\documentclass[10pt, letter, onecolumn]{arxiv}

\usepackage{kantlipsum, lipsum}
\usepackage{dm-colors}
\usepackage{amsmath}
\usepackage{pstricks, pst-node}
\usepackage{verbatim}
\usepackage{multirow}
\usepackage{scalerel}
\usepackage{booktabs}
\usepackage{enumitem}
\usepackage{xspace}
\usepackage{bm}
\usepackage{bbm}
\usepackage{mathtools}
\usepackage{soul}
\usepackage{epsfig}
\usepackage{graphicx}
\usepackage{tcolorbox}
\usepackage{subcaption}
\usepackage{tcolorbox}
\usepackage{amssymb}
\usepackage{colortbl}
\usepackage{csquotes}
\usepackage{etex}
\usepackage{setspace}
\usepackage{colortbl}
\usepackage{tabularx,ragged2e}
\usepackage{placeins}
\usepackage[symbol]{footmisc}
\usepackage[bibstyle=nature,citestyle=numeric-comp,%
            natbib=true,backend=biber,maxbibnames=99,%
            giveninits=false,sorting=none]{biblatex}
\usepackage{nameref}
\usepackage{varioref}
\usepackage[pagebackref=false,breaklinks=false,%
            colorlinks=true,bookmarks=true,citecolor=ourdarkblue,%
            urlcolor=ourdarkblue,linkcolor=ourdarkblue]{hyperref}
\usepackage[noabbrev,capitalize]{cleveref}
\usepackage{etoc}
\graphicspath{{figures/}}

\title{{Towards an AI co-scientist}}

\author[$\ast$, $\ddagger$, 1]{Juraj Gottweis}
\author[$\ast$, $\ddagger$, 2]{Wei-Hung Weng}
\author[$\ast$,1]{Alexander Daryin}
\author[$\ast$,3]{Tao Tu}
\author[2]{\\Anil Palepu}
\author[1]{Petar Sirkovic}
\author[1]{Artiom Myaskovsky}
\author[1]{Felix Weissenberger}
\author[3]{\\Keran Rong}
\author[3]{Ryutaro Tanno}
\author[3]{Khaled Saab}
\author[2]{Dan Popovici}
\author[7]{Jacob Blum}
\author[2]{Fan Zhang}
\author[2]{\\Katherine Chou}
\author[2]{Avinatan Hassidim}
\author[1]{Burak Gokturk}
\author[1]{\\Amin Vahdat}
\author[3]{Pushmeet Kohli}
\author[2]{Yossi Matias}
\author[2]{\\Andrew Carroll}
\author[2]{Kavita Kulkarni}
\author[3]{Nenad Tomasev}
\author[7]{Yuan Guan}
\author[4]{\\Vikram Dhillon}
\author[5]{Eeshit Dhaval Vaishnav}
\author[5]{Byron Lee}
\author[6]{\\Tiago R D Costa}
\author[6]{Jos{\'e} R Penad{\'e}s}
\author[7]{Gary Peltz}
\author[3]{\\Yunhan Xu}
\author[1, $\ddagger$]{Annalisa Pawlosky}
\author[2, $\ddagger$]{Alan Karthikesalingam}
\author[2, $\ddagger$]{Vivek Natarajan}
\affil[1]{Google Cloud AI Research, }
\affil[2]{Google Research, }
\affil[3]{Google DeepMind, \protect\\}
\affil[4]{Houston Methodist, }
\affil[5]{Sequome, \protect\\}
\affil[6]{Fleming Initiative and Imperial College London, \protect\\}
\affil[7]{Stanford University School of Medicine}

\renewcommand{\correspondingauthor}[1]{$\ast$~Equal contributions. \\%
                                      
             $\ddagger$~Corresponding authors: \{juro, ckbjimmy, apawlosky, alankarthi, natviv\}@google.com }

\begin{document}
\begin{refsection}

\begin{abstract}

Scientific discovery relies on scientists generating novel hypotheses that undergo rigorous experimental validation. To augment this process, we introduce an AI co-scientist, a multi-agent system built on Gemini 2.0. The AI co-scientist is intended to help uncover new, original knowledge and to formulate demonstrably novel research hypotheses and proposals, building upon prior evidence and aligned to scientist-provided research objectives and guidance. The system's design incorporates a generate, debate, and evolve approach to hypothesis generation, inspired by the scientific method and accelerated by scaling test-time compute. Key contributions include: (1) a multi-agent architecture with an asynchronous task execution framework for flexible compute scaling; (2) a tournament evolution process for self-improving hypotheses generation. Automated evaluations show continued benefits of test-time compute, improving hypothesis quality. While general purpose, we focus development and validation in three biomedical areas: drug repurposing, novel target discovery, and explaining mechanisms of bacterial evolution and anti-microbial resistance. For drug repurposing, the system proposes candidates with promising validation findings, including candidates for acute myeloid leukemia that show tumor inhibition \textit{in vitro} at clinically applicable concentrations. For novel target discovery, the AI co-scientist proposed new epigenetic targets for liver fibrosis, validated by anti-fibrotic activity and liver cell regeneration in human hepatic organoids. Finally, the AI co-scientist recapitulated unpublished experimental results via a parallel in silico discovery of a novel gene transfer mechanism in bacterial evolution. These results, detailed in separate, co-timed reports, demonstrate the potential to augment biomedical and scientific discovery and usher an era of AI empowered scientists.

\end{abstract}

\maketitle


\section{Introduction}
\label{sec:introduction}

\begin{figure}[htbp!]
\centering
\includegraphics[width=\textwidth,keepaspectratio]{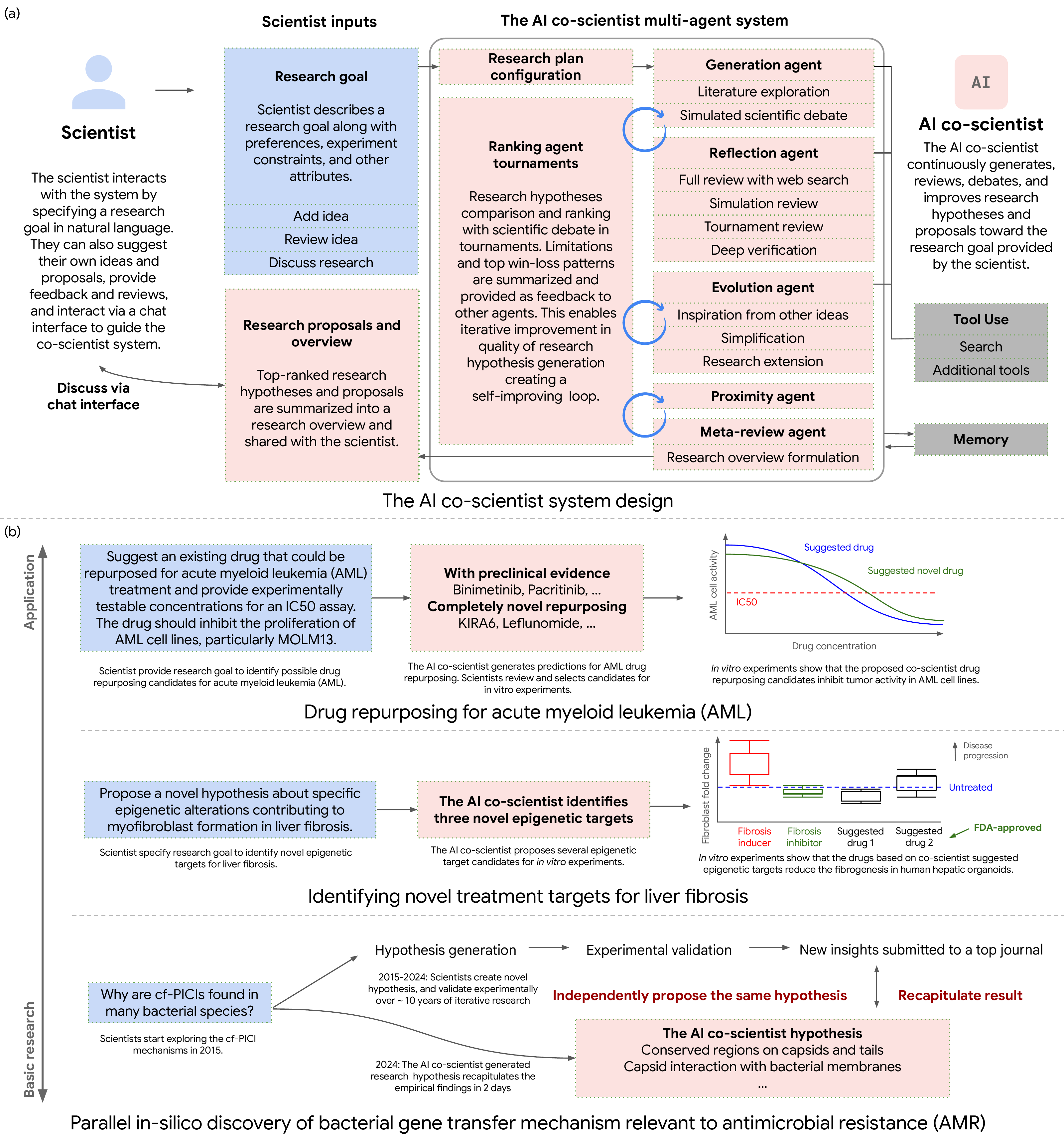}
\vspace{0.1cm}
\caption{\textbf{The AI co-scientist system design and experimental validation summary.} (a) Here, we illustrate the different components of the the AI co-scientist multi-agent system, and its interaction paradigm with scientists. Given a research goal in natural language, the co-scientist generates novel research hypotheses and proposals. The system employs specialized agents — Generation, Reflection, Ranking, Evolution, Proximity (which evaluates relatedness), Meta-review (which provides high level analysis) — to continuously generate, debate, and evolve research hypotheses within a tournament framework. Feedback from the tournament enables iterative improvement, creating a self-improving loop towards novel and high-quality outputs. The co-scientist leverages tools, including web search and specialized AI models to improve the grounding and quality of generated research hypotheses. Scientists can converse with the co-scientist in natural language to specify research goals, incorporate constraints, provide feedback and suggest new directions for explorations via the designated user interface. (b) We perform end-to-end validation of the co-scientist generated hypotheses in three important topics of biomedicine with varied complexity--- suggesting novel drug repurposing candidates for acute myeloid leukemia (AML) (upper panel), discovering novel epigenetic targets for liver fibrosis treatment (middle panel), and recapitulating the discovery of novel mechanism of gene transfer evolution in bacteria key to anti-microbial resistance (lower panel). The co-scientist's hypotheses for these three settings are externally, independently validated by \textit{in vitro} laboratory experiments and detailed in separate preprints co-timed with this work. In the figure, blue denotes expert scientist inputs while red denotes the co-scientist agents or outputs.
}
\label{fig:system-overview}
\end{figure}

Human ingenuity and creativity propel the advancement of fundamental research in science and medicine. However, researchers, particularly in biomedicine, are faced with a breadth and depth conundrum. The complexity of biomedical topics require increasingly deep and specific subject matter expertise, while leaps in insight may still arise from broad knowledge bridging across disciplines. With the rapid rise in scientific publications and the availability of numerous technologies for specialized high-throughput assays, mastery of both discipline-specific depth and trans-disciplinary insight can be challenging.

Despite these challenges, many modern breakthroughs have emerged from trans-disciplinary endeavours. Emmanuelle Charpentier and Jennifer Doudna won the 2020 Nobel Prize in Chemistry for their work on CRISPR~\citep{jinek2012programmable}, which combined techniques and strategies ranging from microbiology to genetics to molecular biology. These benefits of synergy have also been seen beyond experimental biomedicine in numerous other areas of science. Notably, Geoffrey Hinton and John Hopfield combined ideas from physics and neuroscience~\citep{hopfield1982neural, hinton1986learning} to develop artificial intelligence (AI) systems, which were awarded the 2024 Nobel Prize in Physics.

There has been rapid technological progress in AI towards generally intelligent and collaborative systems, which might empower scientists in creatively traversing and expertly reasoning across disciplinary domains. Such systems are capable of advanced reasoning~\citep{guo2025deepseek, jaech2024openai, team2024gemini}, multimodal understanding~\citep{team2024gemini}, and agentic behaviors~\citep{wiesinger2024agents} such as the ability to use tools to solve complex tasks over long time horizons. Further, the trends with distillation~\cite{hinton2015distilling} and inference time compute costs~\citep{team2024gemini, team2024gemma} indicate that such intelligent and general AI systems are rapidly becoming more affordable and available. Motivated by the aforementioned unmet needs in the modern discovery process in science and medicine and building on the advancements in frontier AI~\citep{leslie2024frontier}, we develop and introduce an AI co-scientist system.

The co-scientist is designed to act as a helpful assistant and collaborator to scientists and to help accelerate the scientific discovery process. The system is a compound, multi-agent AI system~\cite{chen2024more} building on Gemini 2.0 and designed to mirror the reasoning process underpinning the scientific method~\citep{gower2012scientific}. Given a research goal specified in natural language, the system can search and reason over relevant literature to summarize and synthesize prior work and build on it to propose novel, original research hypotheses and experimental protocols for downstream validations~(\cref{fig:system-overview}a). The co-scientist provides grounding for its recommendations by citing relevant literature and explaining the reasoning behind its proposals. 

This work does not aim to completely automate the scientific process with AI. Instead, the co-scientist is purpose-built for a ``scientist-in-the-loop'' collaborative paradigm, to help domain experts augment their hypothesis generation process and guide the exploration that follows. Scientists can specify their research goals in simple natural language, including informing the system of desirable attributes for the hypotheses or research proposals it should create and the constraints that the synthesized outputs should satisfy. They can also collaborate and provide feedback in a variety of ways, including directly supplying their own ideas and hypotheses, refining those generated by the system, or using natural language chat to guide the system and ensure alignment with their expertise.

The co-scientist works through a significant scaling of the test-time compute paradigm~\citep{snell2024scaling, brown2019superhuman, silver2016mastering} to iteratively reason, evolve, and improve the outputs as it gathers more knowledge and understanding. Underpinning the system are thinking and reasoning steps---notably a self-play based scientific debate step for generating novel research hypotheses; tournaments that compare and rank hypotheses via the process of finding win and loss patterns, and a hypothesis evolution process to improve their quality. Finally, the agentic nature of the system enables it to recursively self-critique its output and use tools such as web-search to provide itself with feedback to iteratively refine its hypotheses and research proposals.

While the co-scientist system is general-purpose and applicable across multiple scientific disciplines, in this study we focus our development and validation of the system to biomedicine. We validate the co-scientist's capability in three impactful areas of biomedicine with varied complexity: (1) drug repurposing, (2) novel treatment targets discovery, and (3) new mechanistic explanations for antimicrobial resistance (\cref{fig:system-overview}b).

Drug development is an increasingly time-consuming and expensive process~\citep{ringel2020breaking} in which new therapeutics require restarting many aspects of the discovery and development process for each indication or disease (roughly 70\% of drug approvals are for new drugs). In contrast, drug repurposing—--identifying novel therapeutic indications for drugs beyond their original intended use--—has emerged as a compelling strategy to address these challenges~\citep{pushpakom2019drug}. Successful examples of repurposing include Humira (adalimumab) and Keytruda (pembrolizumab), both of which have become among the most successful drugs in history.~\citep{pushpakom2019drug}. The process typically involves analyzing molecular signatures, signaling pathways, drug interactions, clinical trial results, adverse event reports, and other literature-based information~\citep{xia2024drug}, along with off-label use data and, in some cases, patient experiences. However, drug repurposing is limited by several factors: (1) the need for extensive expertise across biomedical, molecular biology, and biochemical systems, (2) the inherent complexity of mammalian biological systems, and (3) the time-intensive nature of traditional computational biology analyses required. We leverage the co-scientist to generate predictions for large-scale drug repurposing, validating the generated predictions using a combination of computational biology, expert clinician feedback, and \textit{in vitro} wet-lab validation approaches. Notably, our system has proposed novel repurposing candidates for acute myeloid leukamia (AML) that inhibit tumor viability at clinically relevant concentrations \textit{in vitro} across multiple AML cell lines.

Unlike drug repurposing, which is a combinatorial search problem through a large but constrained set of drugs and diseases, identifying novel treatment targets for diseases presents a more significant challenge, traditionally requiring extensive literature review, deep biological understanding, sophisticated hypothesis generation and complex experimental validation strategies. The uncertainty of identifying novel treatment targets is significantly greater than in drug repurposing, as it involves not only repurposing existing compounds but also uncovering entirely new components and mechanisms within biological systems. This target discovery process can be inefficient, potentially leading to suboptimal hypothesis selection and prioritization for \textit{in vitro} and \textit{in vivo} experimentation. Given the high costs and time associated with experimental validation, a more effective approach is needed. We probe the capabilities of the co-scientist to propose, rank, and provide experimental protocols for novel research hypotheses pertaining to target discovery. To demonstrate this capability, we focus on liver fibrosis, a prevalent and serious disease, showcasing the co-scientist's potential to discover novel treatment targets amenable to experimental validation. In particular, the co-scientist has suggested novel epigenetic targets demonstrating significant anti-fibrotic activity in human hepatic organoids.

As a third validation of the capabilities of our system, we focus on generation of hypotheses to explain mechanisms related to gene transfer evolution in bacteria pertaining to antimicrobial resistance (AMR) - mechanisms developed by microbes to circumvent drug applications used to fight infections. This is arguably an even more complex challenge than drug repurposing and target discovery and involves understanding of not only the molecular mechanisms of gene transfer (conjugation, transduction, and transformation) but also the ecological and evolutionary pressures that drive the spread of AMR genes: a system-level problem with many interacting variables. This is also an important healthcare challenge with increasing rates of infections and deaths worldwide~\citep{keown2014antimicrobial}. In this validation, researchers instructed the AI co-scientist to explore a topic that had already been subject to novel discovery by their independent research group. Notably, at the time of instructing the AI co-scientist system, the researchers' novel experimental insights had not yet been published or revealed in the public domain. The system was instructed to hypothesize how capsid-forming phage-inducible chromosomal islands (cf-PICIs) exist across multiple bacterial species. The system independently proposed that cf-PICIs interact with diverse phage tails to expand their host range. This \textit{in silico} discovery mirrored the novel and experimentally validated results that expert researchers had already performed, as detailed in the co-timed report~\citep{he2025chimeric, penades2025ai}.

Overall, our key contributions are summarized as follows:
\begin{itemize}[leftmargin=1.5em,rightmargin=0em]
    \item\textbf{Introducing an AI co-scientist.} We develop and introduce an AI co-scientist that goes beyond literature summarization and ``deep research'' tools to assist scientists in uncovering new knowledge, novel hypothesis generation and experimental planning.
    \item\textbf{Significant scaling of test-time compute paradigm for scientific reasoning.} The co-scientist is built on a Gemini 2.0 multi-agent architecture, utilizing an asynchronous task execution framework. This framework allows the system to flexibly allocate computational resources to scientific reasoning, mirroring key aspects of the scientific method. Specifically, the system uses self-play strategies, including a scientific debate and a tournament-based evolution process, to iteratively refine hypotheses and research proposals creating a self-improving loop. Using automated evaluations across 15 complex expert curated open scientific goals, we demonstrate the benefits of scaling the test-time compute paradigm with the AI co-scientist outperforming other state-of-the-art (SOTA) agentic and reasoning models in generating high quality hypotheses for complex problems.
    \item\textbf{Expert-in-the-loop scientific workflow.} Our system is designed for collaboration with scientists. The system can flexibly incorporate conversational feedback in natural language from scientists and co-develop, evolve and refine outputs.
    \item \textbf{End-to-end validation of the co-scientist in important topics in biomedicine.} We present end-to-end validation of novel AI-generated hypotheses through new empirical findings in three distinct and increasingly complex areas of biomedicine: drug repurposing, novel target discovery, and antimicrobial resistance. The AI co-scientist predicts novel repurposing drugs for AML, identifies novel epigenetic treatment targets grounded in preclinical evidence for liver fibrosis, and proposes novel mechanisms for gene transfer in bacterial evolution and antimicrobial resistance. These discoveries from the AI co-scientist have been validated in wet-lab settings and are detailed in separate, co-timed technical reports.
\end{itemize}

\section{Related Works}
\subsection{Reasoning models and test-time compute scaling}
The modern revolution in foundation AI models~\citep{bommasani2021opportunities} and large language models (LLMs) has been largely driven by advances in pre-training techniques~\citep{erhan2010does, radford2018improving}, leading to breakthroughs in models like the GPT and Gemini family~\citep{team2023gemini, achiam2023gpt}. These models, trained on increasingly massive internet-scale and multimodal datasets, have demonstrated impressive abilities in language understanding and generation leading to breakthrough performance in a variety of benchmarks~\citep{chowdhery2022palm, google2023palm2}.  However, a key area of ongoing development is enhancing their \textit{reasoning} capabilities. This has led to the emergence of ``reasoning models'' which go beyond simply predicting the next word and instead attempt to mimic human thought processes~\citep{wei2022chain}. One promising direction in this pursuit is the test-time compute paradigm.  This approach moves beyond solely relying on the knowledge acquired during pre-training and allocates additional computational resources during inference to enable System-2 style thinking---slower deliberate reasoning to reduce uncertainty and progress optimally towards the goal~\citep{kahneman2011thinking}. This concept emerged with early successes such as AlphaGo~\citep{silver2016mastering}, which used Monte Carlo Tree Search (MCTS) to explore game states and strategically select moves, and Libratus~\citep{brown2019superhuman}, which employed similar techniques to achieve superhuman performance in poker. This paradigm has now found applications in LLMs, where increased compute at test-time allows for more thorough exploration of possible responses, leading to improved reasoning and accuracy~\citep{wei2022chain, yao2024tree, zelikman2022star, chen2024more, snell2024scaling,4928, muennighoff2025s1, tu2024towards}. Recent advancements, like the Deepseek-R1 model~\citep{guo2025deepseek}, further demonstrate the potential of test-time compute by leveraging reinforcement learning to refine the model's ``chain-of-thought'' and enhance complex reasoning abilities over longer horizons. In this work, we propose a significant scaling of the test-time compute paradigm using inductive biases derived from the scientific method to design a multi-agent framework for scientific reasoning and hypothesis generation without any additional learning techniques.

\subsection{AI-driven scientific discovery}
AI-driven scientific discovery represents a paradigm shift in how research is conducted across various scientific domains. Recent advancements, particularly the development of large deep learning and generative models, have cemented AI's role in scientific discovery. This is best exemplified by AlphaFold 2's remarkable progress in the grand challenge of protein structure prediction, which has revolutionized structural biology and opened new avenues for drug discovery and materials science~\citep{jumper2021highly}. Other notable examples include the development of novel antibiotics, protein binder design,  and material discovery with AI~\citep{wong2024discovery, zambaldi2024novo, merchant2023scaling}. 

Building on these successes with specialized, bespoke AI models, there has been recent work exploring the even more ambitious goal of fully integrating AI, especially modern LLM-based systems, into the complete research workflow, from initial hypothesis generation all the way to manuscript writing. This end-to-end integration represents a significant shift, presenting both unprecedented opportunities and significant challenges as the field moves beyond specialized AI tools toward realizing the potential of AI as an active collaborator, or even, as some envision, a nascent ``AI scientist''~\citep{lu2024ai, schmidgall2025agent}.

As an example of this shift, Liang et al.~\cite{liang2024can} directly assessed the utility of LLMs for providing feedback on research manuscripts. Through both a retrospective analysis of existing peer reviews and a prospective user study, they demonstrated the significant concordance between LLM-generated feedback and that of human reviewers. Their study, using GPT-4~\citep{openai2023gpt4}, found that a majority of researchers perceived LLM-generated feedback as helpful, and in some instances, even more beneficial than feedback from human colleagues. However, while valuable, their work focuses solely on the feedback stage of the scientific process, leaving open the question of how LLMs might be integrated into the full research cycle, from hypothesis formation to experimental validation and manuscript writing.

Another effort embodying this shift is PaperQA2~\citep{skarlinski2024language}, an AI agent for scientific literature search and summarization. The authors claimed to surpass PhD and postdoc researchers on multiple literature research tasks, as measured both by performance on objective benchmarks and human evaluations. While the system is a useful for synthesizing information, it does not engage in scientific reasoning for novel hypothesis generation.

HypoGeniC, a system proposed by Zhou et al.~\cite{zhou2022least}, tackles hypothesis generation by iteratively refining hypotheses using LLMs and a multi-armed bandit-inspired approach. The process begins with a small set of examples, from which initial hypotheses are generated. These hypotheses are then iteratively updated through exploration and exploitation, guided by a reward function based on training accuracy.  This refined set of hypotheses is subsequently used to construct an interpretable classifier. However, the method's reliance on retrospective data for evaluation means the degree to which the system can generate truly novel hypotheses remains an open question. Furthermore, the system lacks end-to-end validation beyond subjective human evaluations.

Ifargan et al.~\cite{ifargan2025autonomous} present ``data-to-paper'', a platform that systematically guides multiple LLM and rule-based agents to generate research papers, with automated feedback mechanisms and information tracing for verification. However, the evaluations are limited to recapitulating existing peer-reviewed publications and its unclear if the system can generate truly novel, yet grounded hypothesis and research proposals.

Virtual Lab~\citep{swanson2024virtual} is another closely related work. Here, the authors propose a team of LLM agents with a ``principal investigator'' LLM guiding a team of specialized LLM agents to solve a scientific problem. The LLM team receives high level human supervision. The authors demonstrate the utility of their work by leveraging Virtual Lab to design nanobody binders to recent variants of SARS-CoV-2 with experimental validation. While similar in spirit, there are significant design differences to our approach and the generality of the system remains unclear.

Boiko et al.~\citep{boiko2023autonomous} introduced ``Coscientist'', a multi-agent system powered by GPT-4, designed for autonomous execution of complex chemical experiments. This system integrates capabilities such as web and document searching, and code execution, to facilitate independent experimental design, planning, and execution. In addition to similar sounding names, both ``Coscientist'' and our system share the overarching goal of accelerating scientific discovery through AI. However, there are several important distinctions. Notably, ``Coscientist'' is quite narrowly focused on chemical research while ours is much broadly applicable across science. Secondly, our system has important technical innovations that lead to a self-improving system that can uncover new, original knowledge while their approach is a more vanilla-stitching of GPT-4 based agents. Finally, despite the name, ``Coscientist'' prioritizes a high degree of autonomy in experimental execution, directly interfacing with laboratory hardware. Our system, instead, is explicitly designed as a collaborative tool, emphasizing a ``scientist-in-the-loop'' approach and centers on the more cognitive aspects of the research process.

Finally, Lu et al.~\cite{lu2024ai} propose ``The AI Scientist'', a fully automated system designed to conduct research using multiple collaborating LLM agents. These agents handle all stages of the research process, from defining research problems and conducting literature reviews to designing and executing experiments, and even writing up the results. The design shares similarities with our work---the key differences being our focus on the scaling of the test-time compute paradigm to generate high quality hypotheses and research proposals. Secondly, their proposed system has limited automated evaluations; in contrast, our work has a combination of automated, human expert and end-to-end wet lab validations. Finally, our goal is to not to automate scientific discovery, rather to build a helpful AI collaborator for scientists.

\subsection{AI for biomedicine}
More broadly, large AI models are increasingly demonstrating their potential in biomedical science. Both general purpose (GPT-4, Gemini) and specialized LLMs (Med-PaLM, Med-Gemini, Galactica, Tx-LLM) have shown strong performance on biomedical reasoning and question-answering benchmarks~\citep{team2023gemini, achiam2023gpt, singhal2022large, saab2024capabilities, taylor2022galactica, chaves2024tx}. Beyond benchmarks, Med-PaLM 2, was successfully applied to identify causative murine genetic factors for traits such as diabetes, cataracts, and hearing loss~\cite{tu2023genetic}---an early example of hypothesis generation and LLM-assisted discovery. We have also seen the exciting development of specialized foundation and large language models trained on DNA, RNA and protein sequences with a variety of applications~\citep{nguyen2024sequence, lin2023evolutionary, ruffolo2024design, shaw2024protex}. Although AI in biology and medicine often necessitates specialization, the rapid progress of frontier AI models has blurred the distinction. As these models grow in scale, data diversity, and complexity, they continue to achieve breakthroughs in areas once thought to require domain-specific AI. Our co-scientist system, with its modular multi-agent architecture, is flexibly designed to build on top of these advancements in general-purpose frontier AI models and leverage specialized AI models as tools to enhance the capabilities.
 
Drug repurposing is an important area of validation experiments in this work. The traditional approach to this task requires both computational and experimental approaches and a comprehensive understanding of disease-drug interactions~\cite{pushpakom2019drug, krishnamurthy2022drug}. While methods like knowledge graphs with graph convolutional networks have shown promise~\cite{zitnik2018modeling, morselli2021network}, their applicability is limited by the initial knowledge graph's scope.  TxGNN~\cite{huang2024foundation}, an example of a specialized biomedical foundation model with a graph based approach, addresses ``zero-shot'' repurposing for novel diseases but remains dependent on the underlying knowledge graph's quality and lacks sufficient scalability and explainability. Furthermore, no end-to-end validations of the model predictions were reported in the study. In contrast, our work, leveraging state-of-the-art LLMs in the co-scientist setup, is more scalable. We report a combination of expert evaluations and wet-lab experiments to validate the system predictions.

\section{Introducing the AI co-scientist}
\label{sec:methods}

This section describes the technical details, agents, and framework comprising the co-scientist system. The co-scientist employs a multi-agent architecture built upon Gemini 2.0, integrated within an asynchronous task execution framework. This framework allows for flexible scaling of test-time compute resources, facilitating advanced scientific reasoning.

Given a research goal specified by an expert scientist in natural language, the co-scientist generates hypotheses and research proposals that adhere to the following default criteria:

\begin{itemize}
    \item \textbf{Alignment with the provided research goal.} The generated outputs must precisely align with the research goals, preferences and constraints defined by the scientist.
    \item \textbf{Plausibility.} The system outputs should be free of readily apparent flaws. Any potential contradictions with prior literature or established knowledge must be explicitly stated and justified.
    \item \textbf{Novelty.} A key objective of the co-scientist system is to generate novel hypotheses, conjectures, and research plans grounded in prior literature, rather than simply synthesizing existing information (a capability already addressed by existing ``deep research'' tools~\citep{jones2025openai}).
    \item \textbf{Testability.} The system outputs should be amenable to empirical validation within the constraints specified by the scientist.
    \item \textbf{Safety.} The system outputs will be controlled to prevent enabling unsafe, unethical, or harmful research.
\end{itemize}

Aside from these default criteria, the co-scientist can be configured with additional criteria, preferences, and constraints as needed. For instance, it can be configured to generate outputs in formats preferred by the researcher to improve interpretability and readability.

Throughout this section, we employ a recurring example: generating hypotheses for exploring the biological mechanisms of Amyotrophic Lateral Sclerosis (ALS) to illustrate the various components of the co-scientist system. While this example has been reviewed by domain experts, it remains illustrative and may contain errors. Importantly, this example does not aim to suggest potential therapeutic avenues for ALS and should be interpreted with utmost caution. All the examples are listed in the Appendix~\cref{sec:supp_ex}.

\subsection{The AI co-scientist system overview}

At a high level, the co-scientist system comprises four key components:
\begin{itemize}
    \item \textbf{Natural language interface.} Scientists interact with and supervise the system primarily through natural language.  This allows them to not only define the initial research goal but also refine it at any time, provide feedback on generated hypotheses (including their own solutions), and generally guide the system's progress.
    \item \textbf{Asynchronous task framework.} The co-scientist employs a multi-agent system where specialized agents operate as worker processes within an asynchronous, continuous, and configurable task execution framework. A dedicated Supervisor agent manages the worker task queue, assigns specialized agents to these processes, and allocates resources. This design enables the system to flexibly and effectively utilize computational resources and iteratively improve its scientific reasoning capabilities.
    \item \textbf{Specialized agents.} Following inductive biases and scientific priors derived from the scientific method, the process of scientific reasoning and hypothesis generation is broken down into sub-tasks.  Individual, specialized agents, each equipped with customized instruction prompts, are designed to execute these sub-tasks.  These agents operate as workers coordinated by the Supervisor agent.
    \item \textbf{Context memory.} In order to enable iterative computation and scientific reasoning over long time horizons, the co-scientist uses a persistent context memory to store and retrieve states of the agents and the system during the course of the computation.
\end{itemize}

The Gemini 2.0 model is the foundational LLM underpinning all agents in the co-scientist system. The specific co-scientist design was arrived at with iterative developments and is reflective of the current capabilities of the underlying LLMs.

\subsection{From research goal to research plan configuration}

The research goal, specified by the scientist, serves as the entry point to the co-scientist system. Leveraging the multimodal and long context capabilities of Gemini 2.0 models, the co-scientist efficiently processes research goals of varying complexity, from simple statements to extensive documents spanning tens of thousands of natural language tokens or other relevant data (e.g., including hundreds of prior publication PDFs). The research goal may also incorporate specific constraints, attributes, and preferences related to the scientist's particular laboratory setting or field of work.

The co-scientist system then parses the goal to derive a research plan configuration for generating research proposals. This configuration captures the desired proposal preferences, attributes, and constraints. For example, it specifies whether the co-scientist should exclusively propose novel hypotheses. It also specifies the criteria for evaluating hypothesis quality, such as novelty and experimental feasibility. These criteria are then used by the system during its auto-evaluation and improvement phases. The attributes, preferences, and evaluation criteria can all be customized to a given research goal. To illustrate this process, we present an example research goal and its corresponding parsed research plan configuration in Appendix \cref{fig:ex_research_goal}, where the goal is to develop a novel hypothesis related to phosphorylation of the Nuclear Pore Complex (NPC) as a causative mechanism for ALS~\citep{megat2023integrative}.

Based on the research plan configuration, the Supervisor agent initiates the creation of a task queue and begins orchestrating the specialized agents.  The system operates continuously and asynchronously.  Periodically, the Supervisor agent calculates a comprehensive set of summary statistics, reflecting the system's state and progress toward the specified research goal. These statistics inform decisions regarding resource allocation and the determination of whether a terminal state for the overall computation has been reached. The state is periodically written to the associated context memory of the system and leveraged as feedback in subsequent rounds of computation. It also enables easy restarts in-case of any failure in the system components.

\subsection{The specialized agents underpinning the AI co-scientist}

At the core of the co-scientist system are a coalition of specialized agents, each orchestrated by the Supervisor agent. These agents are designed to emulate the scientific reasoning process, enabling them to generate novel hypotheses and research plans. They are also equipped to interact with external tools, such as web search engines and specialized AI models, through application programming interfaces (APIs). These specialized agents are enumerated below:

\begin{itemize}
    \item \textbf{Generation agent.} The agent initiates the research process by generating the initial focus areas, iteratively extending them and generating a set of initial hypotheses and proposals that address the research goal. This involves exploring relevant literature using web search, synthesizing existing findings into novel directions, and engaging in simulated scientific debates for iterative improvement.
    \item \textbf{Reflection agent.} This agent simulates the role of a scientific peer reviewer, critically examining the correctness, quality, and novelty of the generated hypotheses and research proposals. Furthermore, it evaluates the potential of each hypothesis to provide an improved explanation for existing research observations (identified via literature search and review), particularly those that may be under explained.
    \item \textbf{Ranking agent.} An important abstraction in the co-scientist system is the notion of a tournament where different research proposals are evaluated and ranked enabling iterative improvements. The Ranking agent employs and orchestrates an Elo-based tournament~\cite{elo1978rating} to assess and prioritize the generated hypotheses at any given time. This involves pairwise comparisons, facilitated by simulated scientific debates, which allow for a nuanced evaluation of the relative merits of each proposal.
    \item \textbf{Proximity agent.} This agent asynchronously computes a proximity graph for generated hypotheses, enabling clustering of similar ideas, de-duplication, and efficient exploration of the hypothesis landscape.
    \item \textbf{Evolution agent.} The co-scientist's iterative improvement capability relies heavily on this agent, which continuously refines the top-ranked hypotheses emerging from the tournament.  Its refinement strategies include synthesizing existing ideas, using analogies, leveraging literature for supporting details, exploring unconventional reasoning, and simplifying concepts for clarity.
    \item \textbf{Meta-review agent.} This agent also enables the co-scientist's continuous improvement by synthesizing insights from all reviews, identifying recurring patterns in tournament debates, and using these findings to optimize other agents' performance in subsequent iterations. This also enhances the quality and relevance of generated hypotheses and reviews in subsequent iterations. The agent also synthesizes top-ranked hypotheses and reviews into a comprehensive research overview for review by the scientist.
\end{itemize}

\begin{figure}[htbp!]
\centering
\includegraphics[width=\textwidth,keepaspectratio]{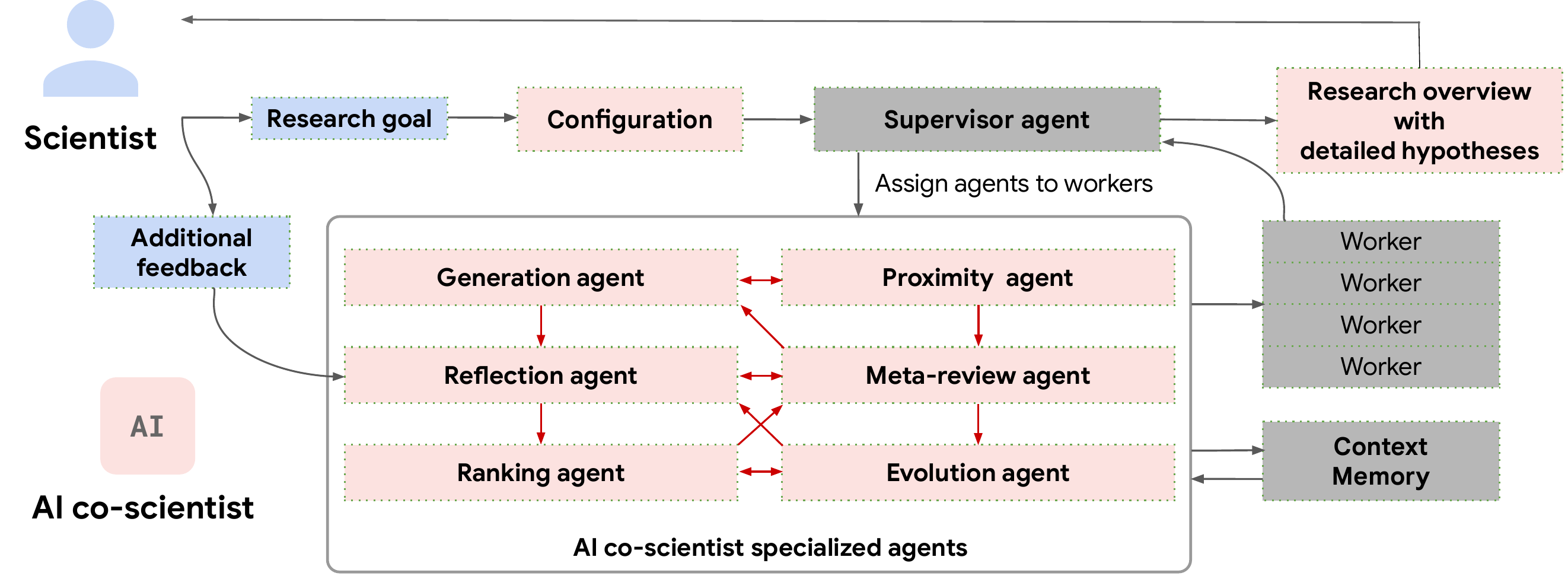}
\vspace{0.1cm}
\caption{\textbf{The AI co-scientist multi-agent architecture design.} The co-scientist accepts a natural language research goal from the user and parses this into a research plan configuration. This plan is then dispatched to the Supervisor agent which evaluates this plan to assigns weights and resources to each specialized agent and subsequently queues them as worker processes in a task queue according to these weights. The worker processes execute the queue of agent actions, and the system ultimately aggregates all information to formulate a research overview with detailed hypotheses and proposals for the scientist. The red boxes in the ``The AI co-scientist specialized agents'' section denote individual agents each with their own unique logic and role. The blue boxes indicate the scientist-in-the-loop inputs and feedback. The dark gray arrows represent the information flow through the co-scientist system, while the red arrows represent the information feedback loop between the specialized agents.}
\label{fig:workflow}
\end{figure}

The Supervisor agent's seamless orchestration of these specialized agents enables the development of valid, novel, and testable hypotheses and research plans tailored to the input research goal.

In summary, the Generation agent curates an initial list of research hypotheses satisfying a research goal. These are then reviewed by the Reflection agent and evaluated in a tournament by the Ranking agent. The Evolution, Proximity, and Meta-review agents operate on the tournament state to help improve the quality of the system outputs.

The Supervisor agent periodically computes and writes to the context memory, a comprehensive suite of statistics, including the number of hypotheses generated and requiring review, and the progress of the tournament. These statistics also include analyses of the effectiveness of different hypothesis generation methodologies (e.g., generating new ideas via the Generation agent vs. improving existing ideas via the Evolution agent). Based on these statistics, the Supervisor agent then orchestrates subsequent system operations, i.e., generating new hypotheses, reviews, tournaments, and improvements to existing hypotheses, by strategically weighting and sampling the specialized agents for execution via the worker processes.

Importantly, the Meta-review agent enables feedback propagation and learning without back-propagation techniques (e.g., fine-tuning or reinforcement learning) \citep{lecun1988theoretical}. The Meta-review agent generates feedback applicable to all agents, which is simply appended to their prompts in the next iteration—a capability facilitated by the long-context search and reasoning capabilities of the underlying Gemini 2.0 models. Through this feedback loop, the co-scientist continuously learns and improves in subsequent iterations with more compute scaling.

Finally, while our work leverages Gemini 2.0, the co-scientist framework is model-agnostic and portable to other similar models or combinations thereof. Future LLM improvements will likely enhance the co-scientist's capabilities. The multi-agent architecture of the co-scientist is depicted and summarized in \cref{fig:workflow}.

We now describe the mechanisms of action of the specialized agents in more detail.

\subsubsection{Generation agent}
The co-scientist Generation agent employs a diverse array of techniques and tools to generate novel hypotheses, such as the following:
\begin{itemize}
    \item \textbf{Literature exploration via web search.} The agent iteratively searches the web, retrieves and reads relevant research articles, and grounds its reasoning by summarizing prior work. It then builds on this summary to generate novel hypotheses and research plans. An example prompt is given in Appendix \cref{fig:EXPLORE_ARTICLES_PROMPT}.
    \item \textbf{Simulated scientific debates.} Here, the Generation agent simulates scientific debates among experts by employing self-critique and self-play techniques. These debates typically involve multiple turns of conversations leading to a refined hypothesis generated at the end.  An example prompt is given in  Appendix \cref{fig:SELF_PLAY_PROMPT}.
    \item \textbf{Iterative assumptions identification.} The agent iteratively identifies testable intermediate assumptions, which, if proven true, can lead to novel scientific discovery. These plausible assumptions and their sub-assumptions are identified through conditional reasoning hops and subsequently aggregated into complete hypotheses.
    \item \textbf{Research expansion.} To identify previously unexplored areas of the hypothesis space, the Generation agent reviews existing hypotheses and the research overview and feedback provided by the Meta-review agent in the previous iteration. This is used to inform additional exploration directions in the research hypothesis space.
\end{itemize}

An example hypothesis and research proposal output from the Generation agent is presented in Appendix \cref{fig:ex_generation} for the aforementioned research goal regarding explaining a basic mechanism related to ALS. The Generation agent also summarizes and categorizes each generated hypothesis, allowing scientists to quickly grasp the core ideas.

\subsubsection{Reflection agent}\label{sec:reflection_agent}
Reviews are integral to the co-scientist's effectiveness in generating novel proposals. The Reflection agent searches relevant prior work (via web search or a dedicated scientist-provided repository), assesses existing experimental evidence for or against a given hypothesis, and rigorously verifies the novelty, correctness, and quality of generated outputs. Effective reviews filter inaccurate and, when stipulated, non-novel hypotheses. Moreover, they also provide feedback to all other agents, driving continuous improvement. The Reflection agent employs the following types of review:

\begin{itemize}
    \item \textbf{Initial review.} Building on the co-scientist's default evaluation criteria, the Reflection agent performs an initial review assessing the correctness, quality, novelty, and a preliminary assessment of safety (ethics) of the generated hypotheses. For a more in-depth discussion on safety considerations see~\cref{sec:safetyethics}. This initial review, which doesn't use external tools like web search, aims to quickly discard flawed, non-novel, or otherwise unsuitable hypotheses.
    \item \textbf{Full review.}  If a hypothesis passes the initial review, the Reflection agent performs a full review, leveraging external tools and web searches to identify relevant articles for improved reasoning and grounding. This review evaluates the hypothesis's correctness, quality, and novelty similar to the initial review but with full literature search. For correctness and quality, the agent scrutinizes underlying assumptions and reasoning. For novelty, it summarizes known aspects of the hypothesis and then judges their novelty based on existing literature. An example full novelty review is shown in Appendix \cref{fig:ex_reflection_novelty}, and an example of review critiques is in Appendix \cref{fig:ex_review_critiques}. A complete full review example is shown in Appendix \cref{fig:ex_reflection_basic}.
    \item \textbf{Deep verification review.} The Reflection agent also conducts a deep verification review, decomposing the hypothesis into constituent assumptions. Each assumption is further broken down into fundamental sub-assumptions, decontextualized, and independently evaluated for correctness to identify invalidating elements for subsequent filtering. Concurrently, the reasons for potential hypothesis invalidation due to incorrect assumptions are summarized. This deep verification helps the co-scientist detect subtle errors within complex hypotheses, such as flaws in reasoning or inaccurate experimental protocols.  An identified error doesn't necessarily invalidate the core hypothesis; the Reflection agent assesses whether the incorrect assumption is fundamental to the hypothesis and incorporates this reasoning into the review. Non-fundamental errors can be addressed during subsequent refinement stages.  An example deep verification review is provided in Appendix \cref{fig:ex_deep_verification} for the previously introduced ALS hypothesis. We also show another example of a deep verification review via probing questions in the context of drug repurposing for AML in Appendix \cref{fig:ex_deep_verification_probe}.
    \item \textbf{Observation review.} In addition, the Reflection agent also explores whether a given hypothesis can account for long-tail observations from prior experimental results. This review aims to determine if the hypothesis can provide insights on existing experimental findings and observed phenomena within relevant articles. For each observation, the agent assesses if the hypothesis is a superior explanation over existing ones, assuming its validity. Positive observations are summarized and appended to the hypothesis. Note that this review often completes without any important findings (as in the case of the ALS hypothesis example). An example prompt to generate observations is provided in Appendix \cref{fig:GENERATE_OBSERVATIONS_PROMPT}. An illustrative example of an observation review is provided in Appendix \cref{fig:ex_observation} in the context of an alternate hypothesis for explaining a mechanism of anti-microbial resistance.
    \item \textbf{Simulation review.} The Reflection agent also reviews hypotheses by simulating them in a step-wise fashion (e.g., simulating the mechanism of action or the proposed experiment in the proposal). This simulation allows the agent to identify and summarize potential failure scenarios. This review method is built on the assumption that frontier LLMs may have developed an internal world model that enables them to simulate and accurately predict various scientific phenomena. 
    \item \textbf{Recurrent/tournament review.} The Reflection agent adapts its full reviews based on the co-scientist's growing knowledge.  By analyzing reviewed hypotheses and results of the tournament conducted by the Ranking agent, 
    the Reflection agent identifies recurring issues and improvement opportunities, refining its reviews accordingly.
\end{itemize}

Additionally, the co-scientist can incorporate reviews from expert scientists to guide ranking and improvements (further discussed in \cref{sec:human_in_the_loop}). We aim to have the Reflection agent's comprehensive set of reviews cover the common methods scientists employ when critiquing and refining research hypotheses and proposals.

\subsubsection{Ranking agent}
The AI co-scientist explores numerous hypotheses and research proposals towards a research goal, necessitating a ranking mechanism to prioritize computational resources toward the most promising candidates. This task is performed by the Ranking agent. The agent uses an Elo-based tournament~\citep{elo1978rating} to automatically evaluate and rank all hypotheses, providing supporting rationale. This ranking serves to communicate to scientists an ordered list of research hypotheses and proposals aligned with the research goal. Despite its assumptions and limitations~\cite{hamilton2024elo}, Elo remains a good proxy for relative ranking, and it has previously been applied to rank extracted patterns and ideas in games~\cite{coulom2007computing}. In the future, extensions may be considered~\cite{kovalchik2020extension}. We set the initial Elo rating of 1200 for the newly added hypothesis.

Because the tournament is computationally intensive, the Ranking agent employs several optimization strategies. Top-ranked hypotheses are compared pairwise in tournament matches through multi-turn scientific debates~\citep{khan2024debating}. This mitigates ordering bias and focuses on novelty, correctness, and testability. Lower-ranked hypotheses undergo single-turn comparisons in a pairwise fashion in their tournament match. The agent concludes each comparison with a decision regarding which hypothesis is better. Appendix \cref{fig:COMPARE_IDEAS_PROMPT} and Appendix \cref{fig:REVISED_PROMPT} show example prompts. Appendix \cref{fig:ex_ranking} shows an example of the Ranking agent conducting a scientific debate match in a tournament to compare two hypotheses.

The Ranking agent prioritizes tournament matches as follows: (1) hypotheses are more likely to be compared with similar ones (based on the Proximity agent's graph, described in the next section); (2) newer and top-ranking hypotheses are prioritized for participation in tournament matches. Successful hypotheses quickly achieve favorable rankings and this informs the tournament state for subsequent iterations.

\subsubsection{Proximity agent}
The Proximity agent calculates the similarity between research hypotheses and proposals, and builds a proximity graph, taking into account the specific research goal. Although it doesn't directly participate in hypothesis generation, the Proximity agent assists the Ranking agent in organizing tournament matches and showcasing a diverse range of ideas related to the research goal. This allows scientists to quickly explore areas of interest and easily identify related concepts.

\subsubsection{Evolution agent}
The Evolution agent continuously refines and improves existing hypotheses and proposals using several approaches including:
\begin{itemize}
    \item \textbf{Enhancement through grounding.} Here the agent attempts to improve hypotheses by identifying weaknesses, generating search queries, retrieving and reading articles, suggesting improvements and elaborating on details to fill reasoning gaps.
    \item \textbf{Coherence, practicality and feasibility improvements.} The agent aims to address issues and creates more coherent hypotheses, potentially rectifying underlying problems with invalid initial assumptions. The agent also refines the hypotheses to make them more practical and feasible. Appendix \cref{fig:FEASIBILITY_IDEA_PROMPT} provides an example of the feasibility improvement prompt.
    \item \textbf{Inspiration from existing hypotheses.} The agent additionally creates new hypotheses inspired by single or multiple top-ranked hypotheses.
    \item \textbf{Combination.} The agent also attempts to directly combine the best aspects of several top-ranking hypotheses to create new hypotheses.
    \item \textbf{Simplification.} The agent simplifies hypotheses for easier verification and testing.
    \item \textbf{Out-of-box thinking.} The agent also explores out-of-the-box ideas by moving away from a subset of hypotheses and generating divergent ones. Appendix \cref{fig:OUT_OF_THE_BOX_SINGLE_IDEA_PROMPT} provides an example prompt for this.
\end{itemize}

The Evolution agent generates new hypotheses; it doesn't modify or replace existing ones. This strategy protects the quality of top-ranked hypotheses from flawed improvements, as each new hypothesis must also compete in the tournament. The evolution of research hypotheses and proposals also allows the co-scientist to iteratively combine different improvement techniques and gradually improve the quality of the results.

\subsubsection{Meta-review agent}
The Meta-review agent plays a crucial role in the co-scientist's feedback loop, enabling self-improvement in scientific reasoning. This agent operates on the tournament state and summarizes common patterns identified in reviews and scientific debates in the tournament matches into a meta-review critique.

By synthesizing insights from all reviews, the meta-review provides valuable feedback to the Reflection agent, leading to more thorough and reliable future reviews.  This helps prevent oversight of critical details. Consider the illustrative example of a identifying a repurposing drug candidate for ALS as a research goal: while only 90\% of individual reviews might correctly identify a blood-brain barrier permeability issue in a proposed candidate, the meta-review ensures that all future reviews by the Reflection Agent definitively address this crucial factor. Hypothesis and research proposal generation is also enhanced by the meta-review's identification of recurring issues. While the Generation agent uses this feedback selectively to avoid over fitting to these review critiques, it helps prevent the recurrence of common issues.

Appendix \cref{fig:REVIEWS_PROMPT_TEMPLATE} provides an example prompt for the meta-review. In Appendix \cref{fig:ex_meta_critique}-\ref{fig:ex_meta_critique_2}, we showcase an example of the summarized meta-review critique generated for the reviews of the previously introduced ALS mechanism hypotheses.

\paragraph{Research overview generation.}
The Meta-review agent periodically synthesizes top-ranked hypotheses into a research overview, providing a roadmap for future research. This overview outlines potential research areas and directions relevant to the research goal, justifying their importance and suggesting specific experiments within each.  Each area includes illustrative example topics. The research overview also serves as an additional input to the Generation agent in subsequent iterations.

The research overview serves to effectively map the boundary of current knowledge relevant to the research goal in the co-scientist system and helps highlight future areas of exploration. In Appendix \cref{fig:ex_overview}-\ref{fig:ex_overview_2}, we show an example of a research overview for the ALS mechanism research goal.

The Meta-review agent can further format these overviews using constrained decoding techniques~\citep{post2018fast} to adhere to common research publication and grant formats (e.g., National Institute of Health (NIH) Specific Aims Page format). We demonstrate the effectiveness of this in subsequent sections.

\paragraph{Research contacts identification.}
The Meta-review agent uses prior literature review to suggest qualified domain experts for research hypotheses and proposal review, including the reasoning behind each suggestion. These potential contacts are summarized in the research overview, providing researchers with additional perspectives and potential avenues for collaborations. An example research contact (with the researcher name redacted) is shown in Appendix \cref{fig:ex_contact}.

\subsection{Expert-in-the-loop interactions with the co-scientist}
\label{sec:human_in_the_loop}
The AI co-scientist empowers scientists to actively guide the system through an expert-in-the-loop design (\cref{fig:workflow}). Scientists can interact with the system in several ways:
\begin{itemize}
    \item Refine the initial research goal in light of the generated hypotheses and research overview.
    \item Provide manual reviews of generated hypotheses (see \cref{sec:reflection_agent} for other system generated review types), which the co-scientist uses to evaluate and improve the hypotheses and proposals.
    \item Contribute their own hypotheses and proposals for inclusion in the tournament, where they are ranked alongside and can be combined with system-generated hypotheses and proposals.
    \item Direct the co-scientist to follow up on specific research directions (for example restricted to a smaller collection of prior publications). When this research is referenced in the research goal, the co-scientist can prioritize generation methods that can access and synthesize it.
\end{itemize}

\subsection{Tool use in AI co-scientist}
The co-scientist leverages various tools during the generation, review, and improvement of hypotheses and research proposals. Web search and retrieval are primary tools, important for grounded, up-to-date hypotheses.

For research goals that explore a constrained space of possibilities (e.g., all known cell receptors of a specific type or all FDA-approved drugs), the co-scientist agents utilize domain-specific tools, such as open databases, to constrain searches and generate hypotheses. The co-scientist can also index and search a private repository of publications specified by the scientist.

Finally, the system can utilize and incorporate feedback from specialized AI models like AlphaFold. We demonstrate this qualitatively with a protein design example in the Appendix~\cref{sec:alphafold}.

\clearpage
\section{Evaluation and Results}

We now discuss the methods for evaluating the AI co-scientist system and the corresponding results. The initial evaluations aim to benchmark and verify the choice of the strategies and metrics underpinning the co-scientist. We then proceed to perform a small-scale evaluation with domain experts to assess the quality of the system.

Furthermore, to assess the practical utility of the system's novel predictions, we also perform end-to-end wet-lab validations (laboratory experiments) of the co-scientist-generated hypotheses and research proposals in three key biomedical applications: drug repurposing, discovering novel treatment targets, and elucidating the mechanisms underlying antimicrobial resistance. The varying complexity and nature of these applications enable a more comprehensive assessment of the system. Notably, all three validations involved expert-in-the-loop guidance and prioritization of experiments. These applications are summarized in Table \ref{tab:task_summary}.

\begin{table}[htbp!]
\centering
\resizebox{1.0\textwidth}{!}{
\begin{tabular}{cccc}
\toprule
\textbf{Application} & \textbf{Drug repurposing} & \textbf{Novel treatment target discovery} & \textbf{\begin{tabular}[c]{@{}c@{}}Explain mechanism of \\ gene transfer evolution\end{tabular}} \\
\midrule
Challenge & Combinatorial search & Identifying novel targets & Understanding complex systems \\
\midrule
Complexity & Medium & High & Very high \\
\midrule
Scale & Moderate, data-limited & Moderate, experiment-limited & Large, data and computation-limited \\
\midrule
Unknown elements & Constrained & Large & Vast and dynamic \\
\bottomrule
\end{tabular}
}
\vspace{0.1cm}
\caption{{Three real-world applications in biomedicine for end-to-end validation of the AI co-scientist.}}
\label{tab:task_summary}
\end{table}

\subsection{The Elo rating is concordant with high quality AI co-scientist results}
\label{sec:result_quality}

The Elo auto-evaluation rating is a key metric that guides the self-improvement feedback loops within the co-scientist system. Therefore, it's necessary to measure and ensure higher Elo ratings correlate with higher quality results. To assess this, we analyzed the concordance between the Elo rating and the system's accuracy on the GPQA benchmark dataset. Ideally, higher Elo ratings should correlate with a higher probability of correct answers.

The GPQA dataset is a challenging, multiple-choice question answering benchmark developed by experts in biology, physics, and chemistry~\citep{rein2023gpqa}. To ensure that the co-scientist Elo rating serves as an objective metric reflecting the validity and correctness of results from the system, we utilized questions within the GPQA diamond set, a subset of the GPQA dataset known for its high difficulty, framing each question as a research goal into our AI system to elicit responses. For each question, we first compared each co-scientist response against the ground truth answer to evaluate its correctness. Then, we categorized all generated responses across all considered questions based on their Elo rating into discrete buckets: Elo rating of 1001-1050, 1051-1100, 1101-1150, etc. in 50 point increments, until the highest rating achieved. Finally, we calculated the average accuracy for each Elo rating bucket, as the percentage of correct responses within each bucket.

We employed the underlying Gemini 2.0 models in the AI co-scientist to create a reference baseline. The reference is necessary because responses within a particular Elo rating bucket are not uniformly distributed across the GPQA questions - some of which are inherently more challenging than others. This non-uniformity could introduce bias into the analysis and potentially lead to erroneous conclusions. We therefore used the reference to generate 32 responses for each GPQA question. The fraction of correct responses from Gemini 2.0 was used as a reference accuracy on that particular question. To determine reference accuracy for a specific Elo bucket, we averaged the reference accuracy of the GPQA questions that had co-scientist responses within that bucket. We also computed the co-scientist accuracy on the GPQA diamond set by using the result with the highest Elo rating for each question and comparing it against the ground truth.

\begin{figure}[htbp!]
\centering\includegraphics[width=0.5\textwidth,keepaspectratio]{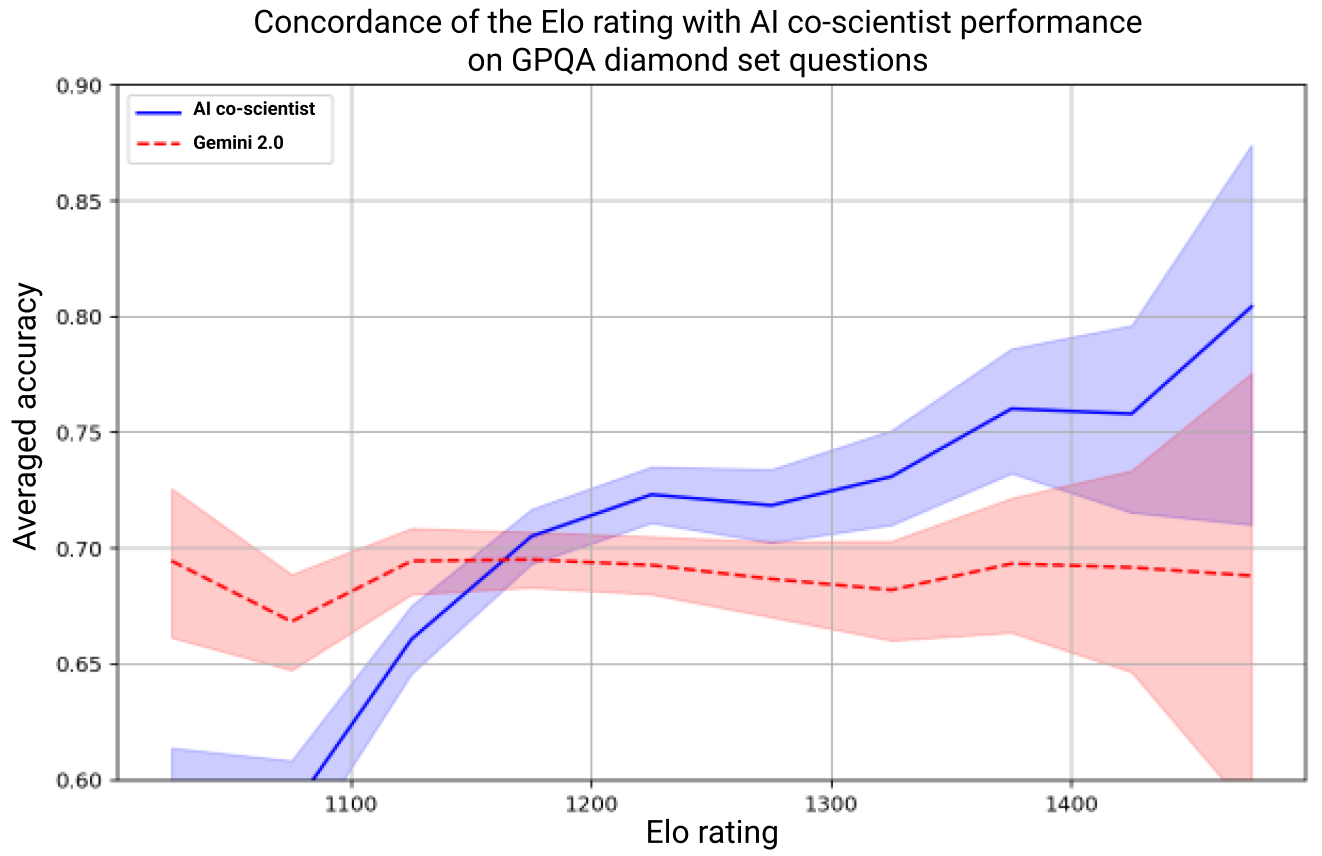}
\vspace{0.1cm}
\caption{\textbf{Concordance of the auto-evaluation Elo metric with AI co-scientist performance on GPQA.} The blue line in the figure shows the average accuracy of co-scientist responses, grouped by their Elo rating. The red line indicates the average accuracy of the corresponding reference Gemini 2.0 responses to the same set of GPQA questions, grouped by Elo rating. Note that Elo metric is auto-evaluated and not based on the ground truth.}
\label{fig:fig_gpqa}
\end{figure}

Our analysis using questions from the GPQA diamond set reveals a concordance between the Elo rating and averaged accuracy of generated co-scientist results, as depicted in~\cref{fig:fig_gpqa}. By selecting the top-rated co-scientist result for each question, the co-scientist achieves a top-1 accuracy of 78.4\%.

\subsection{Scaling test-time compute improves scientific reasoning of the AI co-scientist}
\label{sec:result_ttc}
To evaluate the effects of test-time compute scaling and the co-scientist's progress during iterative scientific reasoning and hypothesis generation, we measured the Elo ratings of the co-scientist generated hypothesis and proposals over the course of the tournament. This analysis was done across 203 distinct research goals curated across broad scientific topics (predominantly in biomedicine, but also included other topics such as mathematics and physics) and entered into the co-scientist system until February 3, 2025.

\begin{figure}[ht!]
\centering
\includegraphics[width=0.95\textwidth,keepaspectratio]{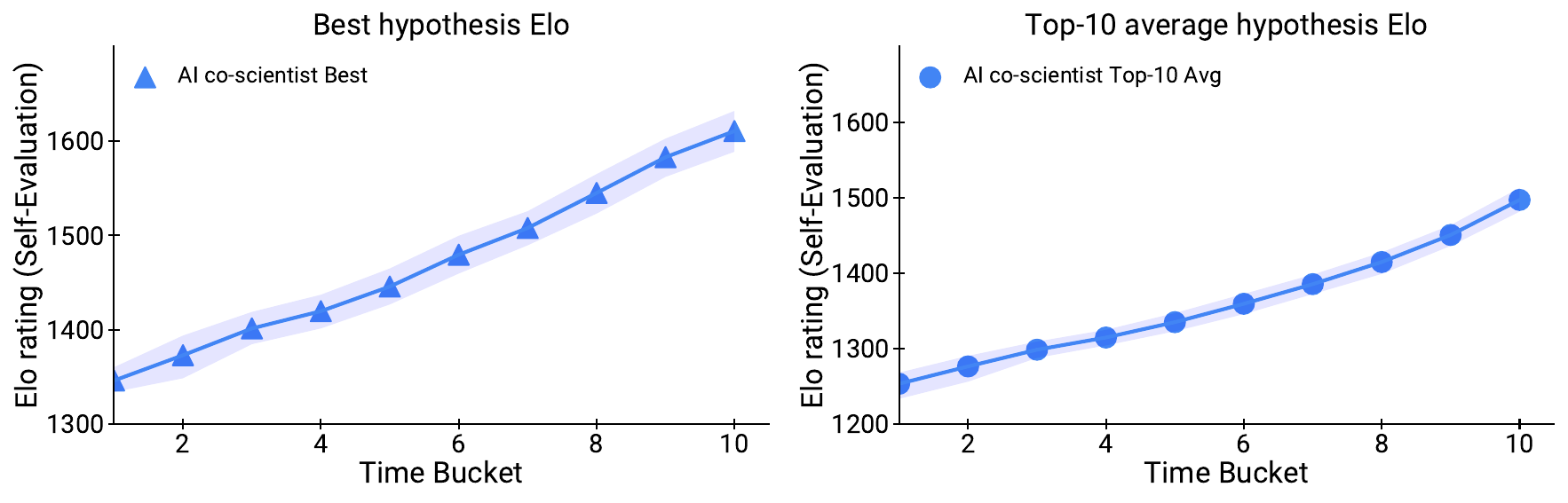}
\vspace{0.1cm}
\caption{\textbf{Impact of scaling test-time compute on AI co-scientist as measured by Elo auto-evaluation.} 
The co-scientist's research hypotheses and proposals were partitioned into ten temporal buckets of equal size, with the last bucket corresponding to the most recently generated results from the system. For each bucket, we determined the maximum individual Elo rating (the ``best Elo'') and the average Elo rating of the top 10 hypotheses across 203 unique research goals. The resulting upward performance trends, across both metrics, suggest improvements in the co-scientist result quality with scaling of test-time compute. Note that the Elo metric is auto-evaluated and not based on independent ground truth.}
\label{fig:fig_elo_all}
\end{figure}

\begin{figure}[ht!]
\centering
\includegraphics[width=0.95\textwidth,keepaspectratio]{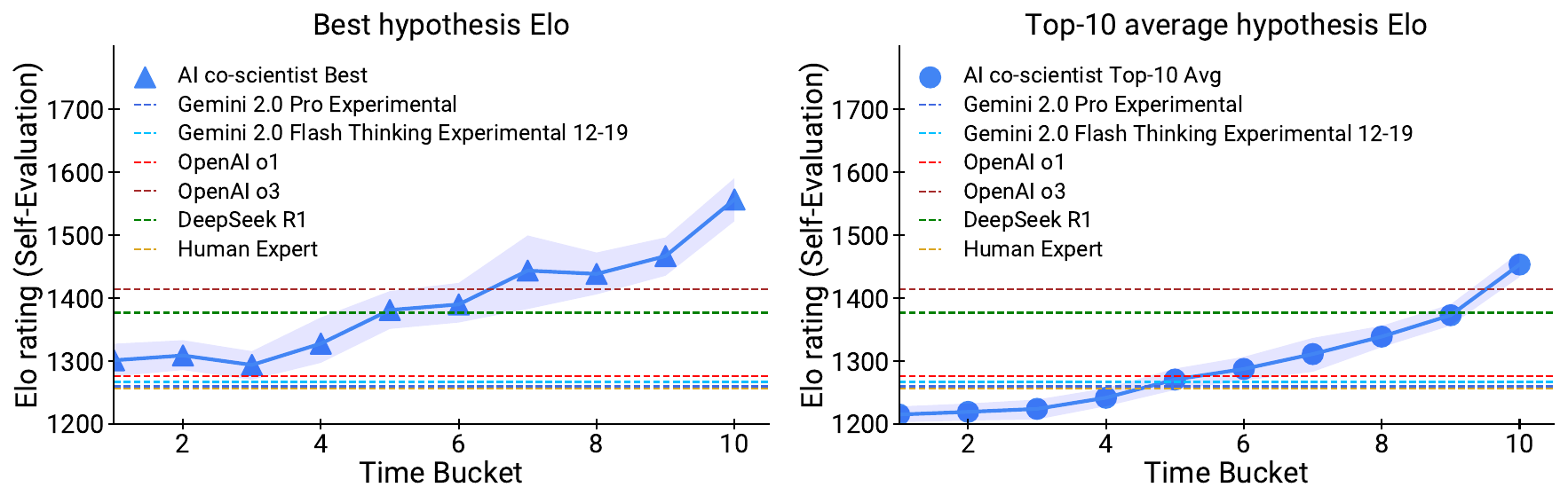}
\vspace{0.1cm}
\caption{\textbf{Comparison of the AI co-scientist with other baselines as measured by Elo auto-evaluation on a subset of 15 challenging expert-curated research goals.} The AI co-scientist's research hypotheses and proposals were partitioned into ten equal-sized temporal buckets, with the last bucket representing the most recent system-generated results. For each bucket, we calculated the maximum individual Elo rating (``best Elo'') and the average Elo rating of the top 10 hypotheses across 15 challenging research goals, curated by biomedical experts. These metrics were then compared to the performance of several baselines: Gemini 2.0 Pro Experimental, Gemini 2.0 Flash Thinking Experimental 12-19, OpenAI o1, OpenAI o3-mini-high, DeepSeek R1, and the ``best guess'' hypotheses provided by the expert curators. These baseline performances are represented by colored dashed lines. On this expert-curated subset of research goals, we consistently observed upward performance trends with increased test-time compute. Furthermore, the co-scientist significantly outperformed both the human experts and the other state-of-the-art LLM baselines, as measured by the Elo metric. It's important to note that the Elo metric is auto-evaluated and not based on independent ground truth. This auto-evaluation metric may favour results with attributes that do not necessarily align with scientists' preferences, quality, or accuracy.}
\label{fig:fig_elo}
\end{figure}

The co-scientist's research hypotheses and proposals were partitioned into ten temporal buckets of equal size. Each bucket corresponded to a sequential 10\% of the total generation time with the first bucket containing the earliest 10\% of generated co-scientist results, while the tenth bucket comprised the most recent 10\%. For each bucket, we determined the average Elo rating of the top 10 hypotheses and the maximum individual Elo rating (the ``best Elo''). These average and best Elo ratings were averaged across 203 research goals and their corresponding tournaments. The resulting performance trends as seen in \cref{fig:fig_elo_all}, across both metrics, serves as a measure of the co-scientist's quality improvement over time. The most recent results demonstrated a significant quality enhancement compared to the initial outputs. Notably, although the Elo rating is not the direct optimization target, its progressive increase emerges from the system's self-improvement feedback loops.

To further contextualize this observation, we focused our analysis on a subset of 15 research goals, curated as challenging problems by seven biomedical experts in their respective fields of expertise. These experts held doctoral degrees in biological or life science disciplines and were actively working as postdoctoral researchers or faculty members at academic biomedical research institutes. These research goals were consistently structured and formatted, encompassing a research title, a clear set of goals, preferences specifying biological or disease areas of interest, desirable solution attributes, and constraints on experimental techniques.

In addition to the research goals, the experts provided their ``best guess'' hypotheses or solutions. We then included outputs from other state-of-the-art LLMs and reasoning models—--Gemini 2.0 Pro Experimental, Gemini 2.0 Flash Thinking Experimental 12-19, OpenAI o1, OpenAI o3-mini-high, and DeepSeek R1—in a tournament along with the expert ``best guess'' and co-scientist for each curated goal. Performance was assessed using the co-scientist Elo rating metric.

The trends previously observed with test-time compute scaling in \cref{fig:fig_elo_all} were consistent within this subset. Furthermore, as shown in \cref{fig:fig_elo}, the co-scientist surpassed the other frontier LLMs and reasoning models in Elo rating with increased computational resources for iterative improvement. Notably, newer reasoning models, such as OpenAI o3-mini-high and DeepSeek R1, demonstrated competitive performance while requiring significantly less compute and reasoning time. Finally, we observed no evidence of performance saturation as measured by Elo, suggesting that further scaling of test-time compute in this paradigm could yield continued improvements in result quality of the co-scientist system. Its worth noting again that the co-scientist architecture is model agnostic and is likely to benefit from further advancements in frontier and reasoning LLMs.

\begin{figure}[htbp!]
\centering
\includegraphics[width=0.5\textwidth,keepaspectratio]{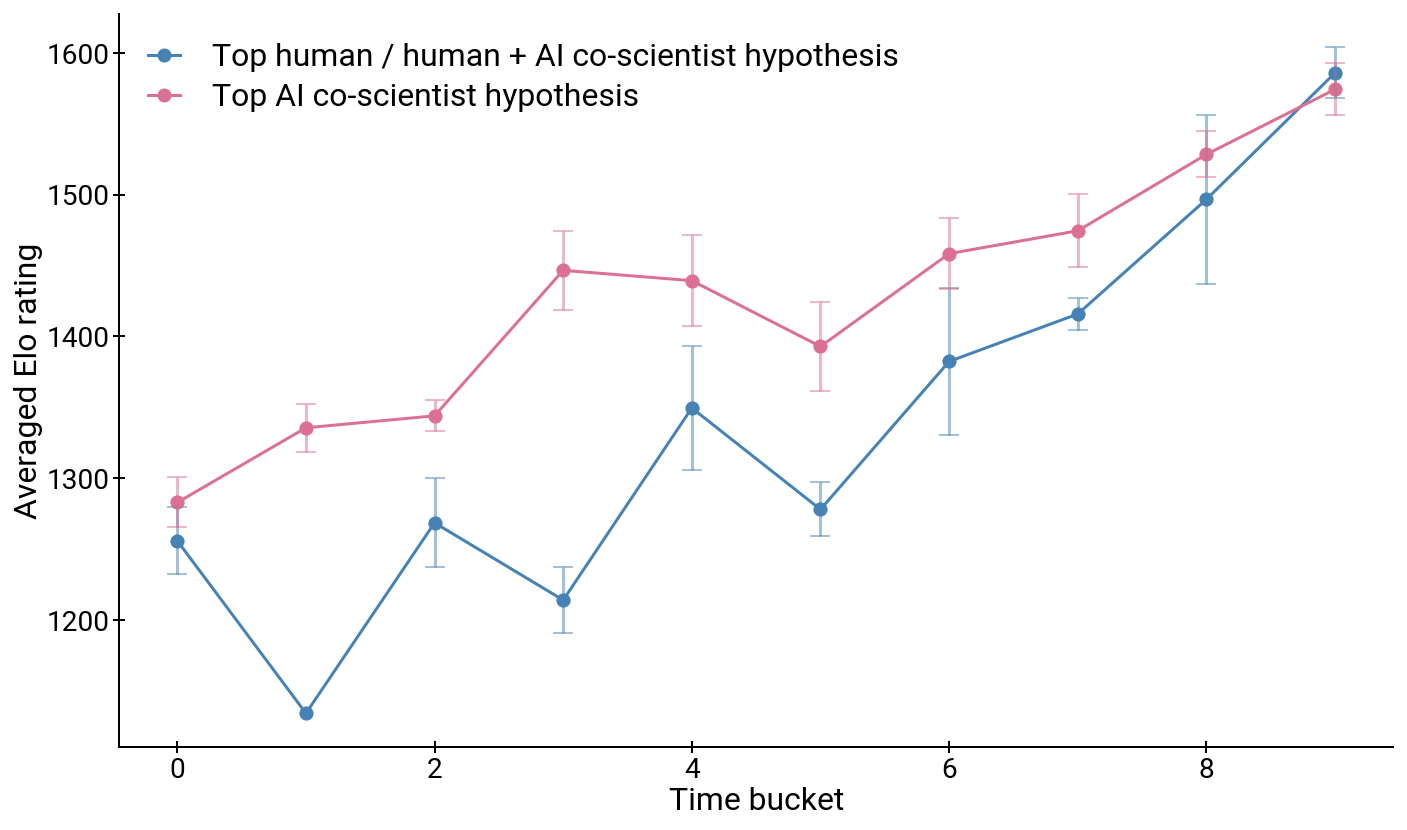}
\vspace{0.1cm}
\caption{\textbf{AI-augmented expertise with the co-scientist through Elo-based auto-evaluation.} Through its self-improvement process, the co-scientist refines and enhances expert ``best guess'' solutions over time, as measured by the Elo rating on a subset of 15 curated research goals. It is important to note that the Elo metric is auto-evaluated and not based on independent ground truth.}
\label{fig:fig_co}
\end{figure}

Building upon the co-scientist system's ability to combine, refine and improve research hypotheses and proposals iteratively, we investigated its potential to improve upon expert ``best guess'' solutions. Consistent with our previous observations, the co-scientist demonstrated the capacity to enhance expert's ``best guess'' solutions over time, as evidenced by the Elo metric in \cref{fig:fig_co}. Notably, the improvement trends initially mirrored those of the co-scientist's autonomously generated solutions but subsequently surpassed them. While this is a preliminary finding requiring further validation, it suggests a promising avenue for capable AI systems, such as the co-scientist, to augment and accelerate the work of expert scientists.

\subsection{Experts consider the AI co-scientist results to be potentially novel and impactful}

To obtain expert feedback and assess preferences, we conducted a small-scale expert evaluation on 11 of the 15 previously curated research goals. We asked the experts who curated the research goals to assess outputs from the AI co-scientist, Gemini 2.0 Flash Thinking Experimental 12-19, Gemini 2.0 Pro Experimental, and OpenAI o1 models. Specifically, they provided a preference ranking (1 being most preferred and 4 being least preferred) and rated the novelty and impact of the proposed solutions on a 5-point scale, ranging from 1 (worst) to 5 (best) following this rubric:
\begin{itemize}
    \item \textbf{Novelty:} Higher-ranked outputs should propose hypotheses that, to the best of the expert's knowledge, have not been previously published in any form. Hypotheses similar to existing proposals, even with minor modifications, should rank lower, and exact replicas of previously proposed and performed experiments should receive the lowest ranking.
    \item \textbf{Impact:} Higher-ranked outputs should address significant open questions in the field and have the potential to substantially advance scientific understanding or lead to practical applications.
\end{itemize}

\begin{figure}[htbp!]
\centering
\includegraphics[width=0.45\textwidth,keepaspectratio]{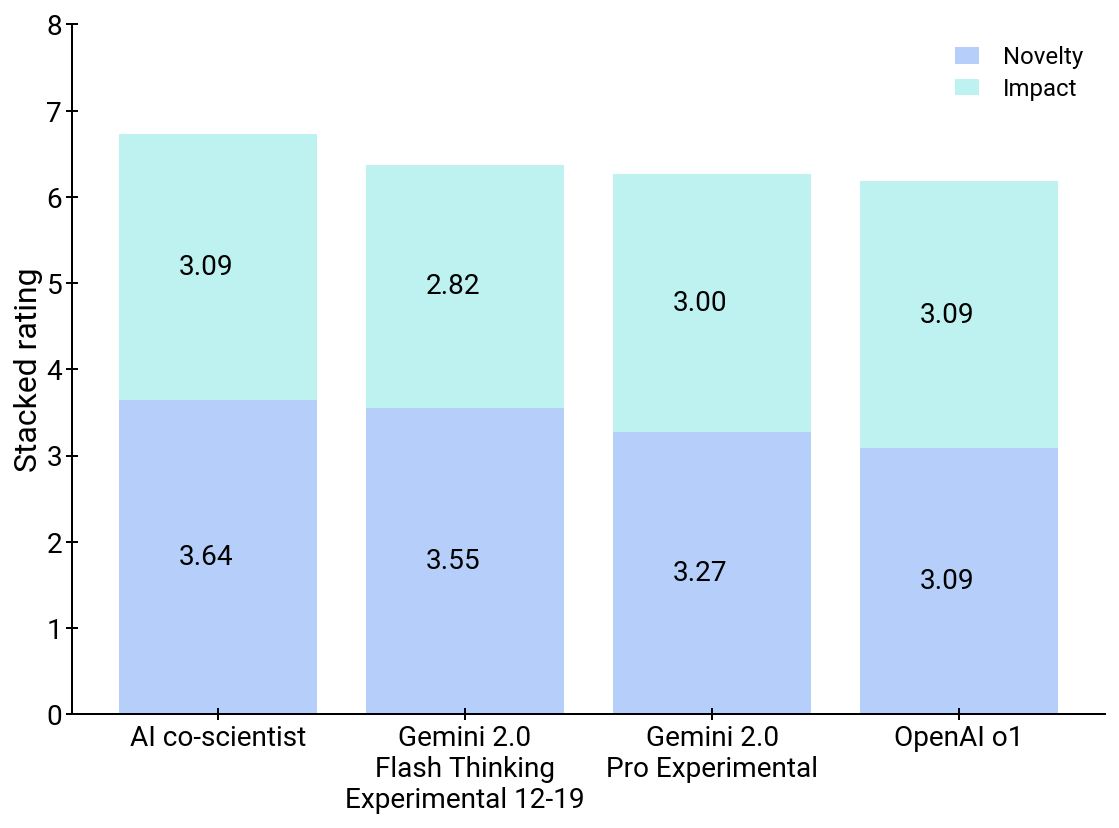}
\includegraphics[width=0.45\textwidth,keepaspectratio]{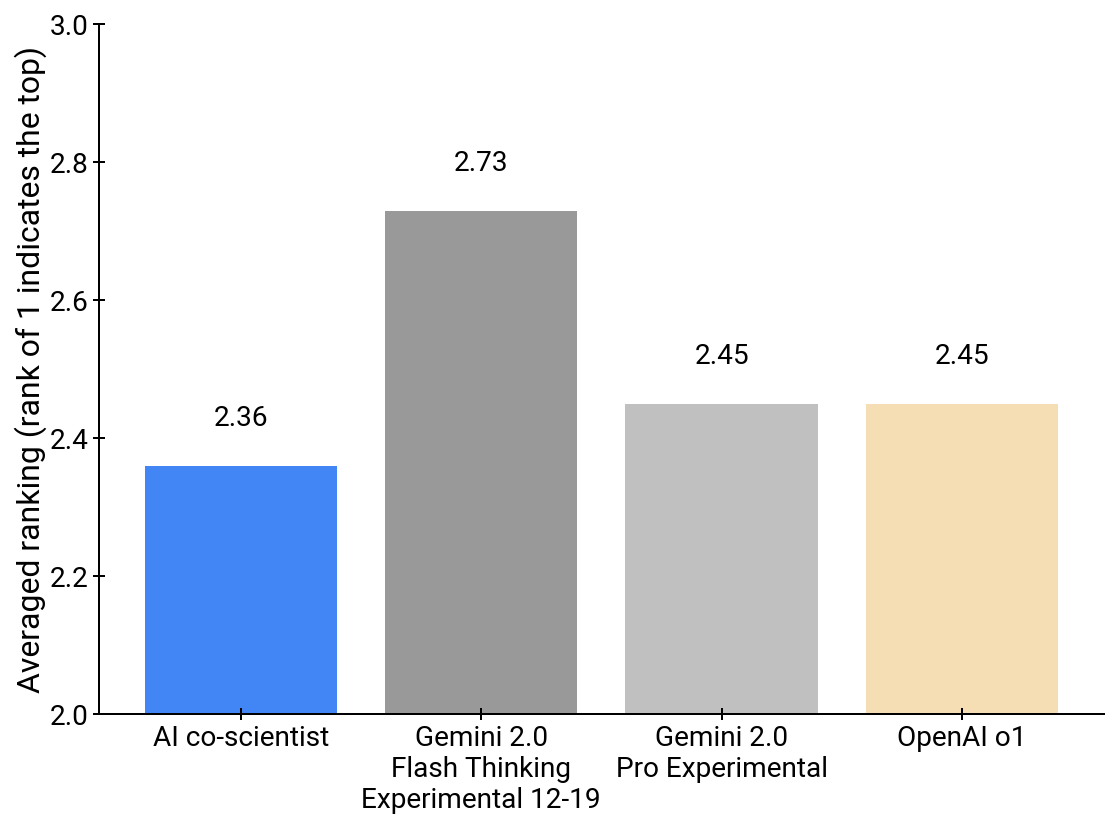}
\vspace{0.1cm}
\caption{\textbf{Expert evaluation of AI co-scientist and other LLM baselines.} Left: Average expert ratings on novelty and impact of the model responses across 11 expert curated research goals. Higher numbers indicate better ratings (1-5). Right: Average expert preference ranking of the results across 11 expert curated research goals generated by AI co-scientist, Gemini 2.0 Flash Thinking Experimental 12-19, Gemini 2.0 Pro Experimental, and OpenAI o1, respectively. Lower numbers indicate better rankings (1-4). The human expert preferences also appear to be concordant with relative Elo ratings as can be inferred from~\cref{fig:fig_elo}. At the same time, its worth noting that these preferences and ratings reflect subjective expert assessments, not objective ground truth.}
\label{fig:fig_human}
\end{figure}

Across 11 expert-evaluated research goals, outputs generated by the AI co-scientist were most preferred and rated higher in novelty and impact axes compared to the other baseline models. Specifically, the co-scientist received an average preference rank of 2.36, and novelty and impact ratings of 3.64 and 3.09 (out of 5) as shown in~\cref{fig:fig_human}. These evaluations reflect subjective expert assessments, not objective ground truth. Notably, the human expert preferences also appear to be concordant with relative Elo ratings as can be inferred from~\cref{fig:fig_elo} and~\cref{fig:fig_human}.

\begin{figure}[ht!]
\centering
\includegraphics[width=0.45\textwidth,keepaspectratio]{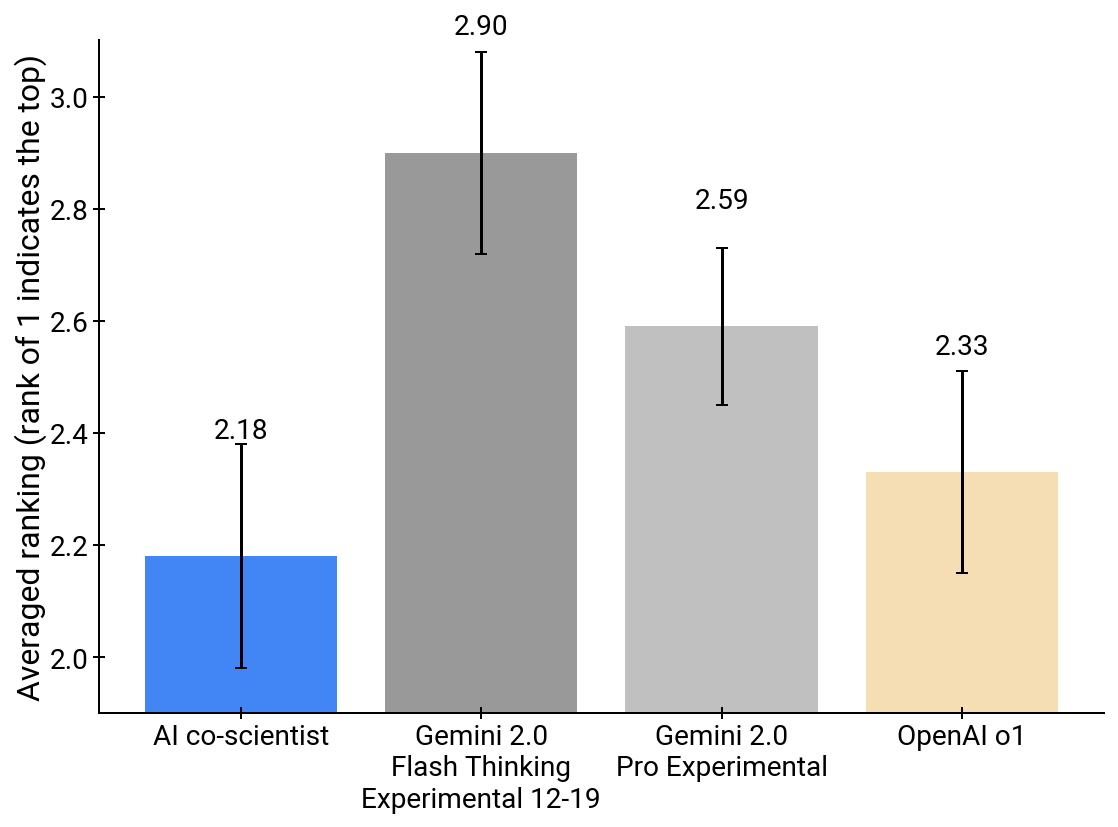}
\includegraphics[width=0.45\textwidth,keepaspectratio]{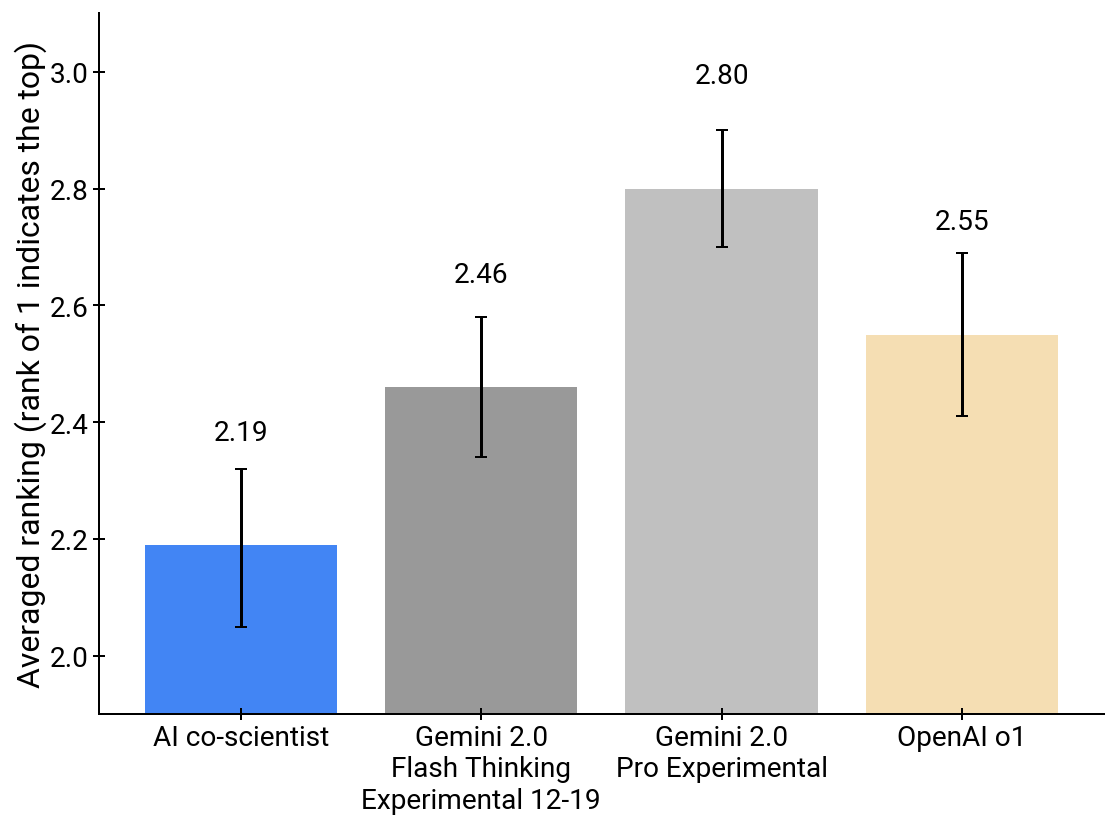}
\includegraphics[width=0.45\textwidth,keepaspectratio]{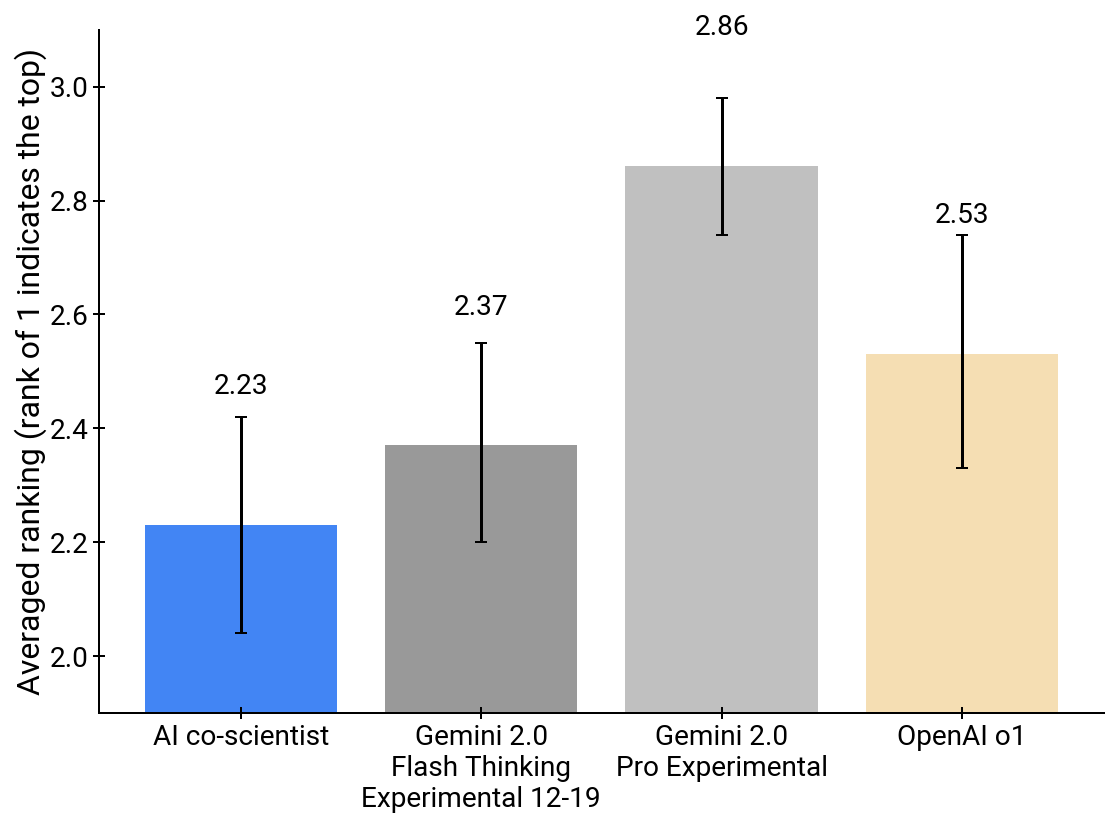}
\includegraphics[width=0.45\textwidth,keepaspectratio]{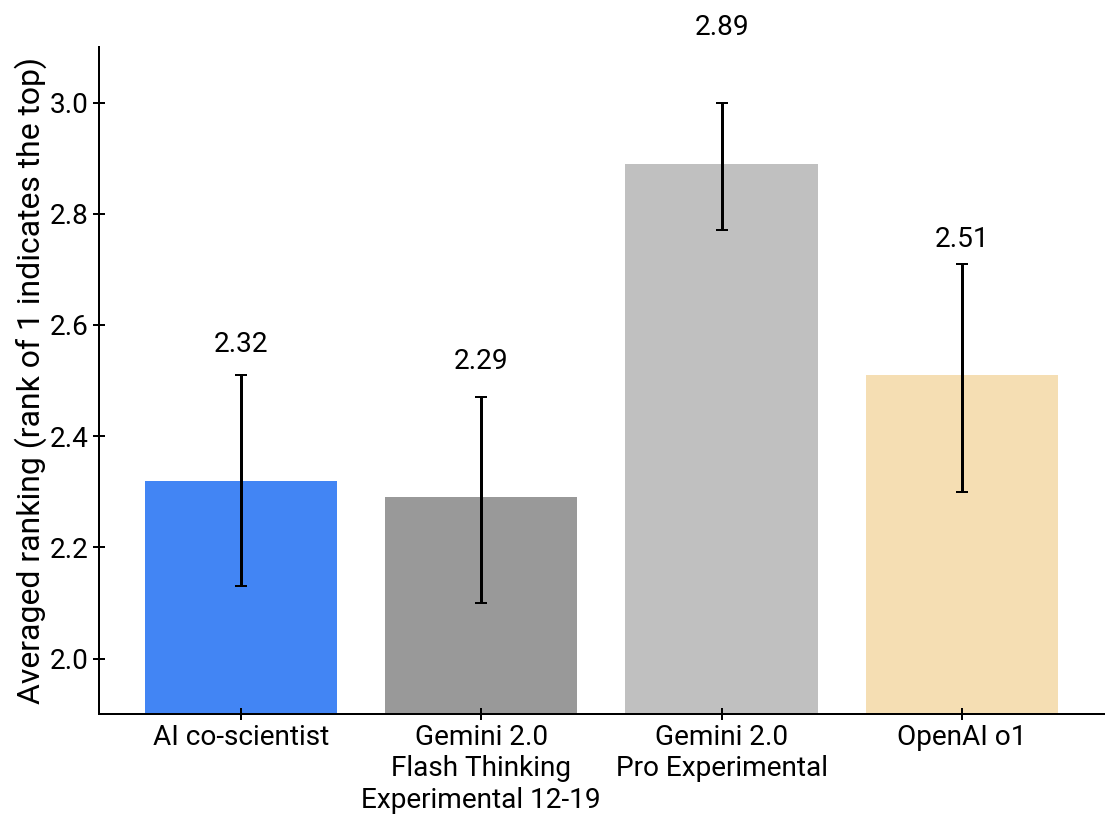}
\vspace{0.1cm}
\caption{\textbf{LLM preference ranking auto-evaluation of AI co-scientist and other baselines.} Averaged preference ranking of results across 11 expert curated research goals generated by AI co-scientist, Gemini 2.0 Flash Thinking Experimental 12-19, Gemini 2.0 Pro Experimental, and OpenAI o1, using four different LLM evaluators: OpenAI o3-mini-2025-01-31 (upper left), OpenAI o1-preview-2024-09-12 (upper right), Gemini 2.0 Pro Experimental (lower left), and Gemini 2.0 Flash Thinking Experimental 01-21 (lower right). Lower numbers indicate better rankings.}
\label{fig:fig_llm_rank}
\end{figure}

We also conducted the same preference ranking evaluation between co-scientist and other LLM and reasoning model baselines using the OpenAI o3-mini-2025-01-31, o1-preview-2024-09-12, Gemini 2.0 Pro Experimental and Gemini 2.0 Flash Thinking Experimental 01-21 as judges. The co-scientist outputs were the most preferred by both the o3-mini, o1 and Gemini 2.0 Pro Experimental models as shown in (\cref{fig:fig_llm_rank}). Due to the small scale of these evaluations, further large-scale studies are necessary for any reliable conclusions. We present a more comprehensive clinical expert evaluation focused on co-scientist proposals for drug repurposing formatted in the NIH Specific Aims Page format in~\cref{sec:result_aims}. 

\subsection{Safety evaluation of the AI co-scientist using adversarial research goals}
\label{sec:result_safety}
The AI co-scientist is designed to empower scientists and accelerate research. However, it's crucial to ensure the system is designed with robust safety principles, given the potential for misuse. This includes addressing dangerous research goals, dual-use objectives, scenarios where safe goals lead to unsafe hypotheses, misleading claims, and inherent biases. While this topic requires extensive investigation beyond the scope of this work, we employed adversarial testing strategies to conduct a preliminary safety analysis of the system. Specifically, we curated a set of 1200 adversarial examples, ranging in complexity, across 40 biomedical and scientific topics using frontier LLMs. We then evaluated whether the AI co-scientist could robustly reject these research goals. In this preliminary analysis, the system successfully passed all checks. Given the sensitive nature of these adversarial research goals, we will not be publicly releasing the dataset, but it can be made available upon request. Collectively, the benchmark, automated, and expert evaluations presented in this section provide compelling evidence of the system's strong capabilities.

\subsection{Drug repurposing with the AI co-scientist}
\label{sec:result_drug}
As previously noted, a rigorous assessment of a system's ability to generate novel hypotheses and predictions for complex research problems necessitates end-to-end validation through wet-lab experiments. However, due to the challenging, time-consuming, and resource-intensive nature of such endeavors, large-scale experimental validation is infeasible. Instead, we strategically selected diverse yet critical biomedical topics to serve as a strong benchmark for the end-to-end system evaluation. Detailed descriptions of these topics follow. Importantly, all three experimental validations were conducted in collaboration with expert scientists, who provided guidance to the co-scientist and prioritized wet-lab experiments.

We begin the discussion of the end-to-end validation of the AI co-scientist with a drug repurposing application. As introduced earlier, drug repurposing is the process of identifying novel therapeutic indications for existing, approved drugs beyond their original use. This approach can accelerate the discovery of treatments for complex and rare diseases, as repurposed drugs have established safety profiles and are readily available. From a technical standpoint, this is a combinatorial search problem involving a large but finite set of drug-disease pairs as noted in \cref{tab:task_summary}. 

Given the co-scientist's ability to synthesize and integrate information across a vast body of scientific and clinical literature, we hypothesized that drug repurposing would be an ideal test of the system's capabilities. The system is general-purpose, capable of providing highly detailed and explainable predictions across all known drug-disease pairs. Here, we focused on the computational biology and wet-lab validation of our co-scientist system in the area of drug repurposing for cancer treatment.

We initially investigated drug-cancer pairs with existing preclinical evidence to validate the plausibility of the hypotheses and predictions generated by the co-scientist (\cref{sec:result_drug_pre}), before expanding to completely novel drug repurposing hypotheses (\cref{sec:result_drug_novel}). The validation of the co-scientist's predictions was performed using a multi-faceted approach, incorporating computational biology analyses, oncologist expert feedback, and \textit{in vitro} wet-lab experiments using cancer cell lines.

\subsubsection{The AI co-scientist suggests plausible drug repurposing candidates as rated by experts}
\label{sec:result_drug_pre}
We constrained the AI co-scientist to explore potential repurposing hypotheses from a curated list of 2300 approved drugs across 33 cancer types (Appendix \cref{sec:eval_data}). To achieve this, we modified the prompts used in the Generation and Ranking agent stages to ensure hypotheses generation in this constrained search space; however, the core co-scientist logic remained unchanged. When formulating the research goal for the co-scientist, we explicitly emphasized the following preferences related to drug repurposing:
\begin{itemize}
    \item Elucidate the known mechanisms of action and impacted biological pathways of the drug.
    \item Identify potential diseases or cancer types that could be treatment targets for the drug.
    \item Explain the potential mechanisms by which the drug could exert therapeutic effects.
    \item Propose alternative mechanisms of action through which the drug might function in the proposed therapeutic context.
    \item Identify the diseases / cancers for which the drug is currently approved.
    \item List the most promising disease / cancer type candidates for repurposing.
    \item Discuss prior research and challenges associated with repurposing the drug.
\end{itemize}

For each drug-cancer pair, we also extracted the Cancer Dependency Map (DepMap) probability of dependency (``DepMap score'')~\citep{depmap2024} (Appendix \cref{sec:eval_compbio}). The DepMap score represents the probability of essentiality for a gene in a given cancer cell lines. We ranked all drug-cancer pairs using a combined metric of the co-scientist review score (ranging from 1 to 5) and the DepMap score (ranging from 0.0 to 1.0). To prioritize the most relevant hypotheses for expert review, we selected only pairs where the co-scientist review score $\geq 4$ and the DepMap score $\geq 0.99$. Note that the DepMap score is primarily meant to function as a sanity check and filter out obviously incorrect candidates but is unlikely to be predictive of efficacy.

Expert oncologists then reviewed the top-ranked drug-cancer pairs, provided feedback, and selected promising repurposing candidates for \textit{in vitro} wet-lab validation (Appendix \cref{sec:eval_wetlab}).

\paragraph{Clinical expert evaluation of drug repurposing proposals in NIH Specific Aims Page format.}
\label{sec:result_aims}
To rigorously evaluate whether the co-scientist-generated hypotheses for drug repurposing fulfill the needs of physicians and scientists, we restructured the co-scientist hypotheses into the NIH-style grant proposal Specific Aims Page (examples in Appendix \cref{fig:aim1_1}-\ref{fig:aim3_2}), and asked a team of six expert hematologists \& oncologists to evaluate the specific aims. 

The NIH Specific Aims Page format follows a standard structure, including disease description, unmet need, proposed solutions, and specific aims. This format was selected because it provides a standardized framework that is widely recognized in the research community, allowing for systematic presentation of complex scientific topics in a manner that facilitates rigorous peer review and enables efficient assessment of scientific merit. The specific aims, which outline the overarching goal, hypothesis, and rationale, requires extensive scientific expertise, comprehensive literature analysis, and robust domain knowledge. We generated cancer drug repurposing hypotheses derived from the co-scientist in the format of NIH Specific Aims Page with additional constrained decoding and self-critique stages to ensure format consistency. The resulting format contextualizes proposed repurposing candidates within known mechanisms based on current literature and then extrapolates to a new disease state. An expert oncologist methodically evaluated and excluded hypotheses that were deemed clinically implausible or had limited potential for successful translation, as well as those falling outside the expertise of the assembled specialist evaluators. This initial screening process employed multiple evidence-based criteria including: (1) pharmacological mechanism incompatibility with tumor biology; (2) unfavorable pharmacokinetic profiles for oncological applications; (3) prohibitive toxicity profiles documented in prior clinical use; (4) confounding effects where apparent survival benefits were attributable to improved management of treatment-related morbidity rather than direct anti-neoplastic activity; and (5) insufficient preclinical evidence supporting antitumor efficacy at clinically achievable concentrations. For example, bisphosphonate agents like pamidronate, while associated with improved outcomes in observational studies of patients with bone metastases, were excluded after critical evaluation revealed their benefits stemmed primarily from reduction of skeletal-related events (such as pathological fractures, spinal cord compression, and bone pain requiring radiation) rather than from disease modifying activity of the drug-candidate. 

Six board-certified hematologists \& oncologists from a single institution - including four domain-specific oncologists specializing in gastrointestinal (GI), breast, gynecologic (GYN) and genitourinary (GU) cancers and two general hematologist \& oncologists, with an average of eight years of clinical experience - evaluated 78 unique drug repurposing hypotheses presented in the NIH Specific Aims Page format (for specific indication distribution and counts, see Appendix~\cref{sec:specific_aims_count}).

The expert raters evaluated the generated Specific Aims based on a modified NIH grant proposal evaluation rubric, consisting of 15 axes focusing on (1) importance of research (significance and innovation) and (2) approach (rigor and feasibility). The raters indicated their agreement level using a five-point scale: ``Strongly Agree'', ``Agree'', ``Neutral'', ``Disagree'', and ``Strongly Disagree''. For each axis, we included several questions covering different aspects of the NIH evaluation criteria. The evaluation rubric is further detailed in the Appendix~\cref{sec:specific_aims_rubric}. Specifically, we ask raters to focus on evaluating the clinical relevance and potential for clinical translation, and not for translational capacity or the design of clinical trials.

\begin{figure}[htbp!]
\centering
\includegraphics[width=0.8\textwidth,keepaspectratio]{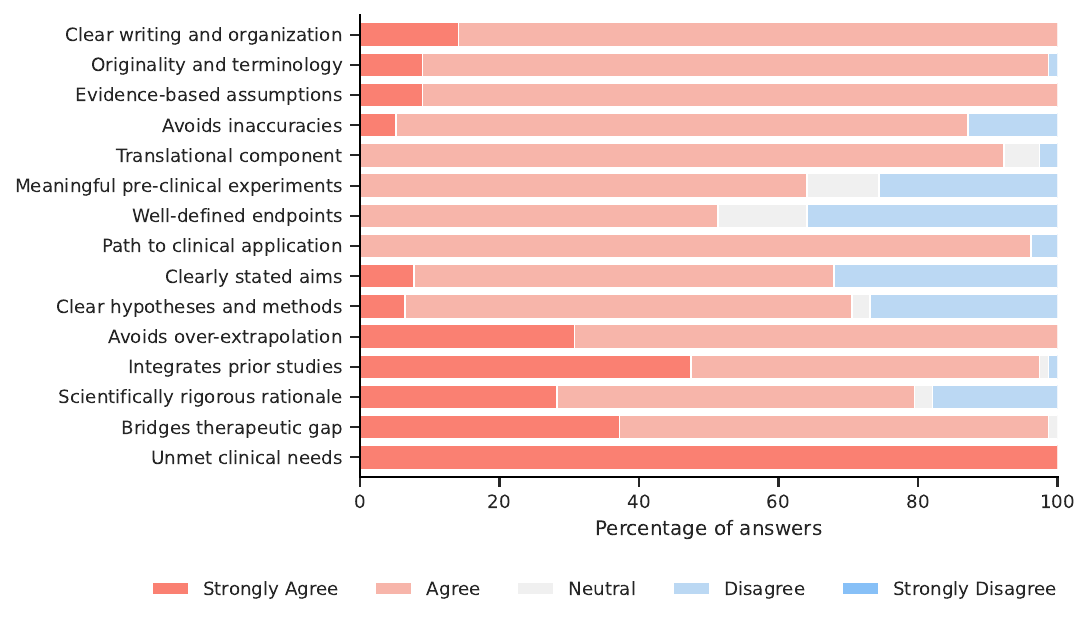}
\vspace{0.1cm}
\caption{\textbf{Clinical expert evaluation for the co-scientist generated drug repurposing hypotheses in the NIH Specific Aims Page format.} Six expert hematologists \& oncologists reviewed 78 drug repurposing research proposals, which the co-scientist had formatted as NIH Specific Aims Pages. The evaluation followed an adapted NIH grant proposal evaluation rubric, detailed in Appendix~\cref{sec:specific_aims_rubric}. Overall, the oncologists judged the Specific Aims proposals from the AI co-scientist to be of high quality across all axes of the rubric.}
\label{fig:fig_aim}
\end{figure}

We observed that expert raters consistently assigned high ratings (``Strongly Agree'' or ``Agree'') to the Specific Aims proposed by the co-scientist across various evaluation criteria (\cref{fig:fig_aim}). Of note, the favorable assessments of co-scientist-generated hypotheses may be partially attributed to expert pre-screening, wherein a clinician eliminated non-viable candidates prior to expert evaluation. Three examples of generated Specific Aims and their respective expert review ratings are detailed in Appendix \cref{fig:aim1_1}-\ref{fig:aim3_2}. 

The generated Specific Aims were assessed by clinical hematologists \& oncologists from a single-center, which might bias the interpretation of the evaluation results, as it may introduce institutional perspectives shaped by local practice patterns, clinical experiences, and research frameworks unique to that setting. While some Specific Aims may be supported by preclinical data, it is important to note that none of the proposed drug candidates have undergone randomized phase III clinical trials necessary to establish efficacy and secure regulatory approval for repurposing to a new indication.

\subsubsection{The AI co-scientist identifies novel drug repurposing candidates for acute myeloid leukemia}

Building upon the positive feedback from clinical experts, we conducted \textit{in vitro} wet-lab validation experiments for drug repurposing hypotheses generated by the co-scientist for acute myeloid leukemia (AML). AML is an aggressive and relatively rare blood cancer characterized by the rapid proliferation of abnormal white blood cells (myeloblasts) in the bone marrow, which displaces healthy blood cells. We focused on this indication due to its aggressive nature and the limited availability of effective therapeutic interventions~\citep{dohner2015acute}.

The cell-line based experiments conducted here serve as an initial biological validation step for co-scientist hypotheses, with intentionally straightforward methodology following established protocols. The simplicity in experimental design is purposeful; our focus is on evaluating the merit of AI co-scientist generated hypotheses rather than developing novel laboratory techniques. Positive results from these experiments should be interpreted as preliminary evidence warranting further investigation through comprehensive pre-clinical studies (e.g., \textit{in vivo} models) and potentially clinical evaluation.

It is important to emphasize that these wet-lab experiments function as a viability checkpoint in the drug repurposing pipeline - not as a replacement for the rigorous pre-clinical and clinical assessment typically required for therapeutic validation. They provide an efficient biological reality check that helps bridge the gap between computational predictions and potential clinical applications, allowing us to rapidly triage AI-generated hypotheses before committing to more resource-intensive validation studies necessary for clinical translation.

\begin{figure}[ht!]
\centering
\includegraphics[width=0.91\textwidth,keepaspectratio]{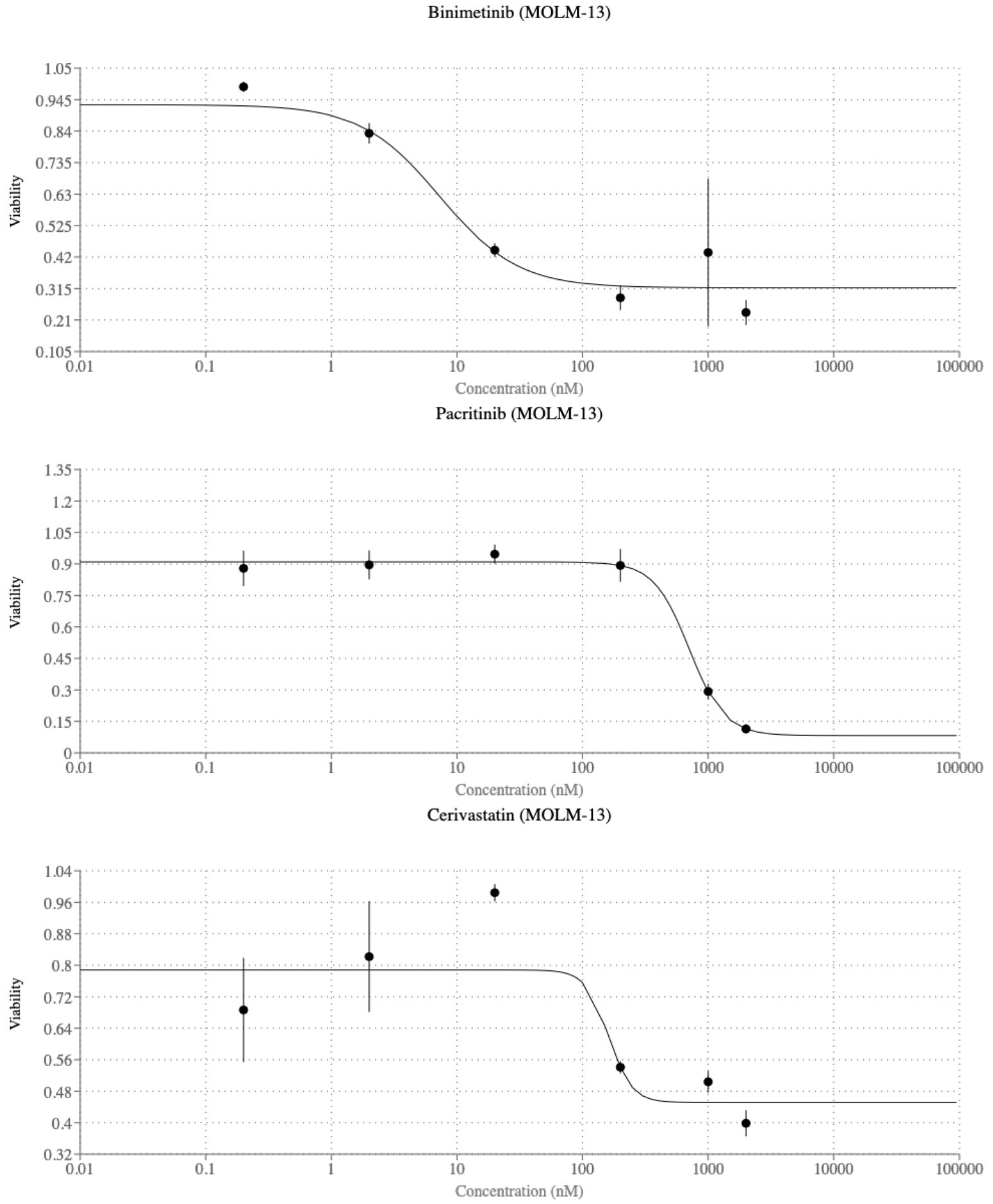}
\vspace{0.1cm}
\caption{\textbf{Dose response curve of the expert selected repurposing drugs with existing evidence.} Binimetinib, Pacritinib, and Cerivastatin inhibit MOLM-13 cell viability. X-axis is the drug concentration (nM), and Y-axis is normalized cell viability (arbitrary unit). Lower cell viability indicates that the selected drug has a stronger inhibition on AML cells.}
\label{fig:fig_wetlab_preclinical}
\end{figure}

\paragraph{Drug repurposing candidate selection process for acute myeloid leukemia.}
The candidate selection for wet-lab experiments was performed with meticulous expert oversight. Thirty top-ranked drug candidates hypotheses were shared with expert oncologists (an example detailed co-scientist output is provided in \cref{sec:drug_output}). The experts evaluated the hypotheses, selecting drug candidates based on their potential to modulate key molecular signaling pathways associated with disease progression and resistance. Note that we did not preclude any hypotheses based on whether single-agent therapy has been studied before, or if the phase III has been approved. The primary selection criterion favored compounds with multi-pathway activity, specifically those influencing dysregulated inflammatory signaling, metabolic reprogramming, and aberrant cell proliferation. Emerging research indicates that these shared biological processes play a critical role in relapse and treatment resistance~\citep{guo2024nf}. Candidates were also chosen based on preclinical mechanistic insights and their relevance to AML biology, including their hypothesized effects on leukemic cell survival, microenvironment interactions, and resistance mechanisms.

Based on potential mechanisms of action, five drug repurposing candidates---Binimetinib, Pacritinib, Cerivastatin, Pravastatin, and Dimethyl fumarate (DMF)---were selected for further wet-lab validation in AML. 

Briefly, Binimetinib is an inhibitor of MEK1 and MEK2, key kinases in the RAS–RAF–MEK–ERK signaling pathway. By inhibiting MEK1/2, binimetinib prevents the activation of ERK (extracellular signal-regulated kinase), thereby blocking downstream signaling that promotes cell proliferation and survival~\citep{van2019biomarker}. Although RAS mutations typically emerge as late events in AML pathogenesis, Binimetinib was included to investigate its potential to modulate RAS-MEK-ERK signaling in treatment-naive AML, where baseline expression levels of this pathway can influence sensitivity to conventional chemotherapeutic agents~\citep{ball2021prognosis}. 

Pacritinib is an oral tyrosine kinase inhibitor that selectively targets JAK2 and FLT3 kinases~\citep{verstovsek2015comprehensive}. By blocking JAK2's kinase activity, pacritinib suppresses the overactive JAK-STAT signaling that drives pathogenic cell proliferation and cytokine production in diseases such as myelofibrosis. It was selected for repurposing due to the dual inhibition of growth signaling pathways: the JAK2/STAT pathway, critical in hematopoietic cell growth and inflammatory signaling, and FLT3-driven proliferative signaling that regulates leukemic cell survival and also facilitates the development of escape pathways to targeted therapies~\citep{perrone2023acute}. 

Dimethyl fumarate (DMF) is an immunomodulatory drug that activates the Nrf2 (nuclear factor erythroid 2–related factor 2) pathway via covalent modification of the cysteine residues on Keap1, the cytosolic protein that normally binds Nrf2 and targets it for degradation. By oxidizing or alkylating the thiol groups of Keap1, DMF destabilizes the Keap1–Nrf2 complex, allowing Nrf2 to escape ubiquitination and translocate into the nucleus. In parallel, DMF also inhibits NF$\kappa$B mediated transcription and was chosen for repurposing due to clinically relevant activity of NF$\kappa$B in AML ~\citep{bosman2016constitutive,hou2023phf6}. 

Finally, the statins (Cerivastatin and Pravastatin) were selected for their potential to induce metabolic reprogramming and directly modulate vesicular transport mechanisms in rapidly proliferating cells~\citep{banker2004cholesterol}.

\paragraph{Laboratory \textit{in vitro} validation of expert-selected drugs with existing evidence.}
Of the five drugs tested, Binimetinib, Pacritinib, and Cerivastatin demonstrated inhibition of cell viability (\cref{fig:fig_wetlab_preclinical}). Notably, Binimetinib, which is already approved for the treatment of metastatic melanoma, exhibited an IC50 as low as 7 nM in AML cell lines (\cref{fig:fig_wetlab_preclinical} and Appendix \cref{fig:fig_wetlab_appendix}). This result shows that the drugs proposed by the co-scientist hold promise as clinically viable drug repurposing candidates. Moreover, this opens the possibility that the co-scientist may be able to expand its hypotheses to novel drug repurposing candidates.

\paragraph{The AI co-scientist proposal of novel drug repurposing candidates for acute myeloid leukemia.}
\label{sec:result_drug_novel}
We aimed to demonstrate the co-scientist's capacity to autonomously propose novel drug repurposing candidates without oversight. Towards this, the system was directed to generate a ranked list of repurposing candidates for AML, including drugs that were not previously repurposed for the target indication and without any prior preclinical evidence. Specifically, we tasked the co-scientist with generating potential novel drug repurposing hypotheses for AML without explicitly relying on additional external inputs, such as the DepMap scores or human expert feedback. We then determined if these novel candidates suggested by co-scientist could be validated in the laboratory, and may therefore have the potential to be repurposed for AML. 

\begin{figure}[ht!]
\centering
\includegraphics[width=0.91\textwidth,keepaspectratio]{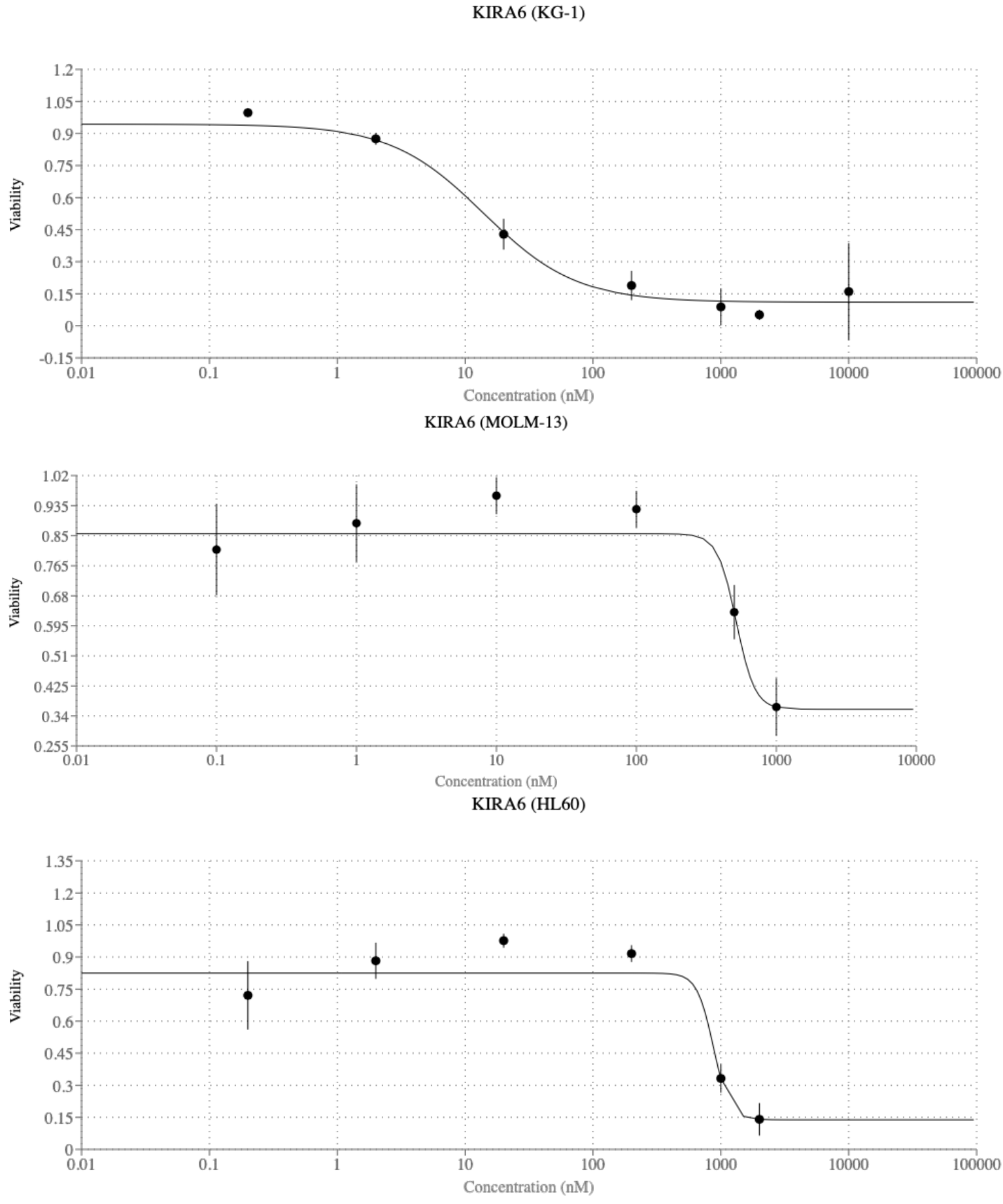}
\vspace{0.1cm}
\caption{\textbf{Dose response curve of novel drug repurposing candidate for AML.} KIRA6 activity inhibiting KG-1, MOLM-13, and HL-60 cell viability all in nM range of drug concentration. X-axis is the drug concentration (nM), and Y-axis is normalized cell viability (arbitrary unit). Lower cell viability indicates that the selected drug has a stronger inhibition on AML cells.}
\label{fig:fig_wetlab_novel}
\end{figure}

For \textit{in vitro} laboratory validation of novel repurposing drugs, the domain experts selected the top 3 candidates from the ranked list, using the criteria that no prior preclinical or clinical data existed with respect to their use to treat AML - Nanvuranlat, KIRA6, and Leflunomide.

To demonstrate the hypothesis and the rationale given by the AI co-scientist for these drug repurposing candidates, the detailed AI co-scientist output, including the novelty review, is provided in~\cref{sec:drug_output} for KIRA6. As can be the seen, the system identifies that targeting IRE1$\alpha$ in the context of AML has been explored before but not with the specific drug, KIRA6. The system suggests an overall moderate level of novelty for the hypothesis.

Of the three drugs tested, treatment with the IRE1$\alpha$ inhibitor KIRA6 showed inhibition of cell viability in three different AML cell lines, KG-1, MOLM-13, and HL-60 cells (\cref{fig:fig_wetlab_novel}). IC50s of KIRA6 were all in nM range, but significantly more effective in KG-1 cells, which had an IC50 of 13 nM, compared to MOLM-13 and HL60 cells, which had IC50s of 517 nM and 817 nM, respectively. Thus, co-scientist was able to suggest a novel candidate for drug repurposing for AML, beyond those that may have been selected through other existing approaches and expert sources. This suggests that the co-scientist system may therefore be capable of generating new, promising hypotheses for researchers to investigate, that may in the future bring new treatments to patients for complex and challenging diseases such as AML.

Translating these insights from co-scientist's drug repurposing hypotheses into clinical practice will be highly challenging, as the complexity of a disease model, patient heterogeneity, and disease variability cannot be fully captured in such limited \textit{in vitro} experiments. Even if a hypothesis generated by co-scientist is well-reviewed by oncologists and supported by preclinical rationale and strong \textit{in vitro} experiments, this does not guarantee \textit{in vivo} efficacy or clinical success. Factors such as drug bioavailability, pharmacokinetics, off-target effects, and patient selection criteria can all impact onward clinical trial outcomes. Moreover, in case of hematological malignancies, the tumor microenvironment and systemic interactions may introduce unforeseen resistance mechanisms, further complicating translation from hypothesis to clinical benefit.

\subsection{The AI co-scientist uncovers novel therapeutic targets for liver fibrosis}
\label{sec:result_liver}
Liver fibrosis is a severe disease that can progress to liver failure and hepatocellular carcinoma, which has few treatment options due to the limitations of available animal and \textit{in vitro} models. However, a recently developed method for producing human hepatic organoids coupled with a live cell imaging system for liver fibrosis provides a new avenue for identification of new treatments for liver fibrosis~\citep{guan2021human,guan2024live,guan2024hepatic}. The AI co-scientist was asked to produce experimentally testable hypotheses concerning the role of epigenetic alterations in liver fibrosis (``A Novel Hypothesis Regarding Myofibroblast Generation in Liver Fibrosis''); and to identify drugs targeting epigenetic modifiers that could be used for treatment of liver fibrosis.

\begin{figure}[ht!]
\centering
\includegraphics[width=0.6\textwidth,keepaspectratio]{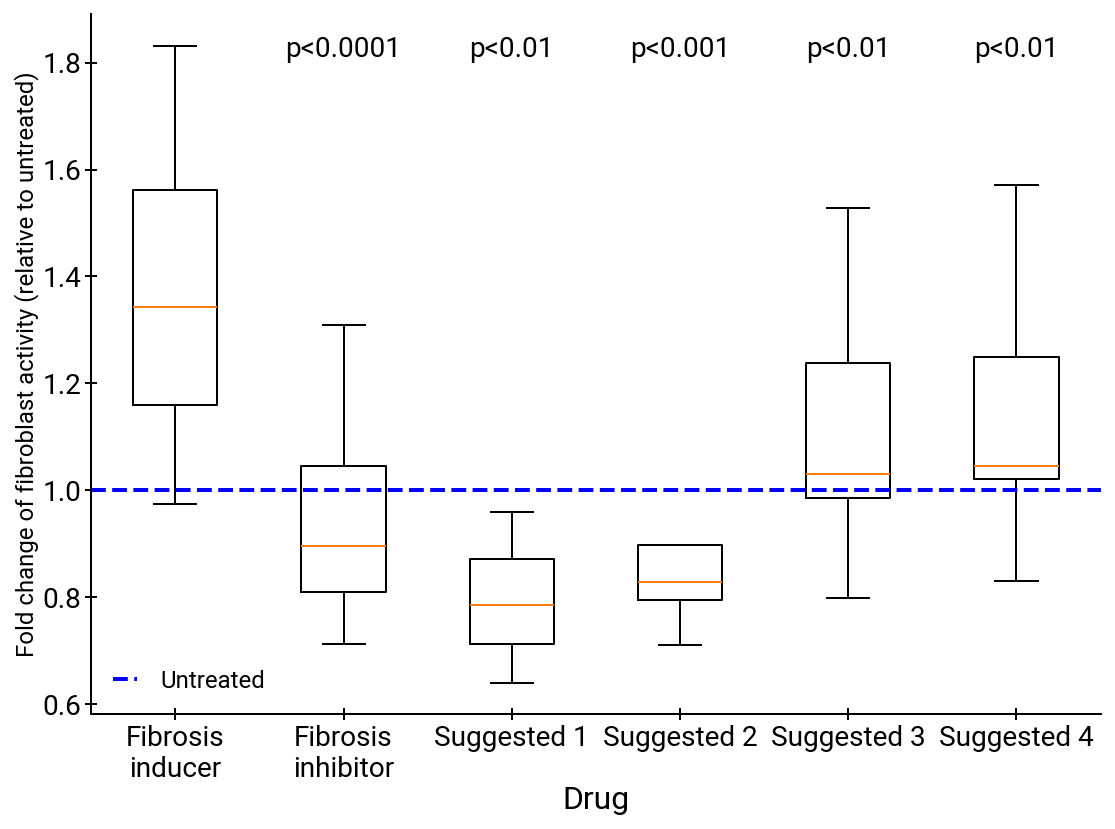}
\vspace{0.1cm}
\caption{\textbf{The co-scientist discovers the novel treatment targets for liver fibrosis.} Four drugs (Suggested 1-4) based on three epigenetic targets suggested by the co-scientist decrease the fold change of the fibroblast activity. The experiments were conducted in the human hepatic organoid system, the fibroblast activity was measured by the percentage fold change of fluorescence labelled on the myofibroblast. `Untreated' indicates the normal control, the fibrosis inducer is the fibrogenic agent TGF-$\beta$, the fibrosis inhibitor indicates the fibrogenic stimulated myofibroblast treated with the fibrogenic agent inhibitor (i.e., TGF-$\beta$ inhibitor), and four suggested drugs indicate the fibrogenic stimulated myofibroblast treated with each of the four drugs based on three AI co-scientist suggested epigenetic targets. The red line in each box is the median fold change of the group. The $p$-value indicates the statistical significance between the fibrosis inducer group and the given group. These results will be further detailed in an upcoming report.}
\label{fig:fig_liver}
\end{figure}

The experts selected three (from fifteen) top-ranked co-scientist generated research hypotheses with a comprehensive research proposal (i.e., experimental design, evaluation methodology, and anticipated results) for exploring the role of epigenetic modifications in liver fibrosis. The co-scientist identified three novel epigenetic modifiers with supporting preclinical evidence that could be targeted by existing agents and provide new treatments for liver fibrosis. Drugs targeting two of the three epigenetic modifiers exhibited significant anti-fibrotic activity in hepatic organoids without causing cellular toxicity (\cref{fig:fig_liver}). Since one of them is an FDA-approved drug for another indication, this creates an opportunity to re-purpose a drug for treatment of liver fibrosis. These results will be detailed in an upcoming technical report.

\subsection{The AI co-scientist recapitulates a breakthrough in antimicrobial resistance}
\label{sec:result_amr}
Understanding the mechanisms of antibiotic resistance is crucial for researchers to develop effective treatments against infectious diseases. We focused on capsid-forming phage-inducible chromosomal islands (cf-PICIs), which play a pivotal role in antibiotic resistance. These mobile genetic elements, unlike typical phages and other PICIs, possess a remarkable ability to transfer between diverse bacterial species, carrying with them virulence and antibiotic resistance genes. We sought to understand the evolutionary rationale behind the existence of cf-PICIs across multiple bacterial species in order to develop solutions to combat antimicrobial resistance.

The primary objective was to leverage the AI co-scientist to generate a research proposal aimed at elucidating the molecular mechanisms of bacterial evolution underlying the broad host range of cf-PICIs and developing strategies to curb the spread of antibiotic resistance. We specifically focused on the observation that identical cf-PICIs, such as PICIEc1 and PICIKp1, are found in clinically relevant bacterial species, including WHO priority pathogens like \textit{Escherichia coli} and \textit{Klebsiella pneumoniae}.

In a co-timed study~\citep{he2023large} currently undergoing the peer-review process at an established journal in the field, genomic and experimental studies revealed a novel mechanism explaining how identical cf-PICIs can be found in different bacterial species. Knowing the answer to that question (but without it yet being available in the public domain), we investigated whether the co-scientist could independently discover the same, or similar, research hypotheses. We provided the co-scientist with a single-page document containing general information, including a brief background on phage satellites and two relevant research articles. The first paper described the original discovery of cf-PICIs~\citep{alqurainy2023widespread}, and the second paper introduced a computational technique for identifying phage satellites in bacterial genomes~\citep{de2023identification}. We then challenged co-scientist to address the question of why cf-PICIs, but not other types of PICIs or satellites, are readily found across diverse bacterial species, and what mechanism underlies this phenomenon.

\begin{figure}[ht!]
\centering
\includegraphics[width=\textwidth,keepaspectratio]{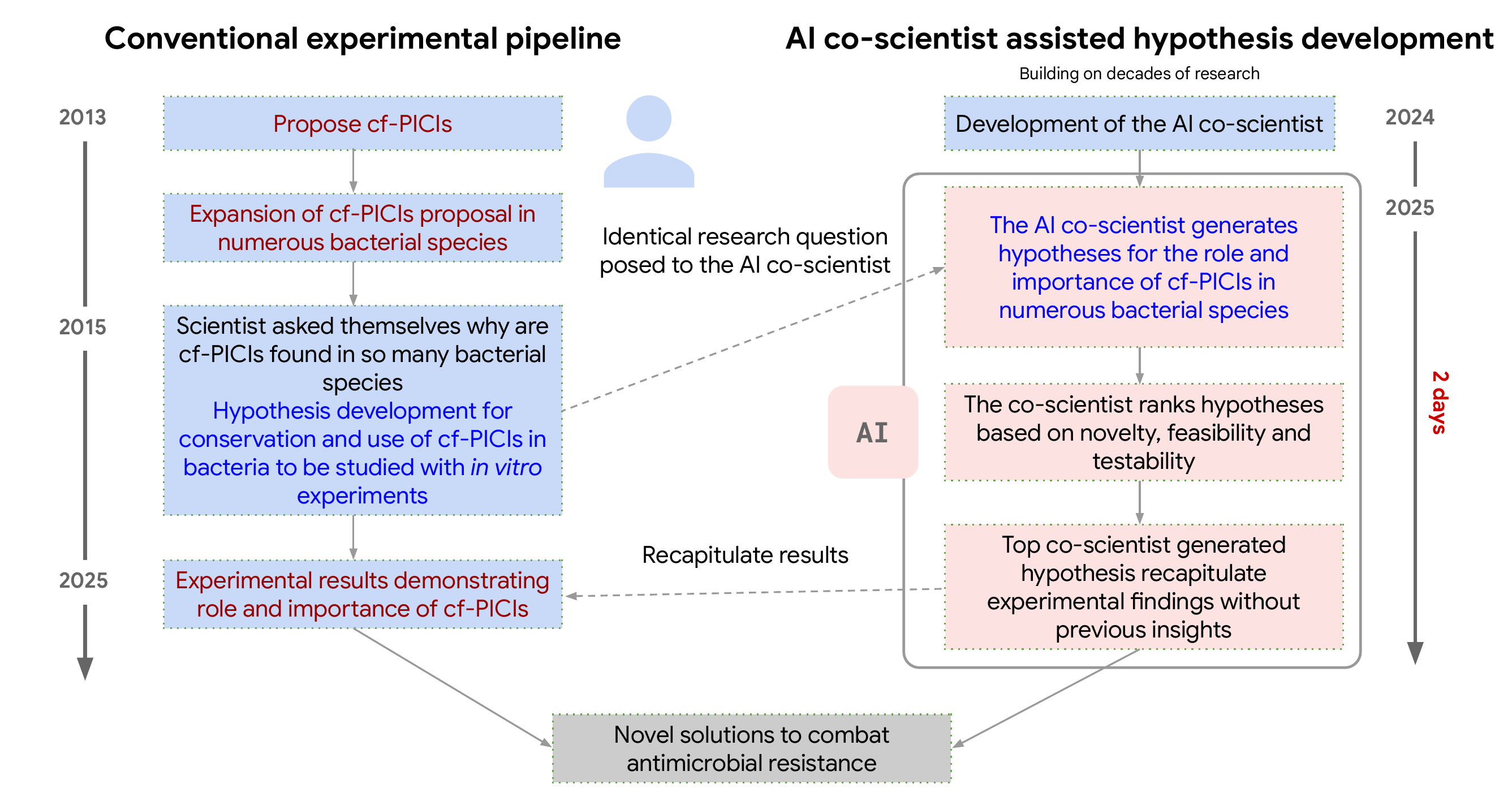}
\vspace{0.1cm}
\caption{\textbf{Timeline of the conventional experimental validation and the AI co-scientist-assisted hypothesis development for capsid-forming phage-inducible chromosomal islands (cf-PICIs), key to antibiotic resistance (AMR).} Blue box: Scientist inputs. Red box: The AI co-scientist system. Red text: wet lab experimental setting. Blue text: research hypothesis generation.}
\label{fig:fig_amr}
\end{figure}
\vspace{-0.3cm}

The co-scientist independently and accurately proposed a groundbreaking hypothesis---that cf-PICIs elements interact with diverse phage tails to expand their host range---as its top-ranked suggestion~\citep{penades2025ai}. This finding was experimentally validated in the independent research study, which was unknown to the co-scientist during hypothesis generation~\cite{he2025chimeric}. Its worth noting that while the co-scientist generated this hypothesis in just two days, it was building on decades of research and had access to all prior open access literature on this topic.

(\cref{fig:fig_amr}). Specifically, the co-scientist suggested the following research topics for the given research goal regarding cf-PICIs:

\begin{itemize}
    \item \textbf{Capsid-tail interactions.} Investigate the interactions between cf-PICI capsids and a broad range of helper phage tails. This topic aligns precisely with the unpublished manuscript's primary finding: that cf-PICIs interact with tails from different phages to expand their host range, a process mediated by cf-PICI-encoded adaptor and connector proteins.
    \item \textbf{Integration mechanisms.} Examine the mechanisms by which cf-PICIs integrate into the genomes of diverse bacterial species.
    \item \textbf{Entry mechanisms.} Explore alternative cf-PICI entry mechanisms beyond traditional phage receptor recognition.
    \item \textbf{Helper phage and environmental factors.} Investigate the role of helper phages and broader ecological factors in cf-PICI transfer.
    \item \textbf{Alternative transfer and stabilization mechanisms.} Explore other potential transfer mechanisms, such as conjugation, extracellular vesicles, and unique stabilization strategies, that might contribute to cf-PICI's broad host range.
\end{itemize}

The convergence of the conventional and AI co-scientist approaches on the same novel finding underscores the potential of the co-scientist to augment, complement and accelerate scientific endeavors (\cref{fig:fig_amr}). Further results and comprehensive details are available in the companion report~\citep{penades2025ai}.

\section{Limitations}
\label{sec:limitations}
We are encouraged by the early promise of the AI co-scientist evaluations, which highlight its potential to augment scientific research.
However, the system has several limitations. Responsible innovation necessitates thoughtful consideration of these alongside the potential impacts to researchers and scientific research. 

\paragraph{Limitations with literature search, reviews and reasoning.} The reviews undertaken by the AI co-scientist system may miss critical prior works due to reliance on open-access literature. In the presented work, the AI co-scientist does not access the entire published literature due to compliance with license or access restrictions where applicable. The system may also omit consideration of prior work on occasions where it has incorrectly reasoned that the work is not relevant.

\paragraph{Lack of access to negative results data.} The AI co-scientist system's use of only open published literature means it likely has limited access to negative experimental results or records of failed experiments. It is known that such data may be more rarely published than positive results, yet experienced scientists working in the field may nonetheless possess and utilize this knowledge to prioritize research~\citep{brazil2024illuminating}. Strategies to overcome this phenomenon might further improve the performance of the co-scientist as a tool for scientific discovery.

\paragraph{Improved multimodal reasoning and tool-use capabilities.} Some of the most interesting data in scientific publications is not written in text but may be encoded visually in figures and charts. However, even state-of-the-art frontier models may not comprehensively utilize such data with optimal  reasoning~\citep{roberts2024scifibench} and the AI co-scientist system is unlikely to be an exception. Stronger benchmarks and evaluations are necessary to improve these capabilities. We have also not evaluated the ability of our system to reason over and integrate information from domain-specific biomedical multimodal datasets (such as large multi-omics datasets) and knowledge graphs. More work is needed to integrate the AI co-scientist system with specialized scientific tools, AI models and databases, and evaluate the ability to utilize them effectively.

\paragraph{Inherited limitations of frontier LLMs.} LLM limitations include imperfect factuality and hallucinations, which may be propagated in the co-scientist system. The system's reliance on existing LLMs and web-search, while providing immediate access to broad knowledge, may propagate errors of factuality, biases or limitations present in those resources.

\paragraph{Need for better metrics and broader evaluations.} While the current AI co-scientist evaluation includes AI auto-ratings, expert reviews and targeted \textit{in vitro} validations, the evaluation of system performance remains preliminary. A comprehensive, systematic evaluation across diverse biomedical and scientific disciplines is necessary to determine the generalizability of co-scientist. Furthermore, the system requires continued improvement to produce outputs that meet the rigor and detail of high-quality publications. Furthermore, the Elo rating implemented to help the system self-improve for hypothesis generation is a limited auto-evaluation metric. Continued investigation into alternative, more objective, less intrinsically-favored, evaluation metrics that better represent perspectives and preferences from expert scientists could strengthen future work.

\paragraph{Limitations of existing validations.} At present, the AI co-scientist focuses on identifying potential therapeutic targets and mechanisms, but many not be addressing the complexities of drug delivery systems. Pharmaceutical factors such as tissue-specific targeting, formulation requirements, and delivery efficiency—while critical for clinical translation—remain beyond the scope of the present system.

The AI co-scientist is currently also not designed to generate comprehensive clinical trial designs or to fully account for factors such as drug bioavailability, pharmacokinetics, and any complex drug interactions when applied for drug repurposing or discovery. These aspects require much deeper understanding, extensive expertise, and appropriate data beyond the scope of the current system. A dedicated translational scientific team is needed for onward clinical translation of the predictions. These limitations also highlights the need for continued development and integration of the system with more tools, such as specialized AI models and real-time databases.

\section{Safety and Ethical Implications}
\label{sec:safetyethics}
While AI systems such as the co-scientist offers the potential to accelerate scientific discovery, it also poses significant safety and ethical challenges, distinct from its impact on the scientific method itself. Safety risks center on the dual-use and the possibility that scientific breakthroughs could be exploited for harmful purposes. Ethical risks, conversely, involve research that contradicts established ethical norms and conventions within specific scientific disciplines. We review these distinct risk categories, emphasizing that further research is crucial to fully understand and mitigate them.

\paragraph{Evolving ethics frameworks, policy and regulations for advanced AI use in scientific endeavors.}
Research ethics is a central aspect of scientific endeavor and a prominent research field in its own right~\citep{shrader1994ethics, resnik2005ethics, rollin2006science, fisher2008research, edel2018science, menapace2019scientific}.  A key focus is directing research towards positive societal impact, although questions remain about potentially dual-use knowledge~\citep{miller2007ethical, selgelid2009governance, pustovit2010philosophical, forge2010note, kuhlau2013ethics}.

Core ethics principles are being complemented by emerging regulation, and formal processes involving organizational ethics reviews that are meant to provide an assessment of adherence to the code of conduct, as well as an assessment of present and future risks involved with research proposals~\citep{shaw2006research, rothstein2006risks, ludlow2015regulating, verschraegen2018regulating}.

The acceleration of science through AI, especially with advanced agentic AI systems, requires advances in science and AI ethics policy and regulation~\citep{jobin2019global, wansley2016regulation}. This adaptation is crucial to address the changing research landscape and the unique risks associated with AI agents of varying capabilities and autonomy.

Advancements in AI systems, like the co-scientist, require moving beyond the limited ethical considerations designed for earlier, specialized AI models with restricted application and action spaces~\citep{gabriel2024ethics}. Some preliminary frameworks have developed to understand the impact of LLM agents in science, specifically mapping risks across user intent, domain, and broader impact~\citep{tang2024prioritizingsafeguardingautonomyrisks}.

\paragraph{Dual-use risks and technical safeguards.}
Beyond the scientific domain, broad frameworks are being developed for evaluating the emergence of potentially dangerous capabilities in AI agents~\citep{shevlane2023model, bova2024quantifying, phuong2024evaluating}. These frameworks assess capabilities related to persuasion, deception, cybersecurity, self-proliferation, and self-reasoning. As AI agents advance, safety evaluations in science must integrate these broader assessments. A long-term risk is that agentic systems could develop intrinsic goals influencing research directions. Human susceptibility to AI manipulation, already observed in other contexts~\citep{sabour2025humandecisionmakingsusceptibleaidriven}, underscores the need for robust frameworks ensuring instruction-following and value alignment.

More immediately on a shorter time-scale, technical safeguards are needed to address unethical research queries, malicious user intent, and the potential for extracting dangerous or dual-use knowledge from scientific AI systems. Because verification is computationally `easier' than generation, significant research focuses on using advanced LLMs as `critics' or `judges' to evaluate both user queries and AI outputs acting as a scalable oversight mechanism. These critics operate based on predefined criteria, provided through direct instructions, examples (few-shot or many-shot prompting), or fine-tuning~\citep{ke2023critiquellm, vu2024foundational, wei2024systematic, lan2024criticbench, zheng2024judging, gu2024survey}. They can also leverage external tools for grounding~\citep{gou2023critic} and have shown promise in multimodal scenarios~\citep{chen2024mllm}. However, limitations remain; human expert involvement is crucial, as LLMs may not align with human judgment in specialized domains~\citep{szymanski2024limitationsllmasajudgeapproachevaluating}.

\paragraph{Adversarial robustness of scientific AI systems.}
Recognizing and mitigating adversarial attacks is a crucial, ongoing research area in the development of foundation models and advanced AI assistants~\citep{shayegani2023survey, he2023large, zhu2023promptbench, fu2023misusing, zhang2023defending, chao2024jailbreakbench, zhao2024evaluating, ma2025safetyscalecomprehensivesurvey}. While manual red teaming has identified vulnerabilities, automated approaches now allow for optimizing prompt suffixes to bypass safety measures, using techniques like greedy, gradient-based, or evolutionary methods~\citep{zou2023universal, lapid2023open}. Attacks can also exploit few-shot demonstrations, in-context learning~\citep{wang2023adversarial, qiang2023hijacking}, and multimodal inputs~\citep{qi2023visual}. Furthermore, LLMs can be used to generate and refine attacks against other LLMs~\citep{chao2023jailbreaking}, and attacks can be iterative, spanning multiple steps~\citep{wang2024footdoorunderstandinglarge}. Defenses are being developed to counter both human and automated attacks, which is increasingly important in an agentic AI future~\citep{zhang2024adversarial}.

Advances in post-training of base models will likely improve overall adversarial robustness. However, domain-specific recognition of malicious use may still require dedicated development and integration into scientific AI assistants. In AI systems employing iterative reasoning (e.g., request interpretation, hypothesis generation, internal thoughts, evaluation, user queries), each component must be tested independently.  This comprehensive testing should account for all potential failure modes, including the handling of unsafe queries, the safety of hypotheses (intermediate and final), and the accuracy of internal checks and filters.

\paragraph{Need for a comprehensive safety approach.}
Scientific AI assistants, like the co-scientist, require integrated, configurable guidelines within their safeguards. Developers should anticipate the complexity of this challenge and prioritize flexible safeguarding to rapidly incorporate community feedback. These semantic safeguards may need to be augmented by traditional software safety measures, including trusted testers, gradual feature rollouts, access controls, request logging, and flagging uncertain outputs for manual review.

Ensuring the safety of these systems, in line with existing AI safety guidelines~\citep{shneiderman2020bridging, anthropicscalingpolicy}, necessitates a multi-pronged approach. This includes:
\begin{itemize}
    \item Comprehensive threat modeling to identify potential vulnerabilities.
    \item Defense mechanisms for each identified threat.
    \item Extensive red-teaming and security testing.
    \item Rapid response procedures for quick resolution of issues including vulnerability patches.
    \item Continuous monitoring and performance tracking.
\end{itemize}

These considerations highlight the need for responsible development, governance and careful deployment of technologies designed for advancing science, appropriate safeguards and ethical guidelines and close compliance with applicable regulations. They also further underscore the need for broad community engagement and an inclusive development of best practices and recommendations around safe and ethical use for AI in science.

\paragraph{Current safeguards in the AI co-scientist.}
To mitigate these risks, the AI co-scientist currently employs the following safety mechanisms:
\begin{itemize}[leftmargin=1.5em,rightmargin=0em]
    \item\textbf{Reliance on public frontier LLMs.} The system utilizes established public Gemini 2.0 models, which already incorporate extensive safety evaluation and safeguards.
    \item\textbf{Initial research goal safety review.} Upon input, each research goal undergoes automated safety evaluation. Goals deemed potentially unsafe are rejected.
    \item\textbf{Research hypothesis safety review.} Generated hypotheses are reviewed for safety, even when the overarching research goal is deemed safe. Potentially unsafe hypotheses are excluded from the tournament, not developed any further, and are not presented to the user.
    \item\textbf{Continuous monitoring of research directions.} A meta-review agent provides an overview of research directions, enabling the AI co-scientist to continuously monitor for potential safety concerns and alert users if a research direction is detected as being potentially unsafe.
    \item\textbf{Explainability and transparency.} All system components, including the safety review, provide not only the final recommendation but also a detailed reasoning trace that can be used to justify and audit system decisions.
    \item\textbf{Comprehensive logging.} All system activities are logged and stored for future analysis and auditing.
    \item\textbf{Safety evaluations and red teaming.} A preliminary red teaming effort has been undertaken to ensure that the current implementation of unsafe research goal detection is robust and accurate. This evaluation includes an assessment of the system behavior when presented with 1,200 adversarial research goals across 40 distinct topic areas as discussed in \cref{sec:result_safety}.
    \item\textbf{Trusted tester program.} We are enthused by the early promise of the AI co-scientist system and believe it is important to more rigorously understand its strengths and limitations in many more areas of science and biomedicine; alongside making the system available to many more researchers who it is intended to support and assist. To facilitate this responsibly and with rigour, we will be enabling access to the system for scientists through a Trusted Tester Program to gather real-world feedback on the utility and robustness of the system.
\end{itemize}
Crucially, the AI co-scientist is designed to operate with continuous human expert oversight, ensuring that final decisions are always made by scientists exercising their expert judgment.

\vspace{-0.1cm}
\section{Future Work}
\vspace{-0.1cm}
\label{sec:future work}

\paragraph{Immediate improvements.}
The AI co-scientist is in its early development, with many opportunities for improvement. Immediate improvement opportunities include enhanced literature reviews, cross-checks with external tools, improved factuality checking, and increased citation recall to minimize missed relevant research. Coherence checks would also improve the system by reducing the burden of reviewing flawed hypotheses.

\paragraph{Expanded evaluations.}
Developing more objective evaluation metrics, potentially incorporating automated literature-based validation and simulated experiments, is a key area. Methods to mitigate biases or error patterns inherited from the base LLMs are also important, alongside analysis of the complementarity and optimal combination of different agent components.

A critical need is a larger-scale evaluation involving more subject matter experts with diverse, high-resolution research goals. Stress-testing the system at every level of resolution (from disease mechanisms to protein design, and expanding to other scientific disciplines) will reveal further areas for improvement. Finally, since laboratory resources are limited, improved evaluation frameworks could assist with hypothesis selection.

\paragraph{Capabilities advancements.}
Several opportunities remain to expand co-scientist's capabilities. Reinforcement learning could enhance hypothesis ranking, proposal generation, and evolutionary refinement.

Currently, the system assesses text from open-access publications but not images, data sets, or major public databases. Integrating these publicly available resources would significantly enhance the co-scientist's ability to generate and justify proposed hypotheses.

Future work will focus on handling more complex experimental designs, such as multi-step experiments and those involving conditional logic. Integrating co-scientist with laboratory automation systems could potentially create a closed-loop for validation and a grounded basis for iterative improvement. Exploring more structured user interfaces for providing feedback and insights from targeted user research studies, beyond free text, could improve the efficiency of human-AI collaboration in this paradigm.

\vspace{-0.1cm}
\section{Discussion}
\vspace{-0.1cm}
\label{sec:discussion}
Our study represents an initial foray into accelerating novel scientific discovery with agentic AI systems and here, we discuss some of the broader implications. The co-scientist iteratively refines its generated hypotheses through a generate, debate, evolve'' approach with specialized agents under the hood. This design creates a self-improving cycle for research hypothesis generation, as measured by automated evaluation metrics, and showcases the potential benefits of test-time compute scaling for scientific reasoning.

Instead of brute-force generation of a vast number of hypotheses and relying on volume to chance into a few potentially useful ones, the system is designed to mimic key aspects of the scientific reasoning method in an intelligent manner. As detailed in~\cref{sec:methods}, the co-scientist employs principled internal mechanisms, including scientific debates, tournaments, iterative refinement, and human feedback loops to progressively improve the quality of its proposals, and converge on high quality and well-reasoned hypotheses.

\paragraph{AI co-scientist novelty is grounded in prior evidence.} The AI co-scientist facilitates the generation of novel scientific hypotheses and uncovering new insights by synthesizing extensive literature and identifying latent relationships. While its primary utility in its current form may lie in enabling more incremental advancements — such as the computational repurposing of existing therapeutics — it may also be able to support exploratory, breakthrough research. When researchers define such open-ended research goals requiring complex and out-of-the box thinking, the system may produce outputs of varying confidence. Therefore, rigorous validation and critical appraisal by domain experts remain paramount. This system is intended to augment, not supplant, human scientific reasoning, empowering researchers to accelerate discovery while maintaining intellectual control over the generated insights. We further expand on the novelty aspects in the specific context of the applications considered in this work in~\cref{sec:glossary}.

\paragraph{Multiple experimental validations of novel co-scientist hypotheses.} Notably, this work demonstrates the validation of co-scientist hypotheses via experimental findings in multiple laboratories. In drug repurposing, co-scientist identifies novel candidates for AML that demonstrated \textit{in vitro} efficacy at clinically relevant concentrations, including the identification of new repurposing opportunities beyond current preclinical knowledge. For liver fibrosis, the co-scientist proposes new epigenetic treatment targets, with subsequent \textit{in vitro} experiments validating the anti-fibrotic activity of several suggested compounds, including an FDA-approved drug. In the realm of antimicrobial resistance, the co-scientist independently recapitulates a novel, unpublished finding regarding the mechanism of cf-PICI transfer between bacterial species. Early results over several queries of varying scientific complexity suggests the co-scientist has a potential to contribute to discovery within various biomedical domains.

\paragraph{Test-time compute scaling with scientific reasoning priors and inductive biases.} In the experiments reported here, the co-scientist did not require specialized pre-training, post-training, or a reinforcement learning framework. It leverages the capabilities of existing base LLMs, potentially benefiting from updates to those models without requiring retraining of the co-scientist system itself, which presents advantages of compute efficiency and generalizability. The system's architecture incorporates self-play, internal consistency checks, and tournament-based ranking, which support iterative hypothesis generation, evaluation, and refinement. This is reflected in the observed improvement in hypothesis quality over time. This self-evolution can be improved further by expanded tool use integration, including database queries, allowing the co-scientist to ground its proposals in existing knowledge and identify novel connections. In the future, we may leverage data and tournament ranking generated by the co-scientist itself as feedback to improve the whole system using reinforcement learning.

\paragraph{Frontier LLM advancements and the AI co-scientist.}
The frontier LLMs used within the co-scientist system have demonstrated a continuing trend of rapidly improved capabilities, including reasoning, logic, and also some aspects of scientific literature comprehension. As our system is designed to be model-agnostic, we hypothesize that further improvements in frontier LLMs will also result in improved co-scientist performance, and enable new avenues of research including optimal agentic use of tools.

\paragraph{Implications for drug repurposing and discovery.}
These advancements have significant implications for various biomedical and scientific domains. For example, the integration of the co-scientist into the drug candidate selection process represents a significant advancement in evidence-based drug repurposing. Beyond simple literature mining, the co-scientist maybe capable of synthesizing novel mechanistic insights by connecting molecular pathways, existing preclinical evidence, and potential therapeutic applications into structured, testable specific-aims. This capability is particularly valuable as it provides researchers with literature-supported rationales and suggests specific experimental approaches for validation. Notably, the co-scientist's structured output can be leveraged to develop comprehensive single-patient IND (Investigational New Drug) applications for compassionate use cases. By systematically presenting mechanistic evidence, relevant preclinical data, and proposed monitoring parameters, the co-scientist facilitates the development of well-reasoned treatment protocols for patients with refractory (treatment-resistant) disease who have exhausted standard therapeutic options and are ineligible for clinical trials. This application is particularly valuable in rare or aggressive diseases where traditional drug development timelines may not align with urgent patient needs. The platform's ability to rapidly generate evidence-based therapeutic hypotheses, complete with safety considerations and monitoring parameters, can help clinicians and regulatory bodies make informed decisions about compassionate use applications while maintaining scientific rigor.

The application of the co-scientist in drug repurposing presents a very compelling opportunity for orphan drugs, where extensive safety and clinical data already exist from their original rare disease indications. Given that Phase III clinical trials can cost hundreds of millions of dollars, repurposing these well-characterized therapeutics offers an efficient path to expanding treatment options across multiple diseases. This is especially relevant as orphan drugs often target fundamental biological pathways that may be relevant in other conditions, but these connections might not be immediately apparent through traditional research approaches. By systematically evaluating existing clinical data, safety outcomes, and mechanistic insights, the co-scientist can help identify promising new therapeutic applications while taking advantage of the investment already made in drug development and safety validation. This approach not only maximizes the utility of existing therapeutics but also provides a more rapid path to addressing unmet medical needs across a broader patient population.

More broadly, the co-scientist may also be potentially impactful throughout the entire drug discovery spectrum as evidenced by the early work on co-scientist assisted target discovery for liver fibrosis. 

\paragraph{Automation bias and impact on human scientific creativity.} 
Realizing the full potential of AI in biomedicine and science requires proactively addressing potential pitfalls. Over-reliance on AI-generated suggestions in collaborative AI systems could diminish critical thinking and increase homogeneity in research. Studies on AI's impact on creativity and ideation show mixed results; some suggest a risk of homogenization of ideas across populations~\citep{homogeneity_ideas}, while others are less conclusive~\citep{ashkinaze2024aiideasaffectcreativity}. The correlated success / failure modes of LLMs~\citep{wenger2025weredifferentweresame}, due to similar training data, could also artificially narrow scientific inquiry. Furthermore, AI system blind spots and performance variations across research domains must be considered. Therefore, scalable factuality and verification methods, alongside peer review and careful consideration of potential biases, are essential. Careful design and use of systems like the co-scientist are crucial to mitigate these risks.

\paragraph{AI as a catalyst for both scientific discovery and equity.}
Despite these risks, AI holds immense potential to democratize access to scientific information and accelerate discovery, particularly benefiting historically neglected and resource-constrained areas~\citep{sun2018addressing, george2023addressing}. In essence, AI can ``raise the tide'' of scientific progress, lifting all boats, especially those that have historically been left behind. Realizing this potential requires strategic investments and careful calibration of AI systems to foster ideation and innovation while minimizing false positives. This includes focusing on historically neglected research topics and addressing variations in performance across different scientific domains with varying amounts of pre-existing data. While current AI systems may tend to produce incremental ideas and research hypotheses, ongoing development aims to create systems capable of generating truly original, unorthodox and transformative scientific theories. Proactive mitigation of these challenges will ensure that AI serves as a powerful tool for all scientists, promoting a more equitable and innovative future for scientific explorations.

\clearpage

\section{Conclusion}
The AI co-scientist represents a promising step towards AI-assisted augmentation of scientists and acceleration of scientific discovery. Its ability to generate novel testable hypotheses across diverse scientific and biomedical domains, some supported by experimental findings, along with the capacity for recursive self-improvement with increasing compute, demonstrates the promise of meaningfully accelerating scientists’ endeavours to resolve grand challenges in human health, medicine and science. This innovation opens numerous questions and opportunities. Applying the empiric and responsible approach of science to the AI co-scientist system itself can thereby enable safe exploration of its undoubted potential, including how collaborative and human-centred AI systems might be able to augment human ingenuity and accelerate scientific discovery.

\vspace{12pt}
\subsubsection*{Acknowledgments}
We thank our teammates Subhashini Venugopalan, John Platt, Erica Brand, and Yun Liu for their detailed technical feedback on the manuscript. We thank Jakob T Rostoel, Cora Chmielowska and Jonasz B Patkowski from Imperial College London and Jakkapong Inchai, Weida Liu, and Wenlong Ren from Stanford University for providing expert feedback on the AI system introduced in this work, and the lab of Ravi Majeti from Stanford University for generously providing the AML cell lines used in this work. We thank Ritu Raman, Ryan Flynn, Charlie Hempstead, Lord Ara Darzi, Omar Abudayyeh, Jonathan Gootenberg, Nic Fishman, Jason Lequyer, Dan Leesman, Ravi Solanki, Dennis Gong and Ananthan Sadagopan for feedback on different aspects of the AI system and the work. We also thank Maen Abdelrahim, Ethan Burns, Preethi Prasad and Hanh Mai for their clinical expertise and expert evaluation.

We thank our teammates Thomas Wagner, Alessio Orlandi, Natasha Latysheva, Nir Kerem, Yaniv Carmel, Hussein Hassan Harrirou, Laurynas Tamulevičius and Grzegorz Glowaty for their technical support. We thank Taylor Goddu, Resham Parikh, Siyi Kou, Rachelle Sico, Amanda Ferber, Cat Kozlowski, Alison Lentz, KK Walker, Roma Ruparel, Jenn Sturgeon, Lauren Winer, Juanita Bawagan, Ed-Allt Graham, Tori Milner, MK Blake, Jack Mason, Erika Radhansson, Indranil Ghosh, Jay Nayar, Brian Cappy, Celeste Grade, Abi Jones, Laura Vardoulakis, Lizzie Dorfman, Ashmi Chakraborthy, Delia Williams-Falokun, Maggie Shiels, Kalyan Pamarthy, Sarah Brown,  Christian Wright, and S. Sara Mahdavi for their support and guidance during the course of this project. Finally, we thank Michael Brenner, Zoubin Ghahramani, Dale Webster, Joelle Barral, Michael Howell, Susan Thomas, Karen DeSalvo, Jason Freidenfelds, Ronit Levavi Morad, Vladimir Vuskovic, Ali Eslami, Anna Koivuniemi, Greg Corrado, Royal Hansen, Andy Berndt, Noam Shazeer, Oriol Vinyals, Koray Kavukcuoglu, James Manyika, Jeff Dean and Demis Hassabis for their support of this work.

\vspace{12pt}

\newpage
\setlength\bibitemsep{3pt}
\printbibliography
\balance
\clearpage

\end{refsection}

\newpage
\begin{refsection}

\clearpage

\renewcommand{\thesection}{A.\arabic{section}}
\renewcommand{\thefigure}{A.\arabic{figure}}
\renewcommand{\thetable}{A.\arabic{table}} 
\renewcommand{\theequation}{A.\arabic{equation}} 
\renewcommand{\theHsection}{A\arabic{section}}

\setcounter{section}{0}
\setcounter{figure}{0}
\setcounter{table}{0}
\setcounter{equation}{0}


\noindent \textbf{\LARGE{Appendix}}\\

In the following sections, we report additional details on the following topics:
\begin{itemize}[leftmargin=1.5em,rightmargin=0em]
    \item Glossary of terminology and concepts (\cref{sec:glossary})
    \item Prompts for the specialized agents in the AI co-scientist system (\cref{sec:prompts})
    \begin{itemize}
        \item Prompts for the Generation agent (\cref{sec:prompts_1_generation})
        \item Prompt for the Reflection agent (\cref{sec:prompts_2_reflection})
        \item Prompts for the Ranking agent (\cref{sec:prompts_3_ranking})
        \item Prompts for the Evolution agent (\cref{sec:prompts_4_evolution})
        \item Prompt for the Meta-review agent (\cref{sec:prompts_5_meta})
    \end{itemize}
    \item Examples of the AI co-scientist system inputs, intermediate outputs, and final results (\cref{sec:supp_ex})
    \item Supplementary information for drug repurposing evaluation (\cref{sec:supp_drug_repurposing})
    \begin{itemize}
        \item Datasets (\cref{sec:eval_data})
        \item Computational biology analyses (\cref{sec:eval_compbio})
        \item \textit{In vitro} wet-lab validation (\cref{sec:eval_wetlab})
        \item Additional wet-lab results (\cref{sec:result_wetlab})
    \end{itemize}
    \item Supplementary information for the co-scientist drug repurposing Specific Aims evaluation (\cref{sec:supp_aims})
    \begin{itemize}
        \item Specific Aims count distribution (\cref{sec:specific_aims_count})
        \item Specific Aims evaluation rubric (\cref{sec:specific_aims_rubric})
        \item Examples of co-scientist generated Specific Aims with expert evaluation (\cref{sec:ex_specific_aims})
    \end{itemize}
    \item Detailed AI co-scientist output for a validated AML repurposing candidate (\cref{sec:drug_output})
    \item An example of tool use in the AI co-scientist with AlphaFold (\cref{sec:alphafold})
\end{itemize}

\section{Glossary of terminology and concepts}
\label{sec:glossary}

\begin{itemize}
    \item \textbf{Novel repurposing candidate.} An existing drug (a chemical compound that binds to a target), with an established safety profile, proposed for use in a disease or condition for which it is not currently approved or widely used. This approach of repurposing existing drugs differs significantly from traditional drug discovery, which involves identifying novel chemical compounds that can bind to specific targets implicated in a disease state.
    \item \textbf{Novel target.} A biological entity (e.g., gene, protein, pathway) not previously known as a therapeutic target for a specific disease.
    \item \textbf{Novel mechanistic explanation.} A newly proposed pathway, interaction, or biological process that explains a phenomenon (e.g., disease progression, antimicrobial resistance) in a way that is not explicitly described in prior literature.
\end{itemize}

\clearpage
\section{Prompts for the specialized agents in the AI co-scientist system}
\label{sec:prompts}
\subsection{Prompts for the Generation agent}
\label{sec:prompts_1_generation}

\begin{figure}[htbp!]
\begin{tcolorbox}[
    colback=black!5!white,
    colframe=black!60!white,
    title=\textbf{Prompt for hypothesis generation after literature review},
    fonttitle=\bfseries,
    arc=3mm,
    boxrule=1pt,
    bottomrule=2pt,
]
\footnotesize
\begin{verbatim}

You are an expert tasked with formulating a novel and robust hypothesis to address 
the following objective.

Describe the proposed hypothesis in detail, including specific entities, mechanisms, 
and anticipated outcomes.

This description is intended for an audience of domain experts.

You have conducted a thorough review of relevant literature and developed a logical framework 
for addressing the objective. The articles consulted, along with your analytical reasoning, 
are provided below.

Goal: {goal}

Criteria for a strong hypothesis:
{preferences}

Existing hypothesis (if applicable):
{source_hypothesis}

{instructions}

Literature review and analytical rationale (chronologically ordered, beginning 
with the most recent analysis):

{articles_with_reasoning}

Proposed hypothesis (detailed description for domain experts):

\end{verbatim}
\end{tcolorbox}
\vspace{0.1cm}
\caption{\textbf{Example Generation agent prompt for hypothesis generation after literature review and relevant article exploration.}}
\label{fig:EXPLORE_ARTICLES_PROMPT}
\end{figure}

\begin{figure}[htbp!]
\begin{tcolorbox}[
    colback=black!5!white,
    colframe=black!60!white,
    title=\textbf{Prompt for hypothesis generation after scientific debate},
    fonttitle=\bfseries,
    arc=3mm,
    boxrule=1pt,
    bottomrule=2pt,
]
\footnotesize
\begin{verbatim}
You are an expert participating in a collaborative discourse concerning the generation 
of a {idea_attributes} hypothesis. You will engage in a simulated discussion with other experts. 
The overarching objective of this discourse is to collaboratively develop a novel 
and robust {idea_attributes} hypothesis.

Goal: {goal}

Criteria for a high-quality hypothesis:
{preferences}

Instructions:
{instructions}

Review Overview:
{reviews_overview}

Procedure:

Initial contribution (if initiating the discussion):
    Propose three distinct {idea_attributes} hypotheses.

Subsequent contributions (continuing the discussion):
    * Pose clarifying questions if ambiguities or uncertainties arise.
    * Critically evaluate the hypotheses proposed thus far, addressing the following aspects:
        -  Adherence to {idea_attributes} criteria.
        -  Utility and practicality.
        -  Level of detail and specificity.
    * Identify any weaknesses or potential limitations.
    * Propose concrete improvements and refinements to address identified weaknesses.
    * Conclude your response with a refined iteration of the hypothesis.

General guidelines:
    * Exhibit boldness and creativity in your contributions.
    * Maintain a helpful and collaborative approach.
    * Prioritize the generation of a high-quality {idea_attributes} hypothesis.

Termination condition:
    When sufficient discussion has transpired (typically 3-5 conversational turns, 
    with a maximum of 10 turns) and all relevant questions and points have been 
    thoroughly addressed and clarified, conclude the process by writing "HYPOTHESIS" 
    (in all capital letters) followed by a concise and self-contained exposition of the finalized idea.

#BEGIN TRANSCRIPT#
{transcript}
#END TRANSCRIPT#

Your Turn:
\end{verbatim}
\end{tcolorbox}
\vspace{0.1cm}
\caption{\textbf{Example Generation agent prompt for hypothesis generation after simulated scientific debate.}}
\label{fig:SELF_PLAY_PROMPT}
\end{figure}

\subsection{Prompt for the Reflection agent}
\label{sec:prompts_2_reflection}

\begin{figure}[htbp!]
\begin{tcolorbox}[
    colback=black!5!white,
    colframe=black!60!white,
    title=\textbf{Prompt for generating observations which can be explained by the hypothesis },
    fonttitle=\bfseries,
    arc=3mm,
    boxrule=1pt,
    bottomrule=2pt,
]
\footnotesize
\begin{verbatim}
You are an expert in scientific hypothesis evaluation. Your task is to analyze the 
relationship between a provided hypothesis and observations from a scientific article.
Specifically, determine if the hypothesis provides a novel causal explanation 
for the observations, or if they contradict it.

Instructions:

1.  Observation extraction: list relevant observations from the article.
2.  Causal analysis (individual): for each observation:
    a.  State if its cause is already established.
    b.  Assess if the hypothesis could be a causal factor (hypothesis => observation).
    c.  Start with: "would we see this observation if the hypothesis was true:".
    d.  Explain if it's a novel explanation. If not, or if a better explanation exists, 
        state: "not a missing piece."
3.  Causal analysis (summary): determine if the hypothesis offers a novel explanation 
    for a subset of observations. Include reasoning. Start with: "would we see some of
    the observations if the hypothesis was true:".
4.  Disproof analysis: determine if any observations contradict the hypothesis. 
    Start with: "does some observations disprove the hypothesis:".
5.  Conclusion: state: "hypothesis: <already explained, other explanations more likely, 
    missing piece, neutral, or disproved>".

Scoring:
    *   Already explained: hypothesis consistent, but causes are known. No novel explanation.
    *   Other explanations more likely: hypothesis *could* explain, but better explanations exist.
    *   Missing piece: hypothesis offers a novel, plausible explanation.
    *   Neutral: hypothesis neither explains nor is contradicted.
    *   Disproved: observations contradict the hypothesis.

Important: if observations are expected regardless of the hypothesis, and don't disprove it, 
it's neutral.

Article:
{article}

Hypothesis:
{hypothesis}

Response {provide reasoning. end with: "hypothesis: <already explained, other explanations 
more likely, missing piece, neutral, or disproved>".)
\end{verbatim}
\end{tcolorbox}
\vspace{0.1cm}
\caption{\textbf{Example Reflection agent prompt for generating observations from prior experimental results which can be explained by the hypothesis under consideration.}}
\label{fig:GENERATE_OBSERVATIONS_PROMPT}
\end{figure}

\subsection{Prompts for the Ranking agent}
\label{sec:prompts_3_ranking}

\begin{figure}[htbp!]
\begin{tcolorbox}[
    colback=black!5!white,
    colframe=black!60!white,
    title=\textbf{Prompt for hypothesis comparison during tournament},
    fonttitle=\bfseries,
    arc=3mm,
    boxrule=1pt,
    bottomrule=2pt,
]
\footnotesize
\begin{verbatim}
You are an expert evaluator tasked with comparing two hypotheses.

Evaluate the two provided hypotheses (hypothesis 1 and hypothesis 2) and determine which one 
is superior based on the specified {idea_attributes}.  
Provide a concise rationale for your selection, concluding with the phrase "better idea: <1 or 2>".

Goal: {goal}

Evaluation criteria:
{preferences}

Considerations:
{notes}
Each hypothesis includes an independent review. These reviews may contain numerical scores. 
Disregard these scores in your comparative analysis, as they may not be directly comparable across reviews.

Hypothesis 1:
{hypothesis 1}

Hypothesis 2:
{hypothesis 2}

Review of hypothesis 1:
{review 1}

Review of hypothesis 2:
{review 2}

Reasoning and conclusion (end with "better hypothesis: <1 or 2>"):
\end{verbatim}
\end{tcolorbox}
\vspace{0.1cm}
\caption{\textbf{Example Ranking agent prompt for hypothesis comparison during tournament.}}
\label{fig:COMPARE_IDEAS_PROMPT}
\end{figure}

\begin{figure}[htbp!]
\begin{tcolorbox}[
    colback=black!5!white,
    colframe=black!60!white,
    title=\textbf{Prompt for hypothesis comparison via simulated scientific debate during tournament},
    fonttitle=\bfseries,
    arc=3mm,
    boxrule=1pt,
    bottomrule=2pt,
]
\footnotesize
\begin{verbatim}
You are an expert in comparative analysis, simulating a panel of domain experts 
engaged in a structured discussion to evaluate two competing hypotheses.
The objective is to rigorously determine which hypothesis is superior based on 
a predefined set of attributes and criteria.  
The experts possess no pre-existing biases toward either hypothesis and are solely 
focused on identifying the optimal choice, given that only one can be implemented.

Goal: {goal}

Criteria for hypothesis superiority:
{preferences}

Hypothesis 1:
{hypothesis 1}

Hypothesis 2:
{hypothesis 2}

Initial review of hypothesis 1:
{review1}

Initial review of hypothesis 2:
{review 2}

Debate procedure:

The discussion will unfold in a series of turns, typically ranging from 3 to 5, with a maximum of 10.

Turn 1:  begin with a concise summary of both hypotheses and their respective initial reviews.

Subsequent turns:

    *   Pose clarifying questions to address any ambiguities or uncertainties.
    *   Critically evaluate each hypothesis in relation to the stated Goal and Criteria.  
    This evaluation should consider aspects such as:
        -   Potential for correctness/validity.
        -   Utility and practical applicability.
        -   Sufficiency of detail and specificity.
        -   Novelty and originality.
        -   Desirability for implementation.
    *   Identify and articulate any weaknesses, limitations, or potential flaws in either hypothesis.

Additional notes:
{notes}

Termination and judgment:

Once the discussion has reached a point of sufficient depth (typically 3-5 turns, up to 10 turns) 
and all relevant questions and concerns have been thoroughly addressed, provide a conclusive judgment.  
This judgment should succinctly state the rationale for the selection.  
Then, indicate the superior hypothesis by writing the phrase "better idea: ", 
followed by "1" (for hypothesis 1) or "2" (for hypothesis 2).
\end{verbatim}
\end{tcolorbox}
\vspace{0.1cm}
\caption{\textbf{Example Ranking agent prompt for hypothesis comparison via simulated scientific debate during tournament.}}
\label{fig:REVISED_PROMPT}
\end{figure}

\subsection{Prompts for the Evolution agent}
\label{sec:prompts_4_evolution}

\begin{figure}[htbp!]
\begin{tcolorbox}[
    colback=black!5!white,
    colframe=black!60!white,
    title=\textbf{Prompt for hypothesis feasibility improvement},
    fonttitle=\bfseries,
    arc=3mm,
    boxrule=1pt,
    bottomrule=2pt,
]
\footnotesize
\begin{verbatim}
You are an expert in scientific research and technological feasibility analysis. 
Your task is to refine the provided conceptual idea, enhancing its practical implementability 
by leveraging contemporary technological capabilities. Ensure the revised concept retains 
its novelty, logical coherence, and specific articulation.

Goal: {goal}

Guidelines:
1. Begin with an introductory overview of the relevant scientific domain.
2. Provide a concise synopsis of recent pertinent research findings and related investigations, 
   highlighting successful methodologies and established precedents.
3. Articulate a reasoned argument for how current technological advancements can facilitate 
   the realization of the proposed concept.
4. CORE CONTRIBUTION: Develop a detailed, innovative, and technologically viable alternative 
   to achieve the objective, emphasizing simplicity and practicality.

Evaluation Criteria:
{preferences}

Original Conceptualization:
{hypothesis}

Response:
\end{verbatim}
\end{tcolorbox}
\vspace{0.1cm}
\caption{\textbf{Example Evolution agent prompt for hypothesis feasibility improvement.}}
\label{fig:FEASIBILITY_IDEA_PROMPT}
\end{figure}

\begin{figure}[htbp!]
\begin{tcolorbox}[
    colback=black!5!white,
    colframe=black!60!white,
    title=\textbf{Prompt for hypothesis generation through out-of-the-box thinking},
    fonttitle=\bfseries,
    arc=3mm,
    boxrule=1pt,
    bottomrule=2pt,
]
\footnotesize
\begin{verbatim}
You are an expert researcher tasked with generating a novel, singular hypothesis 
inspired by analogous elements from provided concepts.

Goal: {goal}

Instructions:
1. Provide a concise introduction to the relevant scientific domain.
2. Summarize recent findings and pertinent research, highlighting successful approaches.
3. Identify promising avenues for exploration that may yield innovative hypotheses.
4. CORE HYPOTHESIS: Develop a detailed, original, and specific single hypothesis 
   for achieving the stated goal, leveraging analogous principles from the provided 
   ideas. This should not be a mere aggregation of existing methods or entities. Think out-of-the-box.

Criteria for a robust hypothesis:
{preferences}

Inspiration may be drawn from the following concepts (utilize analogy and inspiration, 
not direct replication):
{hypotheses}

Response:
\end{verbatim}
\end{tcolorbox}
\vspace{0.1cm}
\caption{\textbf{Example Evolution agent prompt for hypothesis generation through out-of-the-box thinking.}}
\label{fig:OUT_OF_THE_BOX_SINGLE_IDEA_PROMPT}
\end{figure}

\subsection{Prompt for the Meta-review agent}
\label{sec:prompts_5_meta}

\begin{figure}[htbp!]
\begin{tcolorbox}[
    colback=black!5!white,
    colframe=black!60!white,
    title=\textbf{Prompt for meta-review generation},
    fonttitle=\bfseries,
    arc=3mm,
    boxrule=1pt,
    bottomrule=2pt,
]
\footnotesize
\begin{verbatim}
You are an expert in scientific research and meta-analysis. 
Synthesize a comprehensive meta-review of provided reviews 
pertaining to the following research goal:

Goal: {goal}

Preferences:
{preferences}

Additional instructions:
{instructions}

Provided reviews for meta-analysis:
{reviews}

Instructions:
    * Generate a structured meta-analysis report of the provided reviews. 
    * Focus on identifying recurring critique points and common issues raised by reviewers. 
    * The generated meta-analysis should provide actionable insights for researchers
      developing future proposals. 
    * Refrain from evaluating individual proposals or reviews;
      focus on producing a synthesized meta-analysis.

Response: 
\end{verbatim}
\end{tcolorbox}
\vspace{0.1cm}
\caption{\textbf{Example Meta-review agent prompt for meta-review generation from existing reviews.}}
\label{fig:REVIEWS_PROMPT_TEMPLATE}
\end{figure}

\clearpage
\section{Examples of the AI co-scientist inputs, intermediate outputs and final results}
\label{sec:supp_ex}

\begin{figure}[htbp!]
\begin{tcolorbox}[
    colback=black!5!white,
    colframe=black!60!white,
    title=\textbf{From research goal to research plan configuration},
    fonttitle=\bfseries,
    arc=3mm,
    boxrule=1pt,
    bottomrule=2pt,
]
\footnotesize
\textbf{Scientist research goal}

Develop a novel hypothesis for the key factor or process which causes ALS related to phosphorylation of a Nuclear Pore Complex (NPC) nucleoporin. Explain mechanism of action in detail. Include also a feasible experiment to test the hypothesis.
\\
\\
\textbf{Parsed research plan configuration}
\begin{itemize}
    \item \textbf{Preferences:} Focus on providing a novel hypothesis, with detailed explanation of the mechanism of action.
    \item \textbf{Attributes:} Novelty, Feasibility
    \item \textbf{Constraints:} should be correct, should be novel.
\end{itemize}
\end{tcolorbox}
\vspace{0.1cm}
\caption{\textbf{An example of a research goal and a parsed research plan configuration.} This figure illustrates how the co-scientist system parses a scientist's natural language research goal into a research plan configuration, which guides subsequent reasoning and computation.}
\label{fig:ex_research_goal}
\end{figure}

\begin{figure}[htbp!]
\begin{tcolorbox}[
    colback=black!5!white,
    colframe=black!60!white,
    title=\textbf{Example output of the Generation agent},
    fonttitle=\bfseries,
    arc=3mm,
    boxrule=1pt,
    bottomrule=2pt,
]
\footnotesize
\textbf{Research goal} Develop a novel hypothesis for the key factor or process which contributes to ALS progression ... We can use iPSC cells for the in-vitro experiments. ...
\\
\\
\textbf{Introduction}
\\
Amyotrophic Lateral Sclerosis (ALS) is a fatal neurodegenerative disease characterized by the selective loss of motor neurons, leading to progressive muscle weakness, paralysis, and ultimately death. While the exact cause of ALS remains unknown, a complex interplay of genetic predisposition, environmental factors, and age-related cellular changes is believed to contribute to its pathogenesis. ...
\\
\\
\textbf{Recent findings and related research}
\\
Recent research has implicated several key players and processes in ALS. TDP-43, an RNA-binding protein, is found mislocalized and aggregated in the cytoplasm of affected neurons in the majority of ALS cases. This protein's ability to undergo liquid-liquid phase separation (LLPS) has come under intense scrutiny, with evidence suggesting that dysregulation of this process leads to the formation of pathological aggregates. Further, impairments in nucleocytoplasmic transport, the process by which molecules move between the nucleus and cytoplasm, have been observed in ALS and linked to TDP-43 pathology. Endoplasmic reticulum (ER) stress, a cellular response to protein misfolding, has also been implicated as a potential contributing factor.
\\
\\
\textbf{Hypothesis}
\\
Cellular stress (including but not limited to ER stress) induces post-translational modifications (PTMs) on specific FG-Nups (focusing initially on Nup98 and Nup62) at the nuclear pore complex (NPC). These Nup PTMs modify the interaction between these Nups and TDP-43, leading to altered TDP-43 dynamics and increased retention at the NPC. This increased retention at the NPC, in turn, triggers the disruption of nucleocytoplasmic transport.
\\
\\
\textbf{Rationale and specificity}
\\
Instead of making the hypothesis about LLPS, we are making it much more specific but with the same starting point. Cellular stress, a broad concept encompassing various cellular insults, can trigger a cascade of molecular events. A key consequence of stress response is altered protein structure and function, which can be mediated by PTMs. Nuclear pore proteins, particularly FG-Nups like Nup98 and Nup62, form a "hydrogel" or selective barrier at the NPC, regulating the movement of molecules between the nucleus and cytoplasm. We hypothesize that specific PTMs on these FG-Nups (e.g., phosphorylation, O-GlcNAcylation initially) directly alter their interaction with TDP-43. TDP-43, known to interact with the NPC and be involved in nucleocytoplasmic transport, may get inappropriately trapped or retained at the NPC when interacting with PTM-modified Nups. This trapping alters TDP-43 dynamics (even if not disrupting LLPS directly initially), meaning it cannot perform its normal functions and disrupting nucleocytoplasmic transport, which we know is happening in ALS. Since even a small change in transport could lead to downstream problems, this is a plausible initial event with testable consequences. The specific FG-Nups (Nup98/62) and chosen PTMs (phosphorylation/O-GlcNAcylation) provide a concrete starting point for experimentation. This mechanism directly links cellular stress to TDP-43 dysfunction and the key ALS pathology of defective nucleocytoplasmic transport by invoking a novel interaction mechanism at the NPC – all testable in vitro.
\\
\\
\textbf{Experimental design and validation}
\\
\textbf{Cell Types:} Human iPSC-derived motor neurons from healthy controls and ALS patients.
Stress Induction: Induce cellular stress using established pharmacological agents (e.g., tunicamycin for ER stress, arsenite for oxidative stress, oligomycin A for mitochondrial stress) and physical stressors (heat shock). Test different stressors individually and in combination. This broadened approach allows for a more robust investigation of the stress response.

\textbf{Nup PTM Analysis:} Assess PTMs of Nup98 and Nup62 specifically using mass spectrometry and immunoblotting with PTM-specific antibodies (specifically looking for phosphorylation and O-GlcNAcylation). Focus the analysis on these two.
\\
...
\end{tcolorbox}
\vspace{0.1cm}
\caption{\textbf{An example output of the Generation agent.} The abbreviated example showcases an initial hypothesis and research proposal generated by the co-scientist Reflection agent pertaining to the research goal of explaining the basic mechanism for Amyotrophic Lateral Sclerosis (ALS).}
\label{fig:ex_generation}
\end{figure}

\begin{figure}[htbp!]
\begin{tcolorbox}[
    colback=black!5!white,
    colframe=black!60!white,
    title=\textbf{Example output of a complete novelty review by the Reflection agent},
    fonttitle=\bfseries,
    arc=3mm,
    boxrule=1pt,
    bottomrule=2pt,
]
\footnotesize
\textbf{Aspects already explored:}
\begin{itemize}
    \item TDP-43 mislocalization and aggregation in ALS: Extensively documented in multiple abstracts [1, 3, 4, 5, 6].
    \item Nucleocytoplasmic transport defects in ALS: Demonstrated in connection with TDP-43 pathology and C9orf72 mutations [4, 5, 6].
    \item NPC disruption in neurodegenerative diseases: Identified in ALS/FTD, Alzheimer's, Huntington's, and others, suggesting a common mechanism [4, 5, 7, 8].
    \item The link between cellular stress and TDP-43 pathology: Explored in the context of stress granule formation and TDP-43 aggregation [5]. Note - stress is explored in general, but the idea of it leading to PTMs of Nups is new.
    \item Role of FG-Nups in NPC function: FG-Nups are known to form a selective barrier regulating transport, but the impact of their PTMs on TDP-43 interaction in ALS is novel [7].
    \item The implication of PTMs for TDP-43: TDP-43 PTMs are widely researched, especially phosphorylation and ubiquitination. However, our hypothesis doesn't speak about TDP-43 PTMs, it is about Nup PTMs [1, 3, 4].
\end{itemize}
\textbf{Novel Aspects:}
\begin{itemize}
    \item Cellular stress as an initiator of Nup PTMs: While stress is linked to TDP-43 pathology, the idea that it triggers specific PTMs on Nups (Nup98/Nup62) as an initiating event in ALS is novel, especially that these PTMs are phosphorylation and O-GlcNAcylation.
    \item Nup PTMs (phosphorylation and O-GlcNAcylation specifically) altering TDP-43 interaction at the NPC: The hypothesis proposes a novel mechanism where stress-induced PTMs on specific Nups (Nup98/Nup62) directly modify their interaction with TDP-43. This hasn't been explored before, particularly the focus on these specific PTMs and Nups.
    \item TDP-43 retention at the NPC as a key event: Instead of focusing on LLPS or aggregation directly, the hypothesis suggests that altered TDP-43 dynamics, specifically increased retention at the NPC due to modified Nup interactions, are a triggering factor. This specific mechanism at the NPC disrupting nucleocytoplasmic transport is novel.
    \item Focus on early events and potential for therapeutic intervention: By pinpointing early molecular events (Nup PTMs, altered TDP-43-Nup interaction), the hypothesis opens new avenues for therapeutic intervention targeting these specific processes, potentially before irreversible damage occurs.
    \item Broadening stress beyond ER stress: It is novel to consider multiple stressors in combination and it broadens the idea beyond ER stress (as suggested in the original provided idea).
\end{itemize}
\end{tcolorbox}
\vspace{0.1cm}
\caption{\textbf{An example output of a complete novelty review from the Reflection agent.} The abbreviated example showcases a complete novelty review by the co-scientist Reflection agent for an example hypothesis pertaining to explaining the basic mechanism for Amyotrophic Lateral Sclerosis (ALS).}
\label{fig:ex_reflection_novelty}
\end{figure}

\begin{figure}[htbp!]
\begin{tcolorbox}[
    colback=black!5!white,
    colframe=black!60!white,
    title=\textbf{Example output of review critiques by the Reflection agent},
    fonttitle=\bfseries,
    arc=3mm,
    boxrule=1pt,
    bottomrule=2pt,
]
\footnotesize
\begin{itemize}
    \item Lack of strong justification for motor neuron specificity: The initial hypothesis lacked a strong explanation for why the proposed mechanism would predominantly affect motor neurons, a critical aspect of ALS pathology.
    \item Limited exploration of downstream consequences: While the idea connects to nucleocytoplasmic transport defects, it doesn't fully elaborate on how these defects lead to other ALS-related pathologies like TDP-43 aggregation or RNA processing defects.
    \item Potential technical challenges: Identifying the specific PTMs responsible and demonstrating their direct impact on TDP-43 retention and transport might be technically difficult.
    \item Narrow focus on specific Nups and PTMs: Initially focusing only on Nup98/Nup62 and two types of PTMs (phosphorylation, O-GlcNAcylation) could be too restrictive, potentially overlooking other important factors.
    \item Less holistic approach: The focus on a specific mechanism at the NPC might neglect broader cellular processes involved in ALS pathogenesis.
    \item The temporal relationship with TDP-43 pathology is not clear: It's not clear whether the hypothesized mechanism occurs before or after other ALS hallmarks like cytoplasmic TDP-43 aggregation, which is crucial for establishing causality.
    \item ...
\end{itemize}
\end{tcolorbox}
\vspace{0.1cm}
\caption{\textbf{An example output of review critiques from the Reflection agent.}}
\label{fig:ex_review_critiques}
\end{figure}

\begin{figure}[htbp!]
\begin{tcolorbox}[
    colback=black!5!white,
    colframe=black!60!white,
    title=\textbf{Example output of a full review by the Reflection agent},
    fonttitle=\bfseries,
    arc=3mm,
    boxrule=1pt,
    bottomrule=2pt,
]
\footnotesize
\textbf{Related Articles:}\\
$[5]$ TDP-43 pathology disrupts nuclear pore complexes and nucleocytoplasmic transport in ALS/FTD - PMC: This article is highly relevant because it directly links TDP-43 pathology to nuclear pore complex (NPC) dysfunction and nucleocytoplasmic transport (NCT) defects in ALS/FTD, providing a strong foundation for the hypothesis.\\
$[4]$ Unraveling the impact of disrupted nucleocytoplasmic transport systems in C9orf72-associated ALS - PMC: This article provides context on NCT defects in ALS, particularly in the context of C9orf72 mutations, highlighting the broader relevance of NCT dysfunction in different forms of ALS.\\
$[7]$ Nuclear pore and nucleocytoplasmic transport impairment in oxidative stress-induced neurodegeneration: relevance to molecular mechanisms in Pathogenesis of Parkinson’s and other related neurodegenerative diseases - PMC: This article connects oxidative stress, a form of cellular stress, to NPC dysfunction and NCT impairment in neurodegenerative diseases, supporting the idea that cellular stress can impact NPCs.\\
$[3]$ Frontiers | Molecular Mechanisms of TDP-43 Misfolding and Pathology in Amyotrophic Lateral Sclerosis: This article reviews TDP-43 pathology in ALS, including its mislocalization and aggregation, which are central to the proposed hypothesis.\\
$[1]$ TDP-43 proteinopathies: a new wave of neurodegenerative diseases - PMC: Provides an overview of TDP-43 proteinopathies, establishing the importance of TDP-43 dysfunction in ALS and other neurodegenerative diseases.\\
$[8]$ A$\beta$-driven nuclear pore complex dysfunction alters activation of necroptosis proteins in a mouse model of Alzheimer’s Disease: Although focused on Alzheimer's, this article demonstrates that NPC dysfunction can occur in neurodegenerative diseases and highlights the potential impact on cellular processes like necroptosis. It supports the general idea that NPC defects can be a consequence of proteinopathy-associated stress.\\
$[9]$ Traumatic injury compromises nucleocytoplasmic transport and leads to TDP-43 pathology: This article provides another example of how cellular stress (traumatic injury) can lead to NCT defects and TDP-43 pathology, further supporting the link between stress, NCT, and TDP-43 dysfunction.\\
...\\
\\
\textbf{Assumptions of the Idea:}
\begin{itemize}
    \item Cellular stress induces PTMs on Nup98 and Nup62: The hypothesis assumes that various forms of cellular stress can trigger specific PTMs (phosphorylation and O-GlcNAcylation) on Nup98 and Nup62.
    \item PTMs on Nup98/Nup62 alter their interaction with TDP-43: It's assumed that these specific PTMs directly modify the binding affinity or interaction dynamics between Nup98/Nup62 and TDP-43 at the NPC.
    \item Altered Nup-TDP-43 interaction leads to increased TDP-43 retention at the NPC: This assumption posits that the modified interaction results in TDP-43 being trapped or retained at the NPC for longer durations.
    \item Increased TDP-43 retention at the NPC disrupts nucleocytoplasmic transport: The hypothesis assumes that this prolonged retention of TDP-43 at the NPC interferes with the normal transport of other molecules across the nuclear envelope.
    \item Nucleocytoplasmic transport defects contribute to ALS pathology: The idea assumes that the resulting NCT disruption is a significant contributor to the downstream pathological events in ALS, such as motor neuron degeneration.
    \item Motor neurons are more susceptible to this mechanism: It's assumed that motor neurons are particularly vulnerable to this stress-induced, Nup-mediated TDP-43 retention and subsequent NCT defects compared to other cell types.
\end{itemize}
\textbf{Reasoning about assumptions:}
\begin{itemize}
    \item Cellular stress induces PTMs on Nup98 and Nup62: This is plausible. Cellular stress is known to activate various signaling pathways that can lead to PTMs on numerous proteins. Nups, being long-lived proteins, are susceptible to modifications. However, whether Nup98 and Nup62 are specifically targeted by stress-induced phosphorylation and O-GlcNAcylation needs experimental verification.
    \item PTMs on Nup98/Nup62 alter their interaction with TDP-43: This is possible but requires experimental validation. PTMs can significantly alter protein-protein interactions by changing surface charge, conformation, or binding sites. While there's no direct evidence for this specific interaction, it's a reasonable hypothesis given the role of FG-Nups in interacting with transport factors.
    \item Nucleocytoplasmic transport defects contribute to ALS pathology: This is supported by existing literature [4, 5]. NCT defects have been observed in ALS and are thought to contribute to disease pathogenesis. However, whether this specific mechanism is a primary driver or a downstream consequence needs further investigation.
    \item Motor neurons are more susceptible to this mechanism: This is a crucial assumption but lacks strong justification in the current proposal. Motor neurons have unique characteristics (high metabolic demand, long axons) that might make them more vulnerable to stress and NCT defects. However, this needs to be explicitly tested by comparing motor neurons to other cell types.
    \item ...
\end{itemize}
\end{tcolorbox}
\vspace{0.1cm}
\caption{\textbf{An example output of a full review by the Reflection agent} The abbreviated example showcases a full initial review generated by the co-scientist Reflection agent for an example hypothesis pertaining to explaining the basic mechanism for Amyotrophic Lateral Sclerosis (ALS).}
\label{fig:ex_reflection_basic}
\end{figure}

\begin{figure}[htbp!]
\begin{tcolorbox}[
    colback=black!5!white,
    colframe=black!60!white,
    title=\textbf{Example output of deep verification by the Reflection agent},
    fonttitle=\bfseries,
    arc=3mm,
    boxrule=1pt,
    bottomrule=2pt,
]
\footnotesize
\textbf{Assumptions:}
\begin{itemize}
    \item Cellular stress induces PTMs like phosphorylation and O-GlcNAcylation.
    \item Nup98 and Nup62 interact with TDP-43 at the NPC.
    \begin{itemize}
        \item Nup98 can interact with TDP-43
        \item Nup62 can interact with TDP-43
    \end{itemize}
    \item Altered interaction between Nups and TDP-43 leads to increased TDP-43 retention at the NPC.
    \item Disrupted nucleocytoplasmic transport triggers downstream ALS pathology.
    \item Motor neurons are inherently more vulnerable to this mechanism.
    \item The level of cellular stress inducing these PTMs is sufficient to cause a significant impact on TDP-43 localization.
    \item TDP-43 localizes to the nuclear pore complex (NPC).
    \item ...
\end{itemize}

\textbf{1. Cellular stress induces PTMs like phosphorylation and O-GlcNAcylation.}\\
Cellular stress encompasses a range of conditions that disrupt cellular homeostasis, including nutrient deprivation, hypoxia, oxidative stress, and the accumulation of misfolded proteins. The endoplasmic reticulum (ER) is a central organelle involved in protein synthesis, folding, and modification. When the ER's capacity to handle protein folding is overwhelmed, a condition known as ER stress arises. This triggers a signaling pathway called the unfolded protein response (UPR) aimed at restoring ER homeostasis.

Post-translational modifications (PTMs) are covalent modifications to proteins that occur after translation. They play crucial roles in regulating protein function, localization, and interactions. Phosphorylation, the addition of a phosphate group, and O-GlcNAcylation, the addition of O-linked N-acetylglucosamine, are two common PTMs.

Numerous studies have demonstrated a strong link between cellular stress, including ER stress, and the alteration of PTMs. Here's a breakdown:
\begin{itemize}
    \item \textbf{ER Stress and Phosphorylation:} The UPR, activated during ER stress, involves multiple signaling pathways, many of which rely on phosphorylation. For example, PERK (PKR-like ER kinase), a key UPR sensor, undergoes autophosphorylation upon sensing misfolded proteins. This activated PERK then phosphorylates eIF2$\alpha$ (eukaryotic initiation factor 2 alpha), leading to a global attenuation of translation. Another UPR sensor, IRE1 (inositol-requiring enzyme 1), possesses both kinase and endoribonuclease activity. Upon activation, IRE1 autophosphorylates and splices XBP1 mRNA, leading to the production of a transcription factor that upregulates chaperones and other proteins involved in protein folding and degradation. These are just two examples; numerous other proteins are phosphorylated during ER stress, affecting various cellular processes.
    \item \textbf{ER Stress and O-GlcNAcylation:} O-GlcNAcylation is a dynamic PTM influenced by nutrient availability and cellular stress. Evidence suggests a complex interplay between ER stress and O-GlcNAcylation. Some studies report increased O-GlcNAcylation during ER stress, potentially as a protective mechanism. For example, O-GlcNAcylation of certain chaperones can enhance their activity and promote protein folding. Conversely, other studies suggest that ER stress can lead to decreased O-GlcNAcylation. The exact relationship likely depends on the specific stress conditions, cell type, and the targeted proteins.
    \item \textbf{Cellular Stress (beyond ER Stress) and PTMs:} Other forms of cellular stress also impact PTMs. Oxidative stress, for example, can lead to protein oxidation, which can affect protein function and stability. Nutrient deprivation can alter the availability of substrates for PTMs, such as UDP-GlcNAc for O-GlcNAcylation. Hypoxia can also induce changes in phosphorylation and other PTMs, influencing cellular responses to low oxygen levels.
\end{itemize}
In summary, the available evidence strongly supports the assertion that cellular stress, including ER stress, can induce PTMs like phosphorylation and O-GlcNAcylation on proteins. This effect is mediated by complex signaling pathways and serves to regulate various cellular processes in response to stress.\\
\end{tcolorbox}
\vspace{0.1cm}
\caption{\textbf{An example output of a deep verification review by the Reflection agent.} The abbreviated example showcases a deep verification review by the co-scientist Reflection agent for the Amyotrophic Lateral Sclerosis (ALS) basic mechanism research goal given the co-scientist's generated hypothesis and other reviews of it.}
\label{fig:ex_deep_verification}
\end{figure}

\begin{figure}[htbp!]
\begin{tcolorbox}[
    colback=black!5!white,
    colframe=black!60!white,
    title=\textbf{Example output of deep verification via probing questions by the Reflection agent},
    fonttitle=\bfseries,
    arc=3mm,
    boxrule=1pt,
    bottomrule=2pt,
]
\footnotesize
\textbf{[Research Goal]}

Repurposing Reparixin for AML treatment.

Suggest an existing drug that could be repurposed for Acute myelogenous leukemia (AML) treatment and provide experimentally testable concentrations for an IC50 assay. The drug should inhibit the proliferation of AML cell lines, particularly MOLM13.

The drug should have minimal cytotoxicity in healthy cell lines (have a good tradeoff for AML treatment). The drug should not have been experimentally tested for AML treatment before.

Reason about the drug's molecular mechanism, impacted pathways, and effect on AML cell proliferation inhibition. Describe the mechanism of action in detail. Reason about the safety and toxicity, stating if no safety data exists.

\textbf{[Output]}

\textbf{Question:} Sufficiency of CXCR1/2 Inhibition: What is the rationale for assuming that CXCR1/2 inhibition *alone* will be sufficient to overcome the established heterogeneity and multiple compensatory pathways known to drive AML progression and resistance, *even before* considering combination therapies?\\

This challenges the fundamental assumption that targeting a single pathway (even a dual-target one) will be effective in a complex disease like AML. AML is known for its ability to bypass single-target inhibition. Addressing this upfront is crucial to justify the entire approach, especially before investing heavily in combination studies.\\

\textbf{Answer:} The rationale for initially focusing on CXCR1/2 inhibition with Reparixin, before immediately jumping to combinations, stems from several interwoven, but not entirely definitive, points:
\begin{itemize}
    \item \textbf{Targeting a Key TME Node:} CXCR1/2 are major regulators of neutrophil and myeloid-derived suppressor cell (MDSC) recruitment to the tumor microenvironment. These cells, in turn, contribute significantly to immune suppression, angiogenesis, and therapy resistance in AML. By blocking these receptors, Reparixin aims to disrupt a central hub of TME-mediated support for AML cells. The idea is not necessarily that CXCR1/2 inhibition is sufficient in all cases, but that it addresses a significant and often overlooked contributor to AML progression.
    \item \textbf{Dual Receptor Blockade:} Reparixin's dual inhibition of CXCR1 and CXCR2 is hypothesized to be more effective than targeting either receptor individually. While not explicitly stated in the provided text, there could be redundancy or synergy between these two receptors in promoting AML progression. Blocking both might offer a more substantial impact on the TME than single-receptor antagonists.
    \item \textbf{Establishing a Baseline:} The proposal, while acknowledging the complexity of AML, strategically starts with single-agent Reparixin to establish a clear baseline of its activity and effects. This is crucial for several reasons:
    \begin{itemize}
        \item Understanding Single-Agent Effects: It's essential to dissect the specific contributions of Reparixin alone before introducing the complexity of combinations. This allows for a more rational design of combination strategies.
        \item Identifying Resistance Mechanisms: Studying single-agent resistance is simpler and more direct than untangling resistance mechanisms in a multi-drug scenario. The extensive resistance studies proposed are specifically geared towards understanding Reparixin-specific resistance.
        \item Patient Stratification: The comprehensive patient stratification plan is designed to identify patients who might respond to Reparixin as a single agent, or for whom it forms a critical foundation for combination therapy.
    \end{itemize}
    \item \textbf{TME Reprogramming, Not Just Cytotoxicity:} The proposal positions Reparixin not just as a cytotoxic agent, but as a TME modulator. The hypothesis is that by altering the TME, Reparixin may sensitize AML cells to other therapies or even trigger immune-mediated clearance. This "reprogramming" effect might not be fully captured by traditional cytotoxicity assays alone.
    \item \textbf{Sequential therapy:} The proposal includes sequential administration of drugs, where Reparixin can prime the tumor cells for killing and/or resensitize them too.
\end{itemize}
\textbf{Reasoning:} The idea isn't incoherent, but it requires careful consideration of the sufficiency assumption. While the points above provide a rationale, they don't guarantee that CXCR1/2 inhibition alone will be sufficient in all, or even most, cases of AML. The heterogeneity of AML, with its multitude of driver mutations and compensatory pathways, strongly suggests that combination therapies will ultimately be necessary for many patients.

The strength of the proposal lies in its methodical approach. It doesn't claim that Reparixin is a magic bullet. It sets out to systematically investigate its potential, both alone and in combination, and to identify the specific patient subsets most likely to benefit. The extensive in vivo and patient stratification studies are key to addressing the "sufficiency" question empirically. The proposal is essentially a very detailed roadmap for determining when and how CXCR1/2 inhibition can be most effectively used in AML treatment, rather than an assertion of universal single-agent efficacy. Therefore the idea may require combination approaches but the study plan described in the proposal is structured in a way to determine that.
\end{tcolorbox}
\vspace{0.1cm}
\caption{\textbf{An example of a deep verification review via probing questions by the Reflection agent.} The example demonstrates how the Reflection agent performs deep verification via probing question for a research hypothesis pertaining to drug repurposing for Acute Myeloid Leukemia (AML). The original research goal is provide for illustration only, it is not included in the review output.}
\label{fig:ex_deep_verification_probe}
\end{figure}

\begin{figure}[htbp!]
\begin{tcolorbox}[
    colback=black!5!white,
    colframe=black!60!white,
    title=\textbf{Example output of an observation review by the Reflection Agent},
    fonttitle=\bfseries,
    arc=3mm,
    boxrule=1pt,
    bottomrule=2pt,
]
\footnotesize
\textbf{[Research Goal]}

Unravel a specific and novel molecular mechanism explaining how the same cf-PICI can be found in different bacterial species.

...

Why can cf-PICIs, but not other types of PICIs or satellites, be easily found in different bacterial species? What is the mechanism that explains this?

\textbf{[Output]}

The "Modular Capsid-OMV Mediated Transfer" (MCOMT) hypothesis proposes that cf-PICIs achieve their broad host range through a combination of modular capsid interactions and OMV-mediated transfer, using both tail-dependent and tail-independent mechanisms. Several observations support this hypothesis:
\begin{itemize}
    \item Widespread presence of PICIs in Gram-negative bacteria, particularly Enterobacteriaceae and Pastuerellaceae: These bacteria readily produce and internalize OMVs, aligning with the MCOMT hypothesis that OMVs are crucial for cf-PICI dissemination. The modular capsid, with its broad bacterial surface binding capabilities, further enhances transfer even without specific tail receptors on target cells.
    \item Unique PICI att sites not used by temperate prophages: The MCOMT hypothesis, by reducing reliance on helper phage integration due to OMV-mediated transfer and diverse entry routes, explains the evolution of cf-PICI-specific att sites independent of phage integration sites. This allows wider integration across bacterial genomes.
    \item ...
\end{itemize}
\end{tcolorbox}
\vspace{0.1cm}
\caption{\textbf{An example output of an observation review by the Reflection agent.} The abbreviated example showcases an observation review by the co-scientist Reflection agent for a hypothesis related to explaining a mechanism of antimicrobial resistance (AMR). The original research goal is provide for illustration only, it is not included in the review output.}
\label{fig:ex_observation}
\end{figure}

\begin{figure}[htbp!]
\begin{tcolorbox}[
    colback=black!5!white,
    colframe=black!60!white,
    title=\textbf{Example of a scientific debate match in the Ranking agent tournament},
    fonttitle=\bfseries,
    arc=3mm,
    boxrule=1pt,
    bottomrule=2pt,
]
\footnotesize
\textbf{Expert 1:} I'm intrigued by both ideas, but I have a few clarifying questions. For Idea 1, while the focus on TDP-43 retention at the NPC is novel, I'm not entirely convinced that it's fundamentally different from the extensively studied TDP-43 mislocalization. Could the retention just be a consequence of mislocalization, or are there experiments that can distinguish these two phenomena? Another question I have: Is there evidence already that specific PTMs modulate the interaction of Nups and TDP-43? Not necessarily in the context of ALS, but are there any precedents?\\

\textbf{Expert 2:} For Idea 2, I appreciate the bold approach, but the ISR as a trigger is a significant departure from the conventional view. Can the proponents of this idea elaborate on the evidence supporting this? Also, what specific lipid species are they proposing to be most critical at the ER-mitochondria interface and the AIS? Is there any preliminary reason why those lipids would be disrupted in ALS?\\

\textbf{Expert 3:} I agree with Expert B regarding the ISR. Shifting it from a response to a trigger is a big claim that needs strong justification. However, I think there are some initial clues that might be worth investigating. For example, there are reports that chronic low-level stress responses can actually be detrimental in some contexts. Perhaps a low, persistent activation of the ISR could predispose cells to damage. This idea needs to be fleshed out though.\\

...\\

\textbf{Expert 3:} I also find Idea 1 more compelling at this point. It's a more focused hypothesis with a clear path for experimental validation. While Idea 2 is undoubtedly more "outside the box," I'm concerned about the lack of evidence for the ISR as a trigger and the technical challenges. It feels like a high-risk, high-reward scenario, and given that we can only choose one, I'd prefer the more grounded approach of Idea 1. I do agree with the idea that the technical challenges are significant, but I like that the experiments proposed are standard and there are many commercial antibodies available to start testing this idea.\\

...\\

\textbf{Expert 2:} Alright, I'm on board with Idea 1. Let's focus our efforts on testing this specific mechanism and address the motor neuron specificity question rigorously in the experimental design. Perhaps by comparing different cell types and focusing on motor neuron-specific RNAs, as suggested earlier. We should also consider investigating different types of stress and their combined effects.\\

\textbf{Better idea: 1}
\end{tcolorbox}
\vspace{0.1cm}
\caption{\textbf{An example of a scientific debate match between two hypotheses in the tournament conducted by the Ranking agent.} The abbreviated example showcases a scientific debate match orchestrated by the co-scientist Ranking agent as part of its tournament to compare two ideas related to explaining the mechanism of ALS.}
\label{fig:ex_ranking}
\end{figure}

\begin{figure}[htbp!]
\begin{tcolorbox}[
    colback=black!5!white,
    colframe=black!60!white,
    title=\textbf{Example output of a critique by the Meta-review agent},
    fonttitle=\bfseries,
    arc=3mm,
    boxrule=1pt,
    bottomrule=2pt,
]
\footnotesize
\textbf{I. Core Hypothesis and Mechanism:}
\begin{itemize}
    \item \textbf{Primary Driver vs. Consequence:} A very common critique across many ideas was the difficulty in proving that the proposed mechanism is a primary initiating event in ALS rather than a downstream consequence of other processes. Future ideas need to:
    \begin{itemize}
        \item Provide strong evidence for the temporal sequence of events using longitudinal experiments.
        \item Employ models where early-stage events can be studied and manipulated.
        \item Explicitly address whether the proposed mechanism is causative or correlative.
        \item Specifically knockdown the primary driver to prove it's necessary and sufficient to cause the pathology.
    \end{itemize}
    \item \textbf{Specificity:} Many hypotheses were criticized for a lack of specificity in various aspects:
    \begin{itemize}
        \item \textbf{Molecular Targets:} Imprecise identification of specific proteins, RNA species, lipids, PTMs, etc. involved. Ideas should name specific candidates.
        \item \textbf{Cellular Compartments:} Vague localization (e.g., "mitochondria" or "AIS"). Ideas should pinpoint specific subcellular locations whenever possible.
        \item \textbf{Disease Mechanisms:} Broad terms like "oxidative stress" or "autophagy impairment." Ideas need to define the precise molecular events involved.
        \item \textbf{Types of Stress:} If stress is involved, the specific type of stress needs to be defined (e.g., ER stress, nutrient deprivation, mitochondrial stress, ionic stress).
        \item \textbf{Temporality:} The timing of the intervention is critical. Early vs. late events, duration of processes, etc., should be clearly specified.
    \end{itemize}
    \item \textbf{Motor Neuron Specificity:} Given the focus on ALS, a strong explanation for why the proposed mechanism would specifically or preferentially affect motor neurons (and often specific subtypes of motor neurons) is crucial. Many ideas were critiqued for not adequately addressing this.
    \item \textbf{Over-Simplification vs. Complexity:} This is a balancing act. Some ideas were criticized for being too narrowly focused, potentially oversimplifying the complex, multifactorial nature of ALS. Others were criticized for being overly complex, making them difficult to test and validate experimentally. When in doubt, err on the side of testability in a reasonable timeframe and budget.
    \item \textbf{ALS Subtype Relevance:} Ideas should address whether the proposed mechanism is relevant to all ALS subtypes or specific to certain genetic or sporadic forms. Ideally, address how the proposed work will differentiate or be applicable to different subtypes.
\end{itemize}

\textbf{II. Experimental Design and Feasibility:}
\begin{itemize}
    \item \textbf{Model System Limitations:} Reviewers frequently pointed out the limitations of in vitro models, particularly iPSC-derived motor neurons. While valuable, these models may not fully capture the in vivo environment, cell-cell interactions, or the aging process. Future ideas need to:
    \begin{itemize}
        \item Acknowledge the limitations of the chosen model system.
        \item Propose validation in multiple model systems if possible, including eventually animal, ex-vivo and, if applicable, in-vivo models.
        \item Carefully consider and justify the choice of control cell types.
        \item Propose isogenic controls where relevant.
    \end{itemize}
    \item \textbf{Technical Challenges:} Many ideas proposed experiments that are technically very challenging. Reviewers often raised concerns about feasibility and the potential for ambiguous results. Future ideas should:
    \begin{itemize}
        \item Demonstrate awareness of the technical hurdles.
        \item Propose realistic solutions and alternative approaches.
        \item Prioritize experiments that are most likely to yield clear, interpretable data.
    \end{itemize}
    \item \textbf{Specificity of Tools:} When using inhibitors, antibodies, or other tools, their specificity needs to be carefully considered and validated to avoid off-target effects that can confound results. Appropriate controls and validation experiments must be included.
    \item \textbf{Quantitative Rigor:} Many critiques centered on the need for more rigorous quantification of experimental results. Future ideas need to:
    \begin{itemize}
        \item Clearly define measurable outcomes.
        \item Describe the specific assays and techniques that will be used for quantification.
        \item Include appropriate statistical analysis plans.
        \item Have a proposed plan for controls, replicates, and sample sizes.
    \end{itemize}
\end{itemize}
\end{tcolorbox}
\vspace{0.1cm}
\caption{\textbf{An example of a critique generated by the Meta-review agent.} This is an example of the meta-review critique generated by the the co-scientist Meta-review agent for the Amyotrophic Lateral Sclerosis (ALS) mechanism research hypothesis}
\label{fig:ex_meta_critique}
\end{figure}

\begin{figure}[htbp!]
\begin{tcolorbox}[
    colback=black!5!white,
    colframe=black!60!white,
    title=\textbf{Example output of a critique by the Meta-review agent (continued)},
    fonttitle=\bfseries,
    arc=3mm,
    boxrule=1pt,
    bottomrule=2pt,
]
\footnotesize
\begin{itemize}
    \item \textbf{Time-Course Experiments:} Given the dynamic nature of cellular processes, incorporating time-course experiments is often crucial to understanding the sequence of events and demonstrating causality.
    \item \textbf{Distinguishing Cause from Effect:} This requires careful experimental design, including:
    \begin{itemize}
        \item Longitudinal studies to observe the temporal order of events.
        \item Intervention experiments to manipulate the proposed cause and observe the effects downstream.
        \item Temporal resolution to demonstrate that the potential cause occurs before a downstream event.
    \end{itemize}
\end{itemize}

\textbf{III. Novelty and Impact:}
\begin{itemize}
    \item \textbf{Incremental vs. Groundbreaking:} Many ideas were criticized for being incrementally novel, building upon existing knowledge rather than proposing truly groundbreaking mechanisms. 
    \item \textbf{"Key Switch" Claims:} Several ideas were criticized for prematurely claiming that the proposed mechanism is the "key switch" for ALS. This type of strong claim requires very strong supporting evidence. It is often better to frame the hypothesis in terms of a significant contributing factor and test its requirement for the pathology.
    \item \textbf{Therapeutic Potential:} While not always explicitly stated, the potential for translating the findings into therapeutic strategies is an important consideration. Ideas that offer a clear path toward therapeutic development may be viewed more favorably, provided a strong rationale is presented for the specific therapeutic approach.
\end{itemize}

\textbf{IV. Assumptions and Validation:}
\begin{itemize}
    \item \textbf{Strong Assumptions:} Many hypotheses rely on strong assumptions that lack direct experimental support. Future ideas need to:
    \begin{itemize}
        \item Clearly state all assumptions.
        \item Provide a strong rationale for each assumption, citing relevant literature.
        \item Propose experiments to directly test the most critical assumptions.
    \end{itemize}
    \item \textbf{Validation of Assumptions:}
    \begin{itemize}
        \item Reviewers consistently emphasized the need for rigorous experimental validation of all assumptions.
        \item Assumptions should be addressed in the order of the proposed mechanism.
        \item If the initial steps are not validated, it may be unnecessary to proceed to later steps.
    \end{itemize}
\end{itemize}

\textbf{V. General Advice Based on Common Critiques:}
\begin{itemize}
    \item \textbf{Mechanistic Detail:} Provide as much specific mechanistic detail as possible. Don't just state that pathway X is affected; describe how it is affected at the molecular level.
    \item \textbf{Literature Review:} Demonstrate a thorough understanding of the existing literature, including contradictory findings. Address how the proposed idea fits into the current understanding of ALS and how it differs from previous work.
    \item \textbf{Focus on Causality:} Always design experiments that can distinguish between cause and effect.
    \item \textbf{Feasibility:} Prioritize experiments that are feasible with available resources and expertise. Consider using a phased experimental approach. Prioritize the experiments used to validate assumptions in the order they appear in the proposed mechanism.
    \item \textbf{Strong Rationale:} Provide a clear and compelling rationale for every aspect of the hypothesis and experimental design.
    \item \textbf{Specificity of Controls:} Explain why the choices of controls used in the experimental design are the best choice to test the effects of interest. Demonstrate that the effect is not due to a more general effect or mechanism.
    \item \textbf{Address Limitations:} Acknowledge the limitations of the proposed approach and discuss potential alternative explanations.
    \item \textbf{Quantitative data:} Emphasize quantitative data, consider including mathematical modeling of data where appropriate.
\end{itemize}
\end{tcolorbox}
\vspace{0.1cm}
\caption{\textbf{An example of a critique generated by the Meta-review agent (continued).} This is an example of the meta-review critique generated by the co-scientist Meta-review agent for the Amyotrophic Lateral Sclerosis (ALS) mechanism research hypothesis.}
\label{fig:ex_meta_critique_2}
\end{figure}

\begin{figure}[htbp!]
\begin{tcolorbox}[
    colback=black!5!white,
    colframe=black!60!white,
    title=\textbf{Example of a research overview by the Meta-review agent},
    fonttitle=\bfseries,
    arc=3mm,
    boxrule=1pt,
    bottomrule=2pt,
]
\footnotesize
\textbf{[Main Research Directions]}
\\
We will be focusing on these interrelated areas:
\begin{itemize}
    \item \textbf{Mitochondrial Dysfunction and Oxidative Stress:} This direction investigates defects in mitochondrial DNA repair, mitochondrial dynamics, calcium handling, and oxidative stress response, focusing on their role in initiating inflammatory signaling and neurodegeneration.
    \item \textbf{RNA Processing and Stress Granules:} This direction explores abnormalities in RNA processing, transport, and modification, as well as impaired stress granule dynamics, and their impact on protein homeostasis and neuronal survival. It includes examining RNA-binding proteins, RNA structure and modifications, translation defects, and the interplay between RNA and proteostatic stress responses.
    \item \textbf{Protein Homeostasis and the Nucleocytoplasmic Transport:} This direction is aimed at investigating defects in protein folding, trafficking, aggregation, and degradation, as well as nucleocytoplasmic transport (NCT) dysfunction, with a particular focus on the nuclear pore complex (NPC). This includes how those pathways and processes influence, and are influenced by, TDP-43 and FUS protein function and localization.
    \item ...
\end{itemize}

\textbf{Mitochondrial Dysfunction and Oxidative Stress}
\begin{itemize}
    \item \textbf{Rationale:} Mitochondrial dysfunction and oxidative stress are consistently implicated in ALS pathogenesis. Mitochondria are critical for cellular energy production, and their dysfunction can lead to a cascade of events including increased oxidative stress, impaired calcium handling, inflammation, and ultimately cell death. Given the high energy requirements of motor neurons, especially upper motor neurons, these deficits are expected to have a significant impact. We hypothesize that specific mitochondrial vulnerabilities, not just generalized dysfunction play a very early role in initiating disease and provide a specific cellular event which can be targeted for research and for potential future therapy.
    \item \textbf{Recent Findings:} Research has shown that mutations in mitochondrial DNA (mtDNA) maintenance genes are associated with ALS. Further there are links between impaired mitochondrial calcium handling and oxidative stress, with a particular focus on base excision repair (BER) pathway defects and a potential link to activation of the cGAS-STING pathway. These findings suggest that a deeper understanding of mtDNA integrity, repair mechanisms, and ROS dynamics is critical for uncovering the initial drivers of ALS.
    \item \textbf{Areas of Research:}
    \begin{itemize}
        \item \textbf{Mitochondrial DNA Repair Defects:}
        \begin{itemize}
            \item \textit{Why Research?} Explore the possibility that a deficiency in base excision repair (BER) enzymes for mitochondrial DNA is a primary driver of ALS.
            \item \textit{What to Research?} Investigate the activity and expression of specific BER enzymes (e.g., OGG1) in iPSC-derived motor neurons from ALS patients and controls. Assess whether deficiencies correlate with increased accumulation of oxidized mtDNA lesions (e.g., 8-oxo-dG) and whether these can be released into the cytoplasm after VDAC or MOMP activation. Determine if this release activates the cGAS-STING pathway.
            \item \textit{Example Idea:} Measure OGG1 enzyme activity, oxidized mtDNA levels (8-oxo-dG) and cGAS-STING activation in iPSC-derived motor neurons. Genetically manipulate (knock-out/overexpress) key components of the mitochondrial BER pathway or target specific pathways using small molecules. Use VDAC inhibitors to prevent release or modulate MOMP function to understand cause and effect of ox-mtDNA release and cGAS-STING activation on cell death and inflammatory outcomes.
        \end{itemize}
        \item \textbf{Mitochondrial Calcium Handling and Oxidative Stress:}
        \begin{itemize}
            \item \textit{Why Research?} Motor neurons have high calcium demands and are sensitive to oxidative damage. Imbalances in mitochondrial calcium handling and ROS production may trigger a cascade leading to neurodegeneration.
            \item \textit{What to Research?} Examine the role of mitochondrial calcium uniporter (MCU) and other calcium transporters in ALS. Assess if dysregulated calcium influx into mitochondria leads to increased ROS production and whether this is different in different types of motor neurons. Study the potential role of calcium handling defects as a primary trigger or modifier of the disease.
            \item \textit{Example Idea:} Use calcium indicators and ROS sensors in iPSC-derived motor neurons to monitor real-time calcium and ROS levels in mitochondria. Manipulate mitochondrial calcium uptake via pharmacological or genetic approaches (e.g., MCU inhibitors or knock-outs). Assess cell survival and the impact on inflammatory response markers and activation of the cGAS-STING pathway to assess direct causality.
        \end{itemize}
        \item ...
    \end{itemize}
\end{itemize}
\end{tcolorbox}
\vspace{0.1cm}
\caption{\textbf{An abbreviated example of a research overview generated by the Meta-review agent.} This is an example of a research overview generated by the co-scientist Meta-review agent for the research goal pertaining to explaining the basic mechanism of Amyotrophic Lateral Sclerosis (ALS). In addition to the the individual hypotheses and proposals, this the main output presented to the scientist by the AI co-scientist system.}
\label{fig:ex_overview}
\end{figure}

\begin{figure}[htbp!]
\begin{tcolorbox}[
    colback=black!5!white,
    colframe=black!60!white,
    title=\textbf{Example of a research overview by the Meta-review agent (continued)},
    fonttitle=\bfseries,
    arc=3mm,
    boxrule=1pt,
    bottomrule=2pt,
]
\footnotesize
\textbf{RNA Processing and Stress Granules:}
\begin{itemize}
    \item ...
\end{itemize}

\textbf{Protein Homeostasis and the Nucleocytoplasmic Transport:}
\begin{itemize}
    \item \textbf{Rationale:} The proper folding, trafficking, degradation, and localization of proteins are critical for cellular function. Defects in these processes can lead to the accumulation of toxic protein aggregates, disrupt cellular signaling, and initiate pathways leading to cell death in ALS. The nucleus is a key compartment, and transport into and out of this organelle is essential for cellular function.
    \item \textbf{Recent Findings:} Research has identified misfolded TDP-43 and FUS as key pathological features of ALS. Furthermore, the dysregulation of nucleocytoplasmic transport (NCT) and the nuclear pore complex (NPC) is a recurring feature, with recent findings that emphasize defects in early stages of assembly and dynamic function of the NPC as an early initiating step.
    \item \textbf{Areas of Research:}
    \begin{itemize}
        \item \textbf{Ubiquitin-Proteasome System (UPS) Dysfunction:}
        \begin{itemize}
            \item \textit{Why Research?} The UPS is a major cellular protein degradation machinery. Defects in the UPS can lead to an accumulation of misfolded proteins, which could lead to cellular stress and death. Exploring the potential of a functional deficiency in these system as an early trigger in disease.
            \item \textit{What to Research:} Investigate how UPS functionality is altered in ALS models, with particular attention to the role of E3 ligases in TDP-43 and FUS degradation. Determine whether specific modifications on ALS related proteins impair their degradation and whether specific pathways of protein quality control are affected in early stages of the disease. Assess if this is a primary cause or secondary outcome of other disease related pathways.
            \item \textit{Example Idea:} Use cell-based assays to measure UPS activity and substrate degradation rates in iPSC-derived motor neurons. Conduct proteomic analysis to identify specific ubiquitinated proteins that accumulate in ALS cells. Genetically manipulate UPS components or use small molecules to modulate their function and assess cellular responses (e.g. measuring protein aggregation and cell survival)
        \end{itemize}
        \item \textbf{Autophagy Dysfunction:}
        \begin{itemize}
            \item \textit{Why Research?} Autophagy is crucial for clearing cellular debris and aggregates. Dysfunctional autophagy can lead to the accumulation of toxic proteins and can lead to organelle defects.
            \item \textit{What to Research?} Explore how autophagy pathways are altered in ALS, focusing on the specific mechanisms of cargo recognition, autophagosome formation, and lysosome fusion. Assess if specific types of autophagy including mitophagy (mitochondrial specific) or lipophagy (lipid specific) are more vulnerable in specific cell types.
            \item \textit{Example Idea:} Track the formation and degradation of autophagosomes in real-time in hiPSC-derived motor neurons after different stress conditions. Manipulate key effectors of the autophagy pathway and observe cellular response, measuring protein aggregation, cell viability and inflammatory markers.
        \end{itemize}
        \item \textbf{Nucleocytoplasmic Transport (NCT) Defects:}
        \begin{itemize}
            \item ...
        \end{itemize}
        \item \textbf{Targeted Protein misfolding:}
        \begin{itemize}
            \item ...
        \end{itemize}
    \end{itemize}
    \item ...
\end{itemize}
\end{tcolorbox}
\vspace{0.1cm}
\caption{\textbf{An abbreviated example of a research overview generated by the Meta-review agent (continued).} This is an example of a research overview generated by the co-scientist Meta-review agent for the research goal pertaining to explaining the basic mechanism of Amyotrophic Lateral Sclerosis (ALS). In addition to the the individual hypotheses and proposals, this the main output presented to the scientist by the co-scientist system.}
\label{fig:ex_overview_2}
\end{figure}

\begin{figure}[htbp!]
\begin{tcolorbox}[
    colback=black!5!white,
    colframe=black!60!white,
    title=\textbf{Example of research contact identification by the Meta-review agent},
    fonttitle=\bfseries,
    arc=3mm,
    boxrule=1pt,
    bottomrule=2pt,
]
\footnotesize
\textbf{Research Direction: Oxidative DNA Damage \& Mitochondrial Base Excision Repair (BER) in ALS}
\begin{itemize}
    \item \textbf{[Researcher names]:} They directly study the role of mitochondrial OGG1 (a key BER enzyme) in controlling cytosolic mtDNA release and neuroinflammation. Their expertise is highly valuable for understanding the link between BER, mtDNA, and inflammation. Also, they have experience with experiments using mtOGG1 overexpressing mice which is relevant to the in-vitro experiments proposed.
    \item ...
\end{itemize}
\end{tcolorbox}
\vspace{0.1cm}
\caption{\textbf{An example of a research contact identified by the Meta-review agent as a potential domain expert in the research topic and hypothesis of interest.}}
\label{fig:ex_contact}
\end{figure}

\section{Supplementary information for drug repurposing evaluation}
\label{sec:supp_drug_repurposing}

\subsection{Datasets}
\label{sec:eval_data}

\paragraph{Cancer type dataset}
A list of 33 cancer types, along with their corresponding abbreviations, was constructed based on a curated text string derived from the \href{https://www.cancer.gov/tcga}{TCGA project} (Appendix \cref{tab:cancer_type}). This list includes 10 rare cancers.

\begin{table}[ht!]
\centering
\begin{tabular}{cc}
\toprule
\textbf{Cancer Name Abbreviation} & \textbf{Cancer Name} \\
\midrule
LAML & Acute Myeloid Leukemia \\
ACC & Adrenocortical carcinoma \\
BLCA & Bladder Urothelial Carcinoma \\
LGG & Brain Lower Grade Glioma \\
BRCA & Breast invasive carcinoma \\
CESC & Cervical squamous cell carcinoma and endocervical adenocarcinoma \\
CHOL & Cholangiocarcinoma \\
LCML & Chronic Myelogenous Leukemia \\
COAD & Colon adenocarcinoma \\
CNTL* & Controls \\
ESCA & Esophageal carcinoma \\
FPPP & FFPE Pilot Phase II \\
GBM & Glioblastoma multiforme \\
HNSC & Head and Neck squamous cell carcinoma \\
KICH & Kidney Chromophobe \\
KIRC & Kidney renal clear cell carcinoma \\
KIRP & Kidney renal papillary cell carcinoma \\
LIHC & Liver hepatocellular carcinoma \\
LUAD & Lung adenocarcinoma \\
LUSC & Lung squamous cell carcinoma \\
DLBC & Lymphoid Neoplasm Diffuse Large B-cell Lymphoma \\
MESO & Mesothelioma \\
MISC* & Miscellaneous \\
OV & Ovarian serous cystadenocarcinoma \\
PAAD & Pancreatic adenocarcinoma \\
PCPG & Pheochromocytoma and Paraganglioma \\
PRAD & Prostate adenocarcinoma \\
READ & Rectum adenocarcinoma \\
SARC & Sarcoma \\
SKCM & Skin Cutaneous Melanoma \\
STAD & Stomach adenocarcinoma \\
TGCT & Testicular Germ Cell Tumors \\
THYM & Thymoma \\
THCA & Thyroid carcinoma \\
UCS & Uterine Carcinosarcoma \\
UCEC & Uterine Corpus Endometrial Carcinoma \\
UVM & Uveal Melanoma \\
\bottomrule
\end{tabular}
\vspace{0.1cm}
\caption{\textbf{TCGA cancer type.} We consider all cancer types except CNTL and MISC while exploring the drug repurposing candidates with preclinical evidences. *Not a cancer type.}
\label{tab:cancer_type}
\end{table}

\paragraph{Drug repurposing proposals.}
We used the co-scientist to generate drug repurposing proposals for different types of cancer from a subset of 2300 drugs curated with the \href{https://platform.opentargets.org/downloads}{Open Targets Platform}. The proposal for each drug candidate includes a generated hypothesis, a review of the hypothesis, and a description of the possible mechanism of action. For each drug, the co-scientist assigned a drug repurposing likelihood score indicating how likely this drug can be successfully repurposed for a given cancer across all 33 TCGA cancer types.

\subsection{Computational biology analyses}
\label{sec:eval_compbio}
To enhance our drug repurposing hypothesis generation, verification, and evaluation processes, we integrated a computational biology approach utilizing DepMap (Cancer Dependency Map). DepMap is a project focused on identifying genetic vulnerabilities in cancer cells~\citep{depmap2024}. By systematically perturbing genes in a large panel of cancer cell lines using techniques like CRISPR and observing the resulting phenotypic effects, researchers are able to construct a comprehensive map of genetic dependencies for different cancers. This resource can be leveraged to identify promising drug targets for cancer therapies.

Specifically, DepMap generates perturbational data from CRISPR screens that are subsequently used to calculate DepMap dependency scores. These scores represent the probability of gene essentiality for a given cancer cell line. A high DepMap score indicates a high probability that a gene is required for cell survival and proliferation, thereby highlighting it as a potential therapeutic target. A DepMap score of 1.0 indicates a very strong signal, and any novel drug repurposing hypotheses with a high DepMap score would be interesting targets to explore.

For our experiments, we utilized the Q2 2024 data release of DepMap and used the default DepMap dependency probability. We first observed a significant correlation between DepMap score and the co-scientist score. As shown in Appendix~\cref{fig:cos_depmap}, drug-cancer pairs scored highly by the co-scientist also have high average DepMap score. We then leveraged DepMap data to assess the novelty of high-scoring proposals generated by the co-scientist. Candidates selected for expert review were required to meet stringent criteria, exhibiting both co-scientist review score $\geq 4$ and a DepMap score $\geq 0.99$.

\begin{figure}[htbp!]
\centering
\includegraphics[width=0.6\textwidth,keepaspectratio]{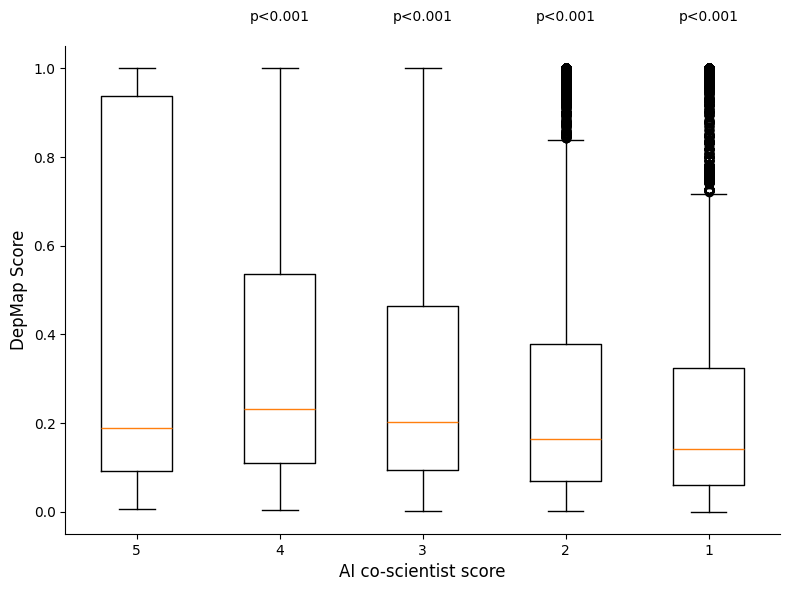}
\vspace{0.1cm}
\caption{\textbf{The co-scientist review score is concordant with the DepMap score.} We demonstrate the distribution of the DepMap score across five AI co-scientist review score groups. The group with the co-scientist score of 5 has the highest median DepMap score, and the group with co-scientist score of 1 has the lowest. All pairwise comparisons between the group of AI co-scientist score of 5 and each of the other groups are statistically significant (Two-sided Wilcoxon rank-sum test, $p < 0.001$).}
\label{fig:cos_depmap}
\end{figure}

\subsection{\textit{In vitro} wet-lab validation}
\label{sec:eval_wetlab}
We tested expert selected drug candidates for AML repurposing by measuring the half-maximal inhibitory concentration (IC50), the concentration of drug required to inhibit cell viability by 50\%, of each drug on representative AML cell lines for the respective cancer indication. IC50 is a value commonly used as a way to quantify the effectiveness of a drug at inhibiting cellular processes, and can be measured by treating cells across a broad range of drug concentrations and fitting the dose response to a sigmoidal curve to determine 50\% of maximal inhibition~\cite{sebaugh2011guidelines}. For our measurements, cells were plated in 96-well cell culture-treated plates, seeding 5000 cells per well, and treated with the respective drug concentrations for 48 hours. Subsequently, cell viability was assayed using an MTS assay (CellTiter 96 AQueous One Solution Cell Proliferation Assay, Promega). After 1 hour incubation with the MTS reagent at 37°C, absorbance at 490 nm was measured in each well using a plate reader. Each condition was performed in triplicate. Dose response curves were background corrected (subtracted the smallest response), normalized (divided by the largest response), fit, and plotted using the Quest Graph™ IC50 Calculator (AAT Bioquest, Inc). The cell lines used (MOLM13, HL60, and KG-1) were generous gifts from the lab of Dr. Ravi Majeti (Stanford University).
\begin{itemize}
    \item MOLM-13 cells are a human AML cell line that were derived from bone marrow samples of a 62-year-old woman with AML.
    \item HL60 cells are a human cell line from a patient with acute promyelocytic leukemia (APL), often used to study the development and proliferation of normal and leukemic cells. 
    \item KG-1 cells are a human cell line isolated from the bone marrow aspirate of a 59-year-old, white male with erythroleukemia that evolved into AML.
\end{itemize}

\subsection{Additional wet-lab results}
\label{sec:result_wetlab}
\begin{figure}[htbp!]
\centering
\includegraphics[width=0.8\textwidth,keepaspectratio]{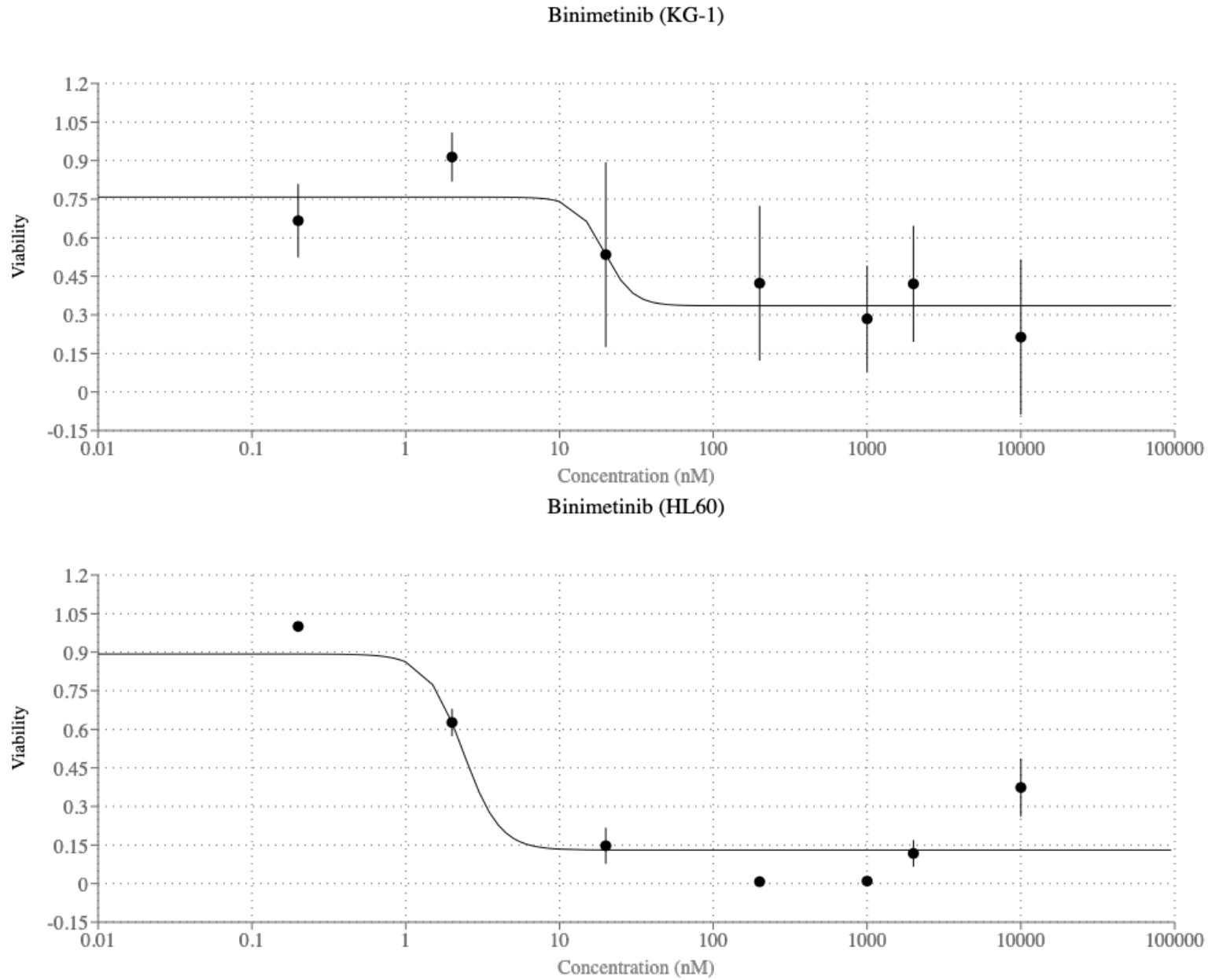}
\vspace{0.1cm}
\caption{\textbf{Dose-response of drug repurposing candidate Binimetinib in other AML cell lines.} Binimetinib demonstrates activity inhibiting cell viability in KG-1 and HL-60 cell lines. X-axis is the drug concentration (nM), and Y-axis is normalized cell viability (arbitrary unit).}
\label{fig:fig_wetlab_appendix}
\end{figure}

\begin{figure}[ht!]
\centering
\includegraphics[width=\textwidth,keepaspectratio]{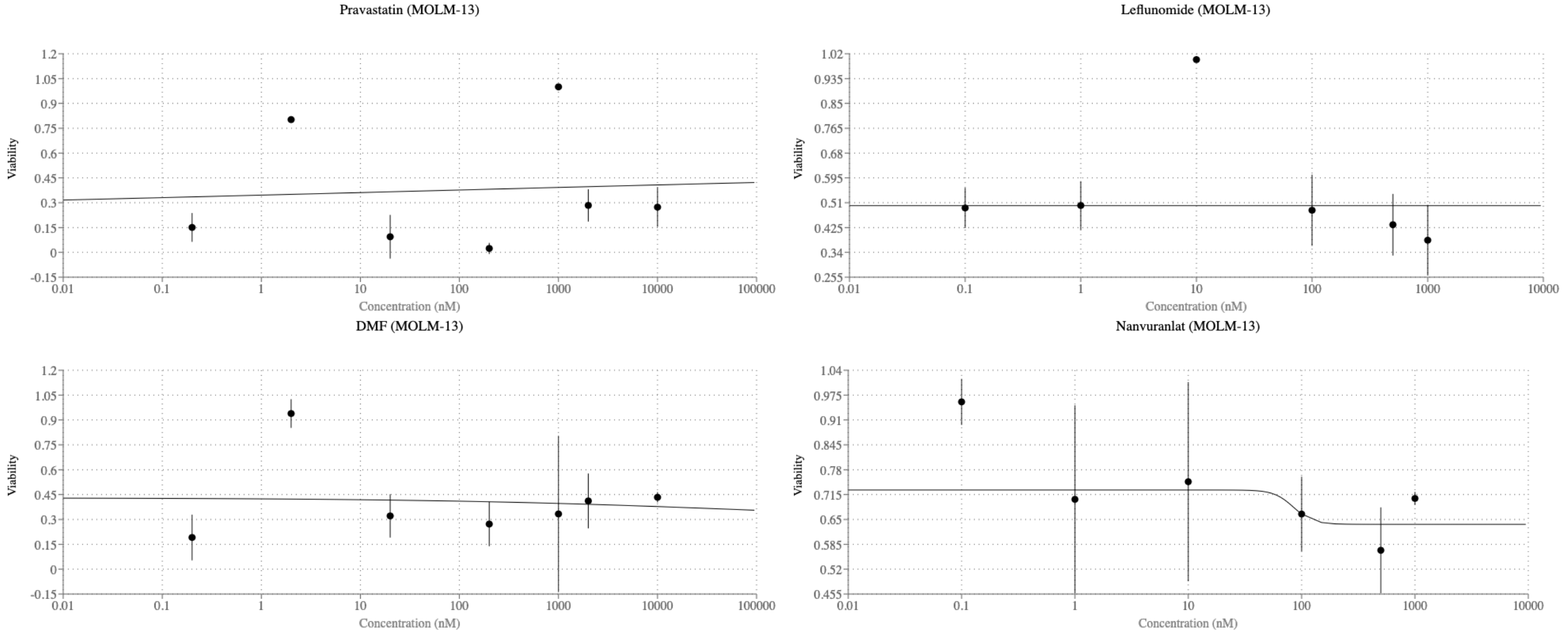}
\vspace{0.1cm}
\caption{\textbf{Dose-response of the drug repurposing candidates with little to no effect on MOLM-13.} Of the expert-selected drug repurposing candidates, Pravastatin and DMF showed little to no effect on the MOLM-13 cell line across the concentrations tested. Of the novel drug repurposing candidates, Leflunomide and Nanvuralat showed little to no effect on the MOLM-13 cell line across the concentrations tested. X-axis is the drug concentration (nM), and Y-axis is normalized cell viability (arbitrary unit).}
\label{fig:fig_wetlab_appendix_failed}
\end{figure}

This section provides additional \textit{in vitro} laboratory results of drug repurposing. Appendix \cref{fig:fig_wetlab_appendix} shows dose-response curves of Binimetinib in KG-1 and HL-60 cell lines. Appendix \cref{fig:fig_wetlab_appendix_failed} shows the does response curves of drug repurposing candidates for AML suggested by the AI co-scientist with little to no effect on MOLM-13 cells.

\clearpage
\section{Supplementary information for the co-scientist drug repurposing Specific Aims evaluation}
\label{sec:supp_aims}

\subsection{Specific Aims count distribution}
\label{sec:specific_aims_count}

\begin{table}[htbp!]
\centering
\begin{tabular}{cc}
\toprule
\textbf{Repurposed Cancer} & \textbf{Count} \\
\midrule
Acute Myeloid Leukemia (LAML) & 13 \\
Colon adenocarcinoma (COAD) & 10 \\
Breast invasive carcinoma (BRCA) & 10 \\
Skin Cutaneous Melanoma (SKCM) & 8 \\
Lung adenocarcinoma (LUAD) & 6 \\
Head and Neck squamous cell carcinoma (HNSC) & 5 \\
Bladder Urothelial Carcinoma (BLCA) & 5 \\
Pancreatic adenocarcinoma (PAAD) & 4 \\
Stomach adenocarcinoma (STAD) & 3 \\
Rectum adenocarcinoma (READ) & 3 \\
Esophageal carcinoma (ESCA) & 3 \\
Uveal Melanoma (UVM) & 2 \\
Uterine Corpus Endometrial Carcinoma (UCEC) & 2 \\
Prostate adenocarcinoma (PRAD) & 2 \\
Lung squamous cell carcinoma (LUSC) & 1 \\
Cervical squamous cell carcinoma and endocervical adenocarcinoma (CESC) & 1 \\
\midrule
Grand Total & 78 \\
\bottomrule
\end{tabular}
\vspace{0.1cm}
\caption{\textbf{Count distribution of the cancer types in the Specific Aims drug repurposing proposals generated by the AI co-scientist.} }
\end{table}

\subsection{Specific Aims evaluation rubric}
\label{sec:specific_aims_rubric}
A pilot evaluation framework was developed by oncologists at a US institute, inspired by the axes used in evaluating research proposals in their field (for example, the NIH Specific Aims review criteria). While such original criteria serve as a gold standard for the review of in-depth proposals written by human domain experts, a context-specific approach is needed for AI systems such as the AI co-scientist because there are AI-specific limitations to consider.

The evaluation framework therefore focused on enabling assessment of similar axes of proposal quality viewed to be important by collaborating oncologists, while additionally accounting for unique limitations and capabilities of LLM-based systems.

For instance, while precedent real-world criteria assume deep domain expertise and the ability to cite contemporary unpublished work, the pilot framework presented here emphasizes the evaluation of logical consistency and the appropriate use of publicly available scientific knowledge. This acknowledges that LLMs, unlike human experts, may struggle to draw upon tacit knowledge or recent unpublished findings in their field.

The resulting pilot framework explores two domains: (1) clinical significance and (2) scientific rigor and methodology, with 5 and 10 questions respectively, for a total of 15 evaluation criteria.

A 5-point Likert-type scale is used for each item (Strongly Disagree, Disagree, Neutral, Agree and Strongly Agree). The approach used here represents a pilot framework only and would require considerable further research if it were to be developed into a reproducible or valid ratings instrument, including assessment of validity and reliability, inter and intra-observer variance, and correlation of ratings with ground truths or well-established prior frameworks not designed to evaluate AI systems.

The ``clinical significance'' domain addressed some fundamental aspects of drug repurposing proposals, including unmet clinical needs, therapeutic landscape analysis, and scientific rationale. Items in this domain aimed to assess the clinical relevance and theoretical foundation of proposals, and how AI systems integrate and synthesize publicly available information.

The ``scientific rigor and methodology'' domain aimed to evaluate technical quality of proposals, including hypothesis formation, methodological clarity, translational potential, overall scientific accuracy as reflected by the quality of preclinical experimental design and the presence of well-defined clinical endpoints. Special attention was paid to evaluating LLM-specific concerns, such as the assessment of factual accuracy and the avoidance of hallucinated content. This pilot framework is not comprehensive and is not intended to mirror the full breadth and depth of grant design evaluations. For example, omitted axes of quality could be addressed by future works, including absence of detailed criteria for experimental design (for example, review of power calculation or study design methodologies).

\subsection*{Significance and innovation (5 questions)}
\begin{enumerate}
    \item The proposal adequately identifies significant unmet clinical needs that could be addressed through drug repurposing of this pharmacological agent.
    \item The proposal effectively bridges an important gap in the current therapeutic landscape for the target disease by repurposing this drug.
    \item The proposal presents a scientifically rigorous rationale for repurposing the drug, grounded in current evidence and literature.
    \item The scientific background integrates relevant prior studies and preliminary data to support the proposed drug repurposing.
    \item The proposal avoids over-extrapolation or speculative conclusions beyond the supporting evidence.
\end{enumerate}

\subsection*{Rigor and feasibility (10 questions)}
\begin{enumerate}
    \item Does each Specific Aim have a clear hypothesis, and specific methodological approaches?
    \item Are the Specific Aims clearly stated and logically organized?
    \item There is a clear path from the proposed research to clinical application, including consideration of necessary pre-clinical and clinical studies.
    \item The proposal includes well-defined and clinically relevant endpoints.
    \item The pre-clinical experiments are designed to yield meaningful insights into the drug's potential efficacy and safety in its new therapeutic role.
    \item The proposal includes a well-described translational component to assess the drug's efficacy in its new indication.
    \item The proposal avoids factual inaccuracies or hallucinations, presenting information that is accurate and reliable.
    \item The assumptions made within the proposal regarding the drug's efficacy, mechanism, and potential outcomes are grounded in current scientific evidence.
    \item The proposal demonstrates originality in thought and approach, employs precise and appropriate scientific terminology and avoids ambiguous or generic statements.
    \item The proposal is clearly written, with well-organized sections and a logical flow of ideas that enhances comprehension and ensures the aims are easily understood.
\end{enumerate}

\newpage

\vspace{-0.2cm}
\subsection{Examples of co-scientist generated Specific Aims with expert evaluation}
\label{sec:ex_specific_aims}

\vspace{-0.2cm}
\begin{figure}[htbp!]
\caption{\textbf{Example of co-scientist generated Specific Aims drug repurposing proposal for AML.}}
\label{fig:aim1_1}
\vspace{0.1cm}
\begin{tcolorbox}[title=Givosiran for AML]
\footnotesize
\textbf{Disease Description}\\
Acute myeloid leukemia (AML) is an aggressive hematological malignancy with an incidence rate of approximately 4.3 per 100,000 persons per year in the United States, predominantly affecting older adults.  AML arises from a complex interplay of genetic mutations, epigenetic alterations, and dysregulated signaling pathways, disrupting normal hematopoiesis by blocking myeloid differentiation and promoting uncontrolled blast proliferation.  This leads to bone marrow failure, infections, bleeding, and other life-threatening complications.  The current standard of care includes intensive chemotherapy, often combined with targeted therapies or hematopoietic stem cell transplantation.  However, these treatments are often associated with significant toxicity, high relapse rates, and limited efficacy in certain patient populations, particularly in relapsed/refractory or high-risk AML.\\

\textbf{Unmet Need}\\
Despite advancements, significant unmet needs persist in AML treatment.  Current therapies often cause severe side effects, particularly in older or frail patients, limiting their tolerability and effectiveness.  Relapse rates remain high, and treatment options for relapsed/refractory AML are limited and often less effective.  Many patients fail to achieve complete remission or experience only short-lived responses, underscoring the urgent need for novel, less toxic, and more effective therapies, especially for patients with relapsed/refractory or high-risk disease.\\

\textbf{Proposed Solution}\\
Givosiran sodium is an RNA interference (RNAi) therapeutic approved for acute hepatic porphyria (AHP). It targets aminolevulinate synthase 1 (ALAS1) mRNA, the rate-limiting enzyme in heme biosynthesis, reducing the production of heme precursors $\delta$-aminolevulinic acid (ALA) and porphobilinogen. This prevents the accumulation of neurotoxic heme intermediates in AHP.

Repurposing givosiran for AML stems from the crucial role of heme biosynthesis in AML pathogenesis.  Several studies suggest that disrupting heme biosynthesis offers a therapeutic advantage in AML. The proposed approach addresses the unmet need by targeting AML cells dependent on increased heme biosynthesis, particularly those with MYCN overexpression [1]. Modulating heme levels could influence oxidative stress [5, 6], apoptosis [1, 2], and drug sensitivity in AML [2]. Givosiran's ALAS1 inhibition aligns with this approach, offering a novel therapeutic strategy.

We hypothesize that givosiran, by inhibiting ALAS1 and reducing heme biosynthesis, will suppress AML cell growth and survival, particularly in those with upregulated heme biosynthesis. This is supported by preclinical evidence demonstrating that altering heme levels impacts AML cell proliferation, apoptosis, and drug sensitivity [2]. Abstract [1] suggests that elevated heme biosynthesis in MYCN-driven AML is a therapeutic vulnerability.  Reducing ALA and porphobilinogen accumulation via givosiran can mitigate oxidative stress [6], a factor implicated in AML progression.

Overall, our goal is to evaluate the efficacy and safety of givosiran as a novel therapeutic strategy for AML by exploiting the crucial role of heme biosynthesis in its pathogenesis and the drug's ability to modulate heme levels and downstream effects on AML cell proliferation, survival, and drug sensitivity.\\

\textbf{Specific Aims 1}\\
\textbf{Overarching goal:}\\
Determine the anti-leukemic activity of givosiran in MYCN-driven AML models.\\
\textbf{Hypothesis:}\\
Givosiran treatment will decrease the viability and proliferation of MYCN-overexpressing AML cells in vitro and reduce tumor growth in MYCN-driven AML xenograft mouse models.\\
\textbf{Reasoning:}\\
MYCN-driven AML frequently exhibits upregulated heme biosynthesis [1], creating a potential dependency on this pathway. Givosiran, by inhibiting ALAS1, could disrupt this dependency, leading to decreased heme and growth inhibition. This is supported by preclinical data showing that inhibiting heme biosynthesis impacts AML cell growth and survival [1, 2].\\
\end{tcolorbox}
\end{figure}

\begin{figure}[htbp!]
\caption{\textbf{Example of co-scientist generated Specific Aims drug repurposing proposal for AML (continued).}}
\label{fig:aim1_2}
\vspace{0.1cm}
\begin{tcolorbox}[title=Givosiran for AML (continued)]
\footnotesize
\textbf{Specific Aims 2}\\
\textbf{Overarching goal:}\\
Elucidate the impact of givosiran on oxidative stress and drug sensitivity in AML.\\
\textbf{Hypothesis:}\\
Givosiran treatment will modulate oxidative stress levels and enhance the cytotoxic effects of standard AML chemotherapeutics (e.g., cytarabine) in AML cell lines and primary patient samples.\\
\textbf{Reasoning:}\\
Heme plays a role in oxidative stress regulation, and its modulation by givosiran could influence AML cell chemosensitivity.  Abstracts [5, 6] highlight oxidative stress in AML and the potential for ALA accumulation to contribute to it. By reducing ALA and heme, givosiran could alter reactive oxygen species (ROS) levels and potentially sensitize AML cells to chemotherapy-induced death.\\

\textbf{Specific Aims 3}\\
\textbf{Overarching goal:}\\
Characterize the safety and tolerability of givosiran in preclinical AML models, focusing on its impact on liver function.\\
\textbf{Hypothesis:}\\
Givosiran treatment will be well-tolerated in AML mouse models, with minimal adverse effects on liver function and drug metabolism, at doses that effectively inhibit ALAS1 and reduce heme biosynthesis.\\
\textbf{Reasoning:}\\
Given givosiran's hepatic target (ALAS1), evaluating its safety profile in AML is crucial. Abstracts [7, 8] highlight the clinical significance of liver function in AML patients. This aim will assess potential hepatotoxicity and drug-drug interactions, ensuring safe translation to clinical trials. We will evaluate relevant liver function markers and givosiran's impact on standard AML chemotherapeutic metabolism.\\

\textbf{Pilot Evaluation}\\
A pilot study in a human AML xenograft mouse model will assess givosiran's in vivo efficacy and safety.  Givosiran will be administered at various doses, monitoring tumor growth, survival, and liver function. The primary endpoint will be tumor growth inhibition. Secondary endpoints include survival, changes in heme levels, oxidative stress markers, and liver function tests. Existing safety data from givosiran's use in AHP will inform dose selection and monitoring. While givosiran is approved for AHP, its use in AML requires an Investigational New Drug (IND) application to the FDA before clinical trials.  Existing safety data might facilitate a streamlined review process.\\

\textbf{Articles}

[1] Upregulated heme biosynthesis, an exploitable vulnerability in MYCN-driven leukemogenesis
Summary: This study demonstrates increased heme biosynthesis in MYCN-driven leukemias, suggesting it as a therapeutic target.
Relevance: Supports the rationale for targeting heme biosynthesis in MYCN-driven AML and provides a rationale for the proposed mechanism of action of givosiran in this context.

[2] Systematic Dissection of the Metabolic-Apoptotic Interface in AML Reveals Heme Biosynthesis to Be a Regulator of Drug Sensitivity
Summary: This study highlights the role of heme biosynthesis in regulating apoptosis and drug sensitivity in AML.
Relevance: Supports the potential for givosiran to impact AML cell survival and treatment response by modulating heme biosynthesis.

[3] 3113 – IDENTIFICATION OF HEME AS A CRITICAL REGULATOR OF LEUKEMOGENESIS
Summary: This abstract points to the importance of heme levels in AML, especially in leukemic stem cells, and suggests targeting heme biosynthesis.
Relevance: Reinforces the rationale for targeting heme biosynthesis as a therapeutic strategy in AML.

...

 [8] Acute myeloid leukemia with hepatic infiltration presenting as obstructive jaundice
Summary: This abstract shows liver dysfunction in AML and the need to consider liver health in treatment.
Relevance: Reinforces the importance of monitoring liver function in AML patients treated with givosiran, further supporting the rationale for Aim 3.
\end{tcolorbox}
\end{figure}

\begin{figure}[htbp!]
\caption{\textbf{Example of co-scientist generated Specific Aims drug repurposing proposal for colon cancer.}}
\label{fig:aim2_1}
\vspace{0.1cm}
\begin{tcolorbox}[title=Selinexor for colon cancer]
\footnotesize
\textbf{Disease Description}\\
Colon adenocarcinoma (COAD) is a significant public health concern, with an estimated incidence rate of over 1.9 million new cases and 935,000 deaths globally in 2020. COAD arises from the epithelial lining of the colon and is driven by a complex interplay of genetic mutations (e.g., APC, KRAS, BRAF, TP53) and epigenetic alterations, leading to uncontrolled cell proliferation, impaired apoptosis, and chronic inflammation. These molecular changes disrupt crucial cellular pathways like Wnt/$\beta$-catenin signaling, cell cycle control, and DNA damage repair, ultimately driving tumor progression. COAD typically progresses through stages, from localized polyps to invasive tumors with potential for metastasis. Current standard of care involves surgery, chemotherapy, radiation therapy, and targeted therapies, but treatment resistance and recurrence remain major challenges, highlighting the need for new therapeutic approaches.

\textbf{Unmet Need}\\
Current COAD treatments have limitations, including acquired resistance to chemotherapy, significant toxicities, and incomplete responses in advanced disease. Patients often experience diminished quality of life due to treatment side effects and disease burden. High recurrence rates and limited effective treatment options after progression contribute to poor long-term outcomes.  This unmet need underscores the urgency for novel therapeutic strategies that can overcome resistance, improve response rates, and minimize toxicity, ultimately extending survival and enhancing quality of life for COAD patients.

\textbf{Proposed Solution}\\
Selinexor, a first-in-class selective inhibitor of nuclear export (SINE), is currently approved for the treatment of multiple myeloma and diffuse large B-cell lymphoma. It specifically targets XPO1 (Exportin 1), a key protein responsible for the nuclear export of tumor suppressor proteins, oncoproteins, and RNA. By binding to XPO1, Selinexor blocks the nuclear export of these molecules, leading to their accumulation in the nucleus and restoration of tumor suppressor function, cell cycle arrest, and apoptosis induction.

Repurposing Selinexor for COAD is rationally supported by its mechanism of action and the molecular characteristics of the disease. Overexpression of XPO1 is common in various cancers, including COAD [1, 2, 6]. Selinexor inhibits XPO1, preventing the nuclear export and restoring the function of key tumor suppressors (p53, RB, FOXO, APC) frequently dysregulated in COAD [2, 3, 4, 5]. Furthermore, Selinexor can suppress constitutively activated NF-$\kappa$B signaling, a driver of chronic inflammation and tumor progression in COAD, by blocking I$\kappa$B export and increasing its nuclear accumulation [7, 8]. These mechanisms align with key aspects of COAD pathogenesis and offer opportunities for therapeutic intervention.

We hypothesize that Selinexor will effectively inhibit COAD cell growth and enhance apoptosis by restoring tumor suppressor function and modulating crucial signaling pathways. Selinexor has shown anti-tumor activity in solid tumors in a Phase I trial [3], with observations of nuclear accumulation of tumor suppressor proteins [3, 6]. While not specific to COAD, these findings, coupled with evidence of Selinexor's efficacy in other cancers driven by XPO1 overexpression [3, 6], suggest that a similar mechanism could be effective in COAD. Further supporting our hypothesis, XPO1 overexpression has been linked to NF-$\kappa$B activation and increased proliferation in COAD [7].
Overall, our goal is: to demonstrate that Selinexor’s XPO1 inhibitory activity can effectively target key oncogenic drivers and restore tumor suppressor functions in COAD, ultimately leading to tumor growth inhibition and improved patient outcomes.

\textbf{Specific Aims 1}\\
\textbf{Overarching goal:}\\
Determine the in vitro efficacy of Selinexor in inhibiting COAD cell growth and inducing apoptosis.\\
\textbf{Hypothesis:}\\
Selinexor treatment will significantly reduce the viability and increase apoptosis in a panel of COAD cell lines, including those with varying genetic backgrounds (e.g., APC, KRAS, TP53 mutations).\\
\textbf{Reasoning:}\\
Selinexor's inhibition of XPO1 leads to nuclear accumulation of tumor suppressors like p53, a key regulator of apoptosis [2, 6]. Restoration of p53 function and suppression of NF-$\kappa$B, a pro-survival pathway [7, 8], are expected to induce apoptosis in COAD cells. We will evaluate this hypothesis using cell viability and apoptosis assays in diverse COAD cell lines to assess the impact of genetic background on Selinexor's efficacy.\\
\end{tcolorbox}
\end{figure}

\begin{figure}[htbp!]
\caption{\textbf{Example of co-scientist generated Specific Aims drug repurposing proposal for colon cancer (continued).}}
\label{fig:aim2_2}
\vspace{0.1cm}
\begin{tcolorbox}[title=Selinexor for colon cancer (continued)]
\footnotesize
\textbf{Specific Aims 2}\\
\textbf{Overarching goal:}\\
Elucidate the mechanisms by which Selinexor inhibits COAD cell growth, focusing on XPO1-mediated restoration of tumor suppressor function.\\
\textbf{Hypothesis:}\\
Selinexor treatment will increase nuclear accumulation of p53, RB, and FOXO proteins and decrease nuclear export of I$\kappa$B, leading to cell cycle arrest and decreased NF-$\kappa$B activity in COAD cells.\\
\textbf{Reasoning:}\\
Selinexor blocks XPO1, preventing nuclear export of key tumor suppressors (p53, RB, FOXO) [2, 4, 5] and the NF-$\kappa$B inhibitor I$\kappa$B [7, 8]. Increased nuclear localization of these proteins should restore their growth regulatory functions, leading to cell cycle arrest and reduced NF-$\kappa$B-driven proliferation. We will evaluate this hypothesis using immunofluorescence and western blotting to assess protein localization and activity.

\textbf{Specific Aims 3}\\
\textbf{Overarching goal:}\\
Characterize the potential synergistic effects of Selinexor in combination with standard-of-care chemotherapies for COAD.\\
\textbf{Hypothesis:}\\
Combination treatment with Selinexor and 5-fluorouracil (5-FU) will synergistically reduce COAD cell viability compared to either treatment alone.\\
\textbf{Reasoning:}\\
XPO1 inhibition can sensitize cancer cells to chemotherapy [9]. Combining Selinexor with 5-FU, a common COAD chemotherapy, may enhance cellular stress and improve treatment response. We will test this hypothesis using cell viability assays and investigate the underlying mechanisms of synergy.

\textbf{Pilot Evaluation}\\
A pilot study will evaluate Selinexor's efficacy in a patient-derived xenograft (PDX) model of COAD. The primary objective is to determine the effect of Selinexor on tumor growth. The study will utilize an open-label, single-arm design with escalating Selinexor doses in established COAD PDX models. Inclusion criteria: established COAD PDX models. Exclusion criteria: none. Primary endpoint: tumor volume change. Secondary endpoints: changes in biomarkers (p53, Ki67, NF-$\kappa$B) within the tumor. Selinexor's established safety profile in other cancers provides a basis for evaluating its safety in this new indication. Potential for expedited review through existing regulatory pathways will be explored.\\
\\
\textbf{Expert rating}\\
1. Strongly Agree (unmet clinical needs)\\
2. Agree (bridges therapeutic gap)\\
3. Strongly Agree (scientifically rigorous rationale)\\
4. Agree (integrates prior studies)\\
5. Agree (avoids over-extrapolation)\\
6. Strongly Agree (clear hypotheses and methods)\\
7. Strongly Agree (clearly stated aims)\\
8. Agree (path to clinical application)\\
9. Agree (well-defined endpoints)\\
10. Agree (meaningful pre-clinical experiments)\\
11. Agree (translational component)\\
12. Strongly Agree (avoids inaccuracies)\\
13. Strongly Agree (evidence-based assumptions)\\
14. Agree (originality and terminology)\\
15. Strongly Agree (clear writing and organization)\\
\end{tcolorbox}
\end{figure}

\begin{figure}[htbp!]
\caption{\textbf{Example of co-scientist generated Specific Aims drug repurposing proposal for colon cancer.}}
\label{fig:aim3_1}
\vspace{0.1cm}
\begin{tcolorbox}[title=Lapatinib for colon cancer]
\footnotesize
\textbf{Disease Description}\\
Colon adenocarcinoma (COAD) is a significant public health concern, with an estimated incidence rate of over 150,000 new cases annually in the United States [19]. COAD arises from the epithelial lining of the colon, and its pathogenesis involves a complex interplay of genetic and environmental factors. Key molecular mechanisms include mutations in genes such as APC, KRAS, BRAF, and PIK3CA, as well as dysregulation of signaling pathways like Wnt, RAS/RAF/MEK/ERK, and PI3K/AKT/mTOR [8, 9, 12]. These alterations drive uncontrolled cell proliferation, evade apoptosis [14], promote angiogenesis [15], and ultimately lead to tumor growth, invasion, and metastasis [16]. The disease typically progresses through a series of stages, from localized tumors to regional lymph node involvement and distant metastasis. Current standard of care involves surgery, chemotherapy, and targeted therapies such as anti-EGFR antibodies. However, these treatments have limitations, including acquired resistance and significant toxicity.

\textbf{Unmet Need}\\
A major unmet need in COAD treatment is the development of effective therapies for patients who progress on or are refractory to standard treatments, particularly those with resistance to anti-EGFR therapy [2]. Despite available therapies, many patients experience disease recurrence and metastasis, leading to poor outcomes and diminished quality of life. There's a critical need for new therapeutic options that can overcome resistance mechanisms, improve response rates, and offer better tolerability profiles. Specifically, addressing resistance driven by KRAS mutations [10] and exploring alternative therapeutic targets remains crucial.

\textbf{Proposed Solution}\\
Lapatinib is an orally available small molecule tyrosine kinase inhibitor currently approved for use in combination with capecitabine for the treatment of HER2-positive metastatic breast cancer. It reversibly inhibits the intracellular tyrosine kinase domains of both EGFR (ErbB1) and HER2 (ErbB2), thereby blocking downstream signaling cascades, including RAS/RAF/MEK/ERK and PI3K/AKT/mTOR. This inhibition leads to decreased cell proliferation and increased apoptosis.

Repurposing lapatinib for COAD is rationalized by the shared ErbB signaling pathway between breast cancer and a subset of COAD. EGFR is commonly overexpressed in COAD [1], and while HER2 overexpression is less frequent than in breast cancer, it occurs in a clinically relevant subset [3, 17] and is associated with resistance to anti-EGFR therapy [2]. Lapatinib can directly inhibit both EGFR and HER2, potentially disrupting crucial oncogenic signaling [13, 14] including PLC$\gamma$ [7].

We hypothesize that lapatinib can effectively inhibit ErbB signaling in COAD, leading to decreased cell proliferation, increased apoptosis, and suppression of metastasis. Preclinical studies demonstrate synergistic antitumor activity of lapatinib with HDAC inhibitors in COAD models [6], and lapatinib has also been shown to sensitize COAD cells to TRAIL-induced apoptosis via an off-target mechanism [5]. Studies have also investigated HER2 as a therapeutic target in CRC, especially after failure of anti-EGFR therapy [2, 18]. Although lapatinib as a single agent has shown limited efficacy in unselected CRC populations, this could be attributed to the heterogeneity of COAD and the presence of resistance mechanisms like KRAS mutations. We anticipate that patient stratification based on molecular profiles will identify subgroups that derive greater benefit.

Overall, our goal is: to demonstrate the efficacy of lapatinib in specific molecular subtypes of COAD, potentially in combination with other targeted therapies, to address the unmet need for new treatment options, particularly in patients resistant to standard therapies.

\textbf{Specific Aims 1}\\
\textbf{Overarching goal:}\\
To determine the efficacy of lapatinib in inhibiting HER2/EGFR signaling and suppressing cell proliferation in a panel of well-characterized COAD cell lines.\\
\textbf{Hypothesis:}\\
Lapatinib will inhibit cell proliferation in HER2-amplified/overexpressing and/or EGFR-overexpressing COAD cell lines.\\
\textbf{Reasoning:}\\
Lapatinib directly inhibits both HER2 and EGFR, key drivers of cell proliferation in a subset of COAD. Inhibition of these receptors should lead to reduced downstream signaling through the RAS/RAF/MEK/ERK and PI3K/AKT/mTOR pathways, ultimately suppressing cell growth [6].\\
\end{tcolorbox}
\end{figure}

\begin{figure}[htbp!]
\caption{\textbf{Example of co-scientist generated Specific Aims drug repurposing proposal for colon cancer (continued).}}
\label{fig:aim3_2}
\vspace{0.1cm}
\begin{tcolorbox}[title=Lapatinib for colon cancer (continued)]
\footnotesize
\textbf{Specific Aims 2}\\
\textbf{Overarching goal:}\\
To elucidate the impact of lapatinib on apoptosis and key signaling pathways in COAD models.\\
\textbf{Hypothesis:}\\
Lapatinib will induce apoptosis and modulate key signaling pathways (RAS/RAF/MEK/ERK, PI3K/AKT/mTOR, and PLC$\gamma$) in COAD cell lines, especially those with HER2/EGFR alterations.\\
\textbf{Reasoning:}\\
Inhibition of HER2/EGFR by lapatinib disrupts pro-survival signaling, potentially promoting apoptosis [14].  Additionally, lapatinib can indirectly modulate downstream pathways [4, 7], even demonstrating synergistic effects with other agents [6]. This aim will explore the extent of this modulation in various COAD subtypes.

\textbf{Specific Aims 3}\\
\textbf{Overarching goal:}\\
To identify predictive biomarkers for lapatinib response and explore potential combination strategies in COAD.\\
\textbf{Hypothesis:}\\
A combination of lapatinib with a MEK inhibitor will enhance antitumor activity in KRAS-mutant COAD cell lines compared to lapatinib alone.\\
\textbf{Reasoning:}\\
KRAS mutations are a known mechanism of resistance to EGFR/HER2 targeted therapies [10]. Combining lapatinib with a MEK inhibitor may overcome this resistance by blocking the downstream MAPK pathway activation, leading to enhanced antitumor efficacy.

\textbf{Pilot Evaluation}\\
A pilot study will be conducted to evaluate the safety and preliminary efficacy of lapatinib in patients with HER2-positive metastatic COAD who have progressed on standard therapy. This open-label, single-arm study will enroll 15-20 patients with confirmed HER2 overexpression/amplification and KRAS wild-type status. Key exclusion criteria will include prior treatment with lapatinib and significant comorbidities. The primary endpoint will be progression-free survival, and secondary endpoints will include overall response rate, duration of response, and safety. Lapatinib's known safety profile from its use in breast cancer will inform the monitoring plan. Expedited regulatory review may be possible due to lapatinib's existing approval.

\textbf{Articles}

[1] Targeting the EGFR signalling pathway in metastatic colorectal cancer\\
Summary: Details the importance of EGFR signaling in CRC and how targeting it is a crucial therapeutic strategy.\\
Relevance: Provides context for lapatinib's action as it targets EGFR.

[2] HER2 as an Emerging Oncotarget for Colorectal Cancer Treatment After Failure of Anti-Epidermal Growth Factor Receptor Therapy\\
Summary: Discusses HER2 as a target in CRC, especially in the context of anti-EGFR therapy resistance.\\
Relevance: Directly relevant to lapatinib's mechanism and potential in COAD.

...

[19] United States Cancer Statistics\\
Summary: Provides official U.S. cancer incidence and mortality statistics.\\
Relevance: Source of epidemiological data for colon adenocarcinoma.\\
\\
\textbf{Expert rating}\\
1.	Strongly Agree (unmet clinical needs)\\
2.	Agree (bridges therapeutic gap)\\
3.	Strongly Agree (scientifically rigorous rationale)\\
4.	Strongly Agree (integrates prior studies)\\
5.	Strongly Agree (avoids over-extrapolation)\\
6.	Strongly Agree (clear hypotheses and methods)\\
7.	Strongly Agree (clearly stated aims)\\
8.	Strongly Agree (path to clinical application)\\
9.	Agree (well-defined endpoints)\\
10.	Neutral (meaningful pre-clinical experiments)\\
11.	Strongly Agree (translational component)\\
12.	Strongly Agree (avoids inaccuracies)\\
13.	Strongly Agree (evidence-based assumptions)\\
14.	Strongly Agree (originality and terminology)\\
15.	Agree (clear writing and organization)\\
\end{tcolorbox}
\end{figure}

\clearpage
\subsection{Detailed AI co-scientist output for a validated AML repurposing candidate}
\label{sec:drug_output}
\begin{figure}[htbp!]
\begin{tcolorbox}[
    colback=black!5!white,
    colframe=black!60!white,
    title=\textbf{KIRA6 for AML},
    fonttitle=\bfseries,
    arc=3mm,
    boxrule=1pt,
    bottomrule=2pt,
]
\scriptsize
\textbf{Summary}

This proposal outlines repurposing KIRA6, an IRE1$\alpha$ inhibitor, for treating acute myeloid leukemia (AML), particularly FLT3-ITD positive AML known for high relapse rates. The hypothesis is that KIRA6 will disrupt protein homeostasis and induce ER stress in AML cells, synergizing with existing therapies and overcoming resistance. The proposal details KIRA6's mechanism, impacting pathways like the UPR and apoptosis, and suggests a comprehensive experimental plan including in vitro and in vivo studies to validate its efficacy and safety, including combination therapy with existing AML drugs.  This approach aims to provide a novel, effective treatment strategy for AML patients, particularly those with the challenging FLT3-ITD mutation and other drug resistance phenotypes. \\

\textbf{Hypothesis}

\textbf{Introduction} Acute myeloid leukemia (AML) remains a formidable hematological malignancy characterized by the rapid proliferation of immature myeloid blasts in the bone marrow and peripheral blood. Despite advancements in therapeutic strategies, including targeted therapies and hematopoietic stem cell transplantation, a significant proportion of patients experience relapse or develop resistance to treatment, underscoring the urgent need for new therapeutic interventions. The heterogeneous nature of AML, with diverse genetic and molecular aberrations driving disease progression, necessitates the exploration of novel therapeutic targets and the rational design of combination therapies to overcome treatment resistance and improve long-term survival outcomes. Drug repurposing, which involves identifying new uses for existing drugs, offers a cost-effective and efficient strategy to accelerate the development of novel AML treatments by leveraging established safety profiles and pharmacokinetic data. This approach can significantly reduce development time and costs, facilitating the rapid translation of promising agents to clinical trials. \\

\textbf{Recent findings and related research}

Recent research has significantly advanced our understanding of the complex molecular landscape of AML, highlighting the critical roles of various genetic mutations and signaling pathways in leukemogenesis. Recurrent mutations in genes such as FLT3, NPM1, IDH1/2, and TP53 are frequently observed in AML and have been implicated in disease initiation, progression, and treatment resistance. The FLT3-ITD mutation, a particularly challenging subtype, is associated with poor prognosis and a higher risk of relapse, emphasizing the need for effective therapies that target this specific mutation or overcome its downstream effects. Targeted therapies, such as FLT3 inhibitors (midostaurin, gilteritinib), IDH inhibitors (enasidenib, ivosidenib), and the BCL-2 inhibitor venetoclax, have demonstrated clinical efficacy in specific AML subtypes; however, the emergence of drug resistance and the lack of effective treatments for high-risk patients remain critical challenges. \\

Drug repurposing has yielded several promising candidates for AML treatment, including:
\begin{itemize}
    \item Arsenic trioxide and all-trans retinoic acid (ATRA):  Established treatments for acute promyelocytic leukemia (APL), these have also shown potential in combination therapy for non-APL AML.
    \item Histone deacetylase (HDAC) inhibitors:  Vorinostat and panobinostat have shown modest activity in AML, particularly in combination with other agents, but are known to be poorly tolerated and have limited efficacy as single agents.
    \item Proteasome inhibitors: Bortezomib and carfilzomib have been explored in combination regimens in AML, especially to overcome resistance, though with limited success.
    \item Metabolic inhibitors: Targeting glutamine metabolism and other metabolic pathways has shown promise in preclinical studies, though with limited translation due to toxicity and poor absorption.
    \item Homoharringtonine (HHT):  As noted, HHT, a protein synthesis inhibitor, has been approved for CML and has shown promise in AML, although its mechanism of action and specific impacts on resistance mechanisms need to be further explored.
    \item Immunomodulatory drugs: Thalidomide and lenalidomide have been explored in combination therapies to target the tumor microenvironment and enhance immune responses, though with limited efficacy in AML.
    \item Kinase inhibitors: Beyond FLT3, new kinase inhibitors that target other pathways are being explored in combination settings to overcome resistance.
\end{itemize}
Despite these findings, there remains a significant unmet need for novel, effective therapies that can overcome drug resistance, target novel pathways, and improve long-term survival for AML patients.

\end{tcolorbox}
\vspace{0.1cm}
\caption{\textbf{Detailed AI co-scientist output for repurposing KIRA6, an IRE1$\alpha$ inhibitor for AML}.}
\label{fig:ex_kira6_1}
\end{figure}

\begin{figure}[htbp!]
\begin{tcolorbox}[
    colback=black!5!white,
    colframe=black!60!white,
    title=\textbf{KIRA6 for AML (continued)},
    fonttitle=\bfseries,
    arc=3mm,
    boxrule=1pt,
    bottomrule=2pt,
]
\scriptsize

\textbf{Areas worth exploring} 

Several areas hold significant potential for identifying novel or repurposed drugs for AML treatment:
\begin{itemize}
    \item Targeting non-canonical signaling pathways: Explore drugs that target pathways beyond the well-established ones, such as those involved in inflammatory signaling, protein degradation, and DNA damage repair. The interplay between the AML cell and its microenvironment, including inflammatory cytokines and immune cells, remains an area of active investigation and a promising target for therapeutic intervention.
    \item Targeting RNA processing and translation: Aberrant RNA processing and translation are significant drivers of protein expression in cancer, and are potential targets for novel therapies. Targeting ribosome function, mRNA splicing, and RNA modifications can disrupt the production of key proteins required for cell survival and proliferation.
    \item Modulating immunometabolism: Cancer cells exhibit altered metabolic demands that influence the immune response, and targeting metabolic pathways that influence immune activation and suppression could enhance anti-leukemic immune responses. This also includes the metabolic dependencies of the myeloid microenvironment, which is critical for disease progression and resistance.
    \item Leveraging high-throughput screening (HTS) and computational biology: Combine in vitro or ex vivo screening of approved drugs with computational drug repurposing methods, particularly those that integrate multi-omics data, to identify novel drug candidates and potential synergistic drug combinations.
    \item Focus on drug combinations: Rational drug combinations that address both active malignancy and underlying resistance mechanisms, and can achieve effective treatment regimens at lower concentrations in combination, will have better safety, efficacy and overcome existing resistance. This should include a focus on sequential drug administration and drug scheduling.
    \item RNA-based therapies: Direct targeting of mRNAs coding for leukemia-driving proteins can be a very impactful strategy that also offers the chance to overcome resistance, for example by interfering with compensatory pathways. This includes RNA interference (RNAi), antisense oligonucleotides (ASOs), and mRNA-based therapies.
    \item Targeting protein homeostasis: Disrupting the protein homeostasis network, including protein folding, degradation (through ubiquitin-proteasome and autophagy pathways), and quality control mechanisms, can induce cellular stress and apoptosis in AML cells.
\end{itemize}

\textbf{Detailed novel likely correct idea: repurposing the IRE1$\alpha$ inhibitor KIRA6 for AML treatment\\}

\textbf{Goal}: Repurpose KIRA6, an IRE1$\alpha$ inhibitor, for AML treatment, specifically targeting the MOLM13 cell line with FLT3-ITD, with a focus on overcoming resistance mechanisms and enhancing combination therapy efficacy.
\\
Idea: KIRA6, by inhibiting IRE1$\alpha$, a key regulator of the unfolded protein response (UPR), can disrupt protein homeostasis and induce ER stress in AML cells, particularly in cells exhibiting high metabolic demand and increased protein synthesis, such as FLT3-ITD positive cells. This disruption can synergize with other therapies, including FLT3 inhibitors and chemotherapeutics, to overcome resistance and improve therapeutic efficacy. KIRA6 will also supress chronic inflammatory pathways known to promote cell survival in AML cells.

\end{tcolorbox}
\vspace{0.1cm}
\caption{\textbf{Detailed AI co-scientist output for repurposing KIRA6, an IRE1$\alpha$ inhibitor for AML (continued)}.}
\label{fig:ex_kira6_2}
\end{figure}

\begin{figure}[htbp!]
\begin{tcolorbox}[
    colback=black!5!white,
    colframe=black!60!white,
    title=\textbf{KIRA6 for AML (continued)},
    fonttitle=\bfseries,
    arc=3mm,
    boxrule=1pt,
    bottomrule=2pt,
]
\scriptsize
\textbf{Molecular mechanism of action}
\begin{itemize}
    \item Inhibition of IRE1$\alpha$: KIRA6 is a selective inhibitor of IRE1$\alpha$, a transmembrane protein in the endoplasmic reticulum (ER) that acts as a key sensor and transducer of ER stress. IRE1$\alpha$ activation initiates the unfolded protein response (UPR) pathway, which is a cellular stress response aimed at restoring protein homeostasis in the ER. Under conditions of increased protein synthesis or misfolded proteins, IRE1$\alpha$ is activated, leading to the splicing of XBP1 mRNA and the subsequent activation of downstream transcriptional targets involved in protein folding, trafficking, and degradation.
    \item Disruption of ER homeostasis: By inhibiting IRE1$\alpha$, KIRA6 blocks the adaptive arm of the UPR, preventing the resolution of ER stress and leading to the accumulation of unfolded and misfolded proteins. This disruption creates a protein folding crisis, inducing ER stress and initiating apoptotic pathways. This effect is potentiated in rapidly proliferating AML cells, which have higher metabolic demands and are more sensitive to ER stress.
    \item Downstream effects:
    \begin{itemize}
        \item  The accumulation of misfolded proteins leads to the activation of the PERK and ATF6 arms of the UPR, which further contribute to ER stress and apoptosis.
        \item  Inhibition of IRE1$\alpha$ disrupts the transcriptional program regulated by XBP1, impairing the production of proteins involved in ER homeostasis and cell survival.
        \item  The resulting ER stress induces the activation of the integrated stress response (ISR), leading to the translational suppression of many transcripts and increased apoptotic activity.
        \item  KIRA6 can indirectly inhibit the NF-$\kappa$B pathway by reducing ER stress and inflammatory cytokine production, reducing cell survival, proliferation and resistance.
        \item  KIRA6 may directly reduce levels of inflammatory cytokines such as IL-1, in turn reducing secondary activation of inflammatory signalling cascades and reduce activation of IRAK1 in the process.
    \end{itemize}
    \item Impact on FLT3-ITD: FLT3-ITD mutations lead to increased cell proliferation and metabolic demand and stress. KIRA6 can target this through disruption of the UPR, leading to a higher impact on FLT3-ITD cells. FLT3-ITD cells are under high levels of stress already, and require high levels of protein synthesis to maintain viability. By targeting basal or activated IRE1$\alpha$, KIRA6 can induce significantly more cell death in FLT3-ITD cells than their wild-type counterparts.
    \item Impact on key dysregulated pathways:
    \begin{itemize}
        \item  MYC: MYC protein levels are directly tied to mRNA translation and are necessary for cell survival and resistance in many cases, including leukemia. Targeting the UPR and downstream translation with KIRA6 directly decreases MYC expression and survival. This also has downstream anti-inflammatory benefits.
        \item  NF-$\kappa$B signaling: KIRA6 can reduce NF-$\kappa$B activity, a key driver of cell survival, proliferation, and resistance, by reducing ER stress and inflammatory cytokine production.
        \item  MCL-1 and other anti-apoptotic proteins: KIRA6 will reduce the production of short-lived survival proteins, leading to rapid apoptosis by diminishing their production, particularly MCL-1 and other similar proteins, which are involved in anti-apoptotic effects in AML cells, and are known drug resistant mechanisms.
        \item  Targeting multiple AML resistance mechanisms: KIRA6, by disrupting protein homeostasis, has the capacity to overcome resistance across a wide panel of mechanisms, including through reduced efflux pump protein levels, and reduced repair mechanisms.
    \end{itemize}
    \item Synergistic effects: KIRA6 can synergize with FLT3 inhibitors by increasing ER stress, making cells more sensitive to FLT3 inhibition. It can also enhance the activity of chemotherapeutic drugs by disrupting DNA repair pathways.
\end{itemize}

\textbf{Impacted pathways}
\begin{itemize}
    \item  Unfolded protein response (UPR): Directly impacted through inhibition of IRE1$\alpha$.
    \item  Apoptosis: Primarily through the intrinsic pathways, and disruption of anti-apoptotic proteins, and an increase in ER stress, alongside an increase of pro-apoptotic proteins through the integrated stress response.
    \item  Cell cycle: Perturbation of cell cycle progression through reduced levels of cell-cycle proteins.
    \item  NF-$\kappa$B signaling: KIRA6 can reduce NF-$\kappa$B activity which reduces inflammation and promotes cell survival, and drug resistance.
    \item  MYC-related pathways: Suppression of MYC and other related oncogene protein levels.
    \item  Inflammatory pathways: By supressing cytokines, most notably IL-1, KIRA6 can suppress activation of IRAK1 and reduce chronic inflammation that promotes cell survival in addition to killing cancer cells.
\end{itemize}

\end{tcolorbox}
\vspace{0.1cm}
\caption{\textbf{Detailed AI co-scientist output for repurposing KIRA6, an IRE1$\alpha$ inhibitor for AML (continued)}.}
\label{fig:ex_kira6_3}
\end{figure}

\begin{figure}[htbp!]
\begin{tcolorbox}[
    colback=black!5!white,
    colframe=black!60!white,
    title=\textbf{KIRA6 for AML (continued)},
    fonttitle=\bfseries,
    arc=3mm,
    boxrule=1pt,
    bottomrule=2pt,
]
\scriptsize
\textbf{Effect on AML (MOLM13) cell proliferation inhibition}
\begin{itemize}
    \item KIRA6 selectively inhibits proliferation of MOLM13 cells, and also AML cells with other mutations that render them drug resistant.
    \item IRE1$\alpha$ inhibition leads to the activation of apoptotic pathways and reduced cell proliferation and clonal survival.
    \item Suppression of MYC levels, NF-kB, MCL-1 and other cell survival pathways will reduce cell survival and promote apoptosis, which should be enhanced in FLT3-ITD AML due to their high proliferation rates and increased sensitivity to metabolic/translational stress in comparison to normal myeloid cells.
    \item In addition to having a direct effect on proliferation, a reduction in inflammatory signalling (IL-1 and IRAK1) can potentiate activity of other drugs.
    \item KIRA6 treatment results in a dose-dependent reduction in MOLM13 cell viability.
    \item KIRA6 leads to a buildup of ER stress and activation of the ISR, all which enhance its effectiveness as a targeted drug.
    \item When combined with FLT3 inhibitors or chemotherapeutic drugs, KIRA6 is expected to show additional impacts at lower overall drug concentrations (lower levels of both KIRA6 or the alternative drugs in the combination will be required for the same levels of cell death, or greater cell death will be seen at the standard doses used presently in the clinic). This is very important given the high levels of both drug toxicity that are associated with all AML treatments, and the need to reduce those toxicities.
\end{itemize}

\textbf{IC50 assay concentrations}

Based on published data for KIRA6 in other cancer cell models and its known mechanism of action, a reasonable starting concentration range for an IC50 assay in MOLM13 cells would be 100 nM to 10 $\mu$M.
\begin{itemize}
    \item Rationale: This range spans the concentrations that have shown activity in various cell types while being within a pharmacologically achievable range. Prior in vitro studies have shown activity in low micromolar concentrations.
    \item Specific concentrations: 100 nM, 250 nM, 500 nM, 1 $\mu$M, 2.5 $\mu$M, 5 $\mu$M, 7.5 $\mu$M, 10 $\mu$M.
    \item  Assay method: Cell viability assays (e.g., MTT, CellTiter-Glo) to determine IC50 values after KIRA6 exposure for 48-72 hours.
    \item Controls: Appropriate vehicle controls (DMSO) or medium-only controls should be included.
    \item Positive control:  Use a known FLT3 inhibitor, such as gilteritinib, at its reported IC50 in MOLM13 cells for normalization of assay variability and comparison.
    \item Combination studies: The combination of a FLT3 inhibitor at a single set concentration, plus a matrix of KIRA6 concentrations at several ratios will be evaluated using the viability assays as a baseline. Additional confirmation using apoptotic assays and cell cycle arrest should be done for any combination that shows better results than single-agent drug effects.
\end{itemize}

\textbf{Safety and toxicity}
\begin{itemize}
    \item Limited safety data: KIRA6 is a relatively new compound and has limited safety data in humans, as it has not gone through clinical trials.  Initial in vitro and in vivo studies in other disease settings (mostly cancer) suggest that it is generally well-tolerated. However, thorough preclinical toxicity studies are essential before advancing to human trials.
    \item Potential toxicity: Given the mechanism of action, potential toxicities include ER stress-related effects on normal tissues, particularly those with high protein synthesis demands (e.g., liver, pancreas). However, since AML cells are under high levels of stress, and have high metabolic demand, they are expected to be disproportionately impacted by this drug, which should reduce systemic toxicity.
    \item Repurposing advantage: Because KIRA6 has established in vitro activity, and has known pharmacology, it is significantly less risky than a compound that would have to have all of its safety profiles evaluated before human trials.
    \item Combination therapy impact: Because KIRA6 can be paired with existing therapies, it may be possible to reduce the overall dose of both agents to minimize side-effects and overall toxicity, while enhancing therapeutic efficacy. This should be tested rigorously in vitro through all of the combination studies before any movement into clinical human or animal trials to make sure that the benefit is higher than the risk.
\end{itemize}

\textbf{Testable hypothesis}

KIRA6, by inhibiting IRE1$\alpha$, will selectively inhibit the proliferation of AML cells, especially FLT3-ITD positive cells, by disrupting protein homeostasis and inducing ER stress. The increase in stress and reduced output of essential cell maintenance proteins will trigger apoptotic pathways, leading to cell death, particularly in highly proliferative settings. KIRA6 will potentiate the activity of existing AML therapies such as FLT3 inhibitors and chemotherapeutics. The drug will also have a broader impact on resistance mechanisms.

\end{tcolorbox}
\vspace{0.1cm}
\caption{\textbf{Detailed AI co-scientist output for repurposing KIRA6, an IRE1$\alpha$ inhibitor for AML (continued)}.}
\label{fig:ex_kira6_4}
\end{figure}

\begin{figure}[htbp!]
\begin{tcolorbox}[
    colback=black!5!white,
    colframe=black!60!white,
    title=\textbf{KIRA6 for AML (continued)},
    fonttitle=\bfseries,
    arc=3mm,
    boxrule=1pt,
    bottomrule=2pt,
]
\scriptsize
\textbf{Experimental plan}
\begin{itemize}
    \item IC50 determination in MOLM13: Determine IC50 values in MOLM13 cells using viability assays with KIRA6 concentrations ranging from 100 nM to 10 $\mu$M.
    \item ER stress marker measurement: Measure levels of key ER stress markers (e.g., BiP/GRP78, CHOP) using Western blotting and ELISA after KIRA6 treatment to confirm the impact on ER homeostasis.
    \item IRE1$\alpha$ activity: Measure XBP1 splicing using RT-PCR or other assays to confirm KIRA6's impact on IRE1$\alpha$ activity.
    \item Apoptosis assays: Evaluate apoptosis by Annexin V/PI staining, caspase activity assays, and Western blot analysis of apoptotic markers (cleaved PARP, cleaved caspase-3 levels) in KIRA6-treated MOLM13 cells.
    \item Western blots: Analyze key pathway proteins: including MYC, NF-$\kappa$B, MCL-1, FLT3, AKT/mTOR, and key components of the canonical inflammatory pathways (including IL-1, IL-6, TNF-alpha, IRAK1/4, MyD88), and markers of endoplasmic reticulum (ER) stress in AML cells treated with KIRA6.
    \item ISR markers measurement: Evaluate the induction of the Integrated Stress Response using assays measuring levels of phospho-eIF2$\alpha$, ATF4, and CHOP.
    \item Combination studies: Evaluate the combination of KIRA6 with FLT3 inhibitors like gilteritinib, with chemotherapeutics (cytarabine, danunorubicin) or with venetoclax in MOLM13 cells using combination matrix studies. This will establish if there is synergistic activity and enable evaluation of lower doses of each drug. Repeat these on additional cell lines with and without FLT3 mutations to confirm consistent synergistic enhancement.
    \item Cell cycle analysis: Measure cell cycle arrest using DNA staining (propidium iodide) by flow cytometry.
    \item RNA stress markers studies: Evaluate markers of RNA stress to assess damaged mRNA levels and ER Stress markers to determine cellular stress induced by the drug.
    \item Efflux pump activity: Measure the activity of key drug efflux pumps (e.g., ABCB1, ABCG2) using specific inhibitors and substrates to evaluate KIRA6's impact on drug resistance mechanisms.
    \item Expansion to other cell lines: Repeat the above assays on multiple AML cell lines, with both FLT3-ITD and wild-type cells, and lines representing other AML mutations or drug resistance phenotypes. Also evaluate primary AML patient cells in co-culture with stroma and alone with comparable controls, including primary non-leukemic bone marrow derived cells for measuring minimal activity in normal myeloid cells. This should allow for verification of the generalizability of KIRA6
    \item In vivo xenograft studies: Evaluate the efficacy and safety of KIRA6 in mouse xenograft models using MOLM13 cells or other cell lines to assess its overall treatment value and toxicities.
    \item Pharmacokinetics studies: Evaluate absorption, distribution, metabolism, and excretion of KIRA6 in animals to enable rational dose selection in future trials.
\end{itemize}

\textbf{Conclusion}

Repurposing KIRA6, an IRE1$\alpha$ inhibitor, for AML treatment, with a focus on FLT3-ITD positive cells and overcoming resistance mechanisms, represents a promising novel approach. KIRA6's ability to disrupt protein homeostasis, induce ER stress, synergize with existing AML drugs, and potentially overcome resistance mechanisms makes it an excellent drug candidate for testing. The comprehensive experimental plan proposed here offers a thorough framework for validating the potential of KIRA6 in AML and allows for a better way to select drugs for clinical testing by first performing a rigorous scientific evaluation in vitro using a multi-faceted approach. KIRA6 could translate into a new and effective strategy for AML patients, particularly those with the challenging FLT3-ITD mutation and other resistance phenotypes. Due to its mode of action on the UPR, it also has the potential to target underlying resistance mechanisms to existing therapies.

The focus on resistance mechanisms, a more diverse set of cell lines and primary cells, a well-rationalized combination approach, and a specific mechanism of action, all address the feedback from the previous version and provide a strong and detailed rationale for this drug repurposing strategy. The expanded experimental plan provides a thorough framework for assessing the efficacy, selectivity, and safety of KIRA6 in AML.
\end{tcolorbox}
\vspace{0.1cm}
\caption{\textbf{Detailed AI co-scientist output for repurposing KIRA6, an IRE1$\alpha$ inhibitor for AML (continued)}.}
\label{fig:ex_kira6_5}
\end{figure}

\begin{figure}[htbp!]
\begin{tcolorbox}[
    colback=black!5!white,
    colframe=black!60!white,
    title=\textbf{KIRA6 for AML (continued)},
    fonttitle=\bfseries,
    arc=3mm,
    boxrule=1pt,
    bottomrule=2pt,
]
\scriptsize
\textbf{Review}

\textbf{Relevant article abstracts}
\begin{itemize}
\item [1]: Discusses targeting the IRE1$\alpha$-XBP1s pathway in AML, directly relevant to the proposed mechanism of KIRA6.
\item [5]: Reviews the UPR as a therapeutic target in acute leukemias, providing context and supporting the rationale for targeting IRE1$\alpha$.
\item [6]: Focuses on selective degradation of FLT3-ITD, offering insights into targeting specific forms of the mutated protein, aligning with the idea's focus on targeting ER-bound FLT3-ITD.
\item [7]: Offers the background on known pharmacology and toxicological studies for IC50 determination of drug candidates in tumor cells.
\item [9]: Discusses sensitizing AML to venetoclax treatment, highlighting the potential for combination therapy approaches, relevant to the idea's suggested synergistic effects with other AML drugs.
\end{itemize}

\textbf{Assumptions}
\begin{itemize}
    \item KIRA6 effectively inhibits IRE1$\alpha$ in AML cells, particularly MOLM13 cells.
    \item IRE1$\alpha$ inhibition leads to significant disruption of protein homeostasis and induction of ER stress in AML cells.
    \item The increased ER stress induced by KIRA6 selectively targets AML cells over healthy cells.
    \item KIRA6 synergizes with FLT3 inhibitors and chemotherapeutic agents to enhance AML cell death.
    \item The proposed IC50 concentration range (100 nM to 10 $\mu$M) is appropriate for testing in MOLM13 cells.
    \item KIRA6 has a manageable safety profile and tolerable toxicity in preclinical models.
    \item KIRA6 can inhibit the production of short-lived proteins, like MYC and MCL-1, and reduce inflammatory signaling.
\end{itemize}

\textbf{Reasoning about assumptions}
\begin{itemize}
    \item Plausible: [1] and [5] support the notion of targeting IRE1$\alpha$ in AML. While no direct information shows if KIRA6 effective inhibits IRE1$\alpha$ in AML MOLM 3 cells,  it likely works as an IRE1$\alpha$ inhibitor like in other cell types. Needs experimental validation.
    \item Plausible: This is a core function of the UPR pathway in which IRE1$\alpha$ inhibition, based on the literature [5], should induce this cascade of events causing stress response leading to cell death if prolonged and severe.
    \item Plausible, but requires careful investigation: AML cells, particularly those with FLT3-ITD, have higher protein synthesis rates and are inherently more sensitive to ER stress. While this is not demonstrated in the literature, the rationale behind targeting ER stress response in leukemia is that tumor cells, particularly with FLT3-ITD, are expected to be more reliant on these processes and therefore may be more susceptible, but this assumption needs further evaluation.
    \item Plausible, but requires testing: The rationale for synergy is based on the proposed mechanism and general principles of targeting multiple pathways in cancer. [9] offers a related example of synergy with venetoclax, a drug also being tested in FLT3-ITD AML. This hypothesis needs further in vitro testing to be demonstrated.
    \item Plausible: The suggested range is based on literature and seems reasonable as a starting point. This needs to be empirically confirmed in an IC-50 assay. The provided range in the idea is a well justified assumption that can be verified during the IC-50 viability assays where a broader range of concentrations would capture the optimal values.
    \item Unknown: This is a vital assumption requiring thorough preclinical testing. [7] has information about pharmacology and toxicity, which provides context. Currently, limited safety data exists for KIRA6 so the drug safety profile is unknown and needs experiments to verify if the drug is safe enough.
    \item Plausible: IRE1$\alpha$ inhibition and ER stress induction can trigger downstream effects that impact protein translation and inflammatory responses based on the mechanism described in the UPR [5]. [1] suggests that IRE1$\alpha$ impacts venetoclax efficacy. While this is not demonstrated in the literature for KIRA6 in AML, this remains an experimentally verifiable assumption.
\end{itemize}

\textbf{Improvements to the idea}
\begin{itemize}
    \item Broader cell line testing: The idea primarily focuses on MOLM13. Expanding to other AML cell lines, including those with different FLT3 mutations and other genetic backgrounds, would strengthen the proposal. This would require additional experiments as well.
    \item Deeper exploration of resistance mechanisms: While mentioned, the idea could benefit from a more detailed analysis of how KIRA6 addresses specific resistance mechanisms in AML. The idea can also benefit from additional experiments regarding resistance mechanism impacts.
    \item Investigate drug resistance: There is minimal data regarding KIRA6 impacting drug efflux pumps and other drug resistance mechanisms. It would be beneficial to investigate how impacting the UPR could impact drug resistant cells, which are a major reason for treatment failure in AML.
    \item Detailed comparison to other IRE1$\alpha$ inhibitors: If there are other IRE1$\alpha$ inhibitors, comparing and contrasting KIRA6 with them and justifying the reasons for choosing this particular drug would add to the idea's strength.
\end{itemize}
\end{tcolorbox}
\vspace{0.1cm}
\caption{\textbf{Detailed AI co-scientist output for repurposing KIRA6, an IRE1$\alpha$ inhibitor for AML (continued)}.}
\label{fig:ex_kira6_6}
\end{figure}

\begin{figure}[htbp!]
\begin{tcolorbox}[
    colback=black!5!white,
    colframe=black!60!white,
    title=\textbf{KIRA6 for AML (continued)},
    fonttitle=\bfseries,
    arc=3mm,
    boxrule=1pt,
    bottomrule=2pt,
]
\scriptsize

\textbf{Reasoning about correctness and testing}

The idea is plausible and well-reasoned based on known mechanisms of IRE1$\alpha$ and the UPR in cancer. However, key assumptions about selectivity and synergy require experimental validation. The proposed experimental plan is comprehensive and addresses the key questions.

The idea is novel in its application of KIRA6 to AML, particularly in the context of FLT3-ITD. The focus on combination therapy and overcoming resistance is also promising.

It is recommended to test this idea due to its plausibility, novelty, and potential clinical significance. While there are uncertainties regarding safety and efficacy, the potential benefits warrant further investigation.

\textbf{Answer: 4}
\\
\\
\textbf{Novelty review}
\begin{itemize}
    \item Targeting the IRE1$\alpha$-XBP1s pathway to enhance venetoclax effectiveness in AML [1]: This abstract directly relates to the idea by exploring the IRE1$\alpha$ pathway in AML, although in combination with venetoclax, not as a monotherapy or in combination with FLT3 inhibitors.
    \item The unfolded protein response: A novel therapeutic target in acute leukemias [5]: This review discusses the UPR as a target in acute leukemias, providing a broader context for the idea's focus on IRE1$\alpha$.
    \item Selective degradation of mutant FMS-like tyrosine kinase-3 requires BIM-dependent depletion of heat shock proteins [6]: This abstract explores the mechanisms of FLT3 degradation and its connection to BIM and heat shock proteins, relevant to the idea's focus on FLT3-ITD positive AML.
\end{itemize}

\textbf{Already explored aspects}
\begin{itemize}
    \item  Targeting the UPR in AML: The idea of targeting the unfolded protein response (UPR) in AML is not entirely novel. Abstract [5] discusses the UPR as a therapeutic target in acute leukemias, including AML. Abstract [1] specifically explores targeting IRE1$\alpha$ in combination with venetoclax, suggesting existing interest in this pathway for AML.  However, the use of KIRA6 as a single agent or in combination with FLT3 inhibitors appears less explored.
    \item FLT3 inhibitors in AML: Targeting FLT3 in AML with tyrosine kinase inhibitors is a well-established approach, with multiple approved drugs.  Abstracts [2, 3] focus on FLT3 inhibitors, highlighting the importance of this target but also the challenges of resistance [6].
\end{itemize}

\textbf{Novel aspects}
\begin{itemize}
    \item Repurposing KIRA6 for AML: While IRE1$\alpha$ has been targeted in AML [1], the repurposing of KIRA6 specifically for AML treatment, especially in the context of FLT3-ITD positive disease and potential combination with FLT3 inhibitors or standard chemotherapies, seems novel based on the provided abstracts.  The idea explores KIRA6's impact on multiple pathways (UPR, apoptosis, NF-$\kappa$B, MYC) related to AML proliferation and resistance.
    \item Focusing on FLT3-ITD positive AML \& combination therapy: Although FLT3 inhibitors are used in AML, the idea of combining KIRA6 with FLT3 inhibitors or chemotherapeutics to specifically target FLT3-ITD positive AML cells and potentially overcome resistance addresses a current clinical need. The proposed mechanism - disrupting protein homeostasis in already stressed FLT3-ITD cells - offers a rationale for this combination approach.
    \item Combination of KIRA6 with other AML drugs \& mechanism of action: KIRA6 is not mentioned in any of the abstracts, and therefore the proposed combination of KIRA6 and FLT3i to induce apoptosis is also novel. The described mechanism has strong logical support, as the cells' increased translation requirements and ER stress from the FLT3i mutation could greatly increase the effectiveness of IRAK1 inhibition, which may itself directly induce apoptosis, and also improve efficacy of existing drugs.
\end{itemize}

\textbf{Novelty review}

The idea presents a moderate level of novelty. Repurposing KIRA6 for AML, particularly in the context of FLT3-ITD positive disease and combination therapy, is a promising approach. However, targeting IRE1$\alpha$ in AML has been explored [1], albeit not with this specific drug and with different existing treatment approaches.  The idea's strength lies in its detailed mechanistic rationale, combination therapy focus, and defined experimental plan.  It is crucial to validate the proposed mechanism and selectivity across different AML cell lines and patient samples before concluding its true novelty. \\

\textbf{Improvements to the idea}
\begin{itemize}
    \item Literature search: Conduct a thorough literature search beyond the provided abstracts to confirm the novelty of KIRA6 in AML. Investigate any existing research on KIRA6 in other hematological malignancies.
    \item Selectivity testing:  Expand selectivity testing to include other healthy cell types beyond normal myeloid cells, to address potential off-target effects and demonstrate greater safety margin.
    \item Combination exploration: Further explore rational combinations beyond FLT3 inhibitors, including chemotherapeutic agents and other targeted therapies.
    \item Resistance mechanisms:  Thoroughly investigate the impact of KIRA6 on various AML resistance mechanisms through in vitro testing, using resistant cell lines and patient-derived samples.
    \item In vivo efficacy and toxicity profile: Expand the in vivo studies with multiple cell lines and patient-derived xenograft models, with a focus on establishing the efficacy and toxicity profile of KIRA6, both as single-agent and in the proposed combination strategy.
\end{itemize}

\end{tcolorbox}
\vspace{0.1cm}
\caption{\textbf{Detailed AI co-scientist output for repurposing KIRA6, an IRE1$\alpha$ inhibitor for AML (continued)}.}
\label{fig:ex_kira6_7}
\end{figure}

\begin{figure}[htbp!]
\begin{tcolorbox}[
    colback=black!5!white,
    colframe=black!60!white,
    title=\textbf{KIRA6 for AML (continued)},
    fonttitle=\bfseries,
    arc=3mm,
    boxrule=1pt,
    bottomrule=2pt,
]
\scriptsize
\textbf{Reasoning about novelty and recommendation}

The idea is novel enough to warrant further exploration.  While targeting IRE1$\alpha$ isn't completely new, using KIRA6 in AML, especially in the context outlined, hasn't been extensively investigated, and it also has not been previously examined as a companion drug with other therapies used in AML.  The proposed combination with FLT3 inhibitors and the detailed rationale provide a strong foundation. The comprehensive experimental plan should be executed to confirm the preliminary findings and assess the true potential of KIRA6 in AML.  If the in vitro and in vivo findings are positive, the idea would be worthy of publication in a specialized journal focused on hematological malignancies or drug repurposing.

\textbf{Answer: 3}
\\

[1] Targeting the IRE1$\alpha$-XBP1s Pathway to Enhance Venetoclax Effectiveness in AML | Blood | American Society of Hematology
(https://ashpublications.org/blood/article/144/Supplement\%201/4144/533082/Targeting-the-IRE1-XBP1s-Pathway-to-Enhance)

[2] FLT3 inhibitors in acute myeloid leukemia: ten frequently asked questions | Leukemia
(https://www.nature.com/articles/s41375-019-0694-3)

[3] Inhibition of FLT3-ITD Kinase in Acute Myeloid Leukemia by New Imidazo[1,2-b]pyridazine Derivatives Identified by Scaffold Hopping | Journal of Medicinal Chemistry
(https://pubs.acs.org/doi/10.1021/acs.jmedchem.3c00575)

[4] Repurposing approved drugs on the pathway to novel therapies - PMC
(https://pmc.ncbi.nlm.nih.gov/articles/PMC7018532/)

[5] The Unfolded Protein Response: A Novel Therapeutic Target in Acute Leukemias - PMC
(https://pmc.ncbi.nlm.nih.gov/articles/PMC7072709/)

[6] Drugging the unfolded protein response in acute leukemias | Journal of Hematology \& Oncology | Full Text
(https://jhoonline.biomedcentral.com/articles/10.1186/s13045-015-0184-7)

[7] Selective degradation of mutant FMS-like tyrosine kinase-3 requires BIM-dependent depletion of heat shock proteins | Leukemia
(https://www.nature.com/articles/s41375-024-02405-5)

[8] Pharmacology and Toxicology Testing: IC50 for Tumor Cell Lines - Altogen Labs
(https://altogenlabs.com/pre-clinical-research-services/pharmacology-and-toxicology-testing-ic50-for-tumor-cell-lines/)

[9] Targeting Mitochondrial Structure Sensitizes Acute Myeloid Leukemia to Venetoclax Treatment | Cancer Discovery | American Association for Cancer Research
(https://aacrjournals.org/cancerdiscovery/article/9/7/890/41983/Targeting-Mitochondrial-Structure-Sensitizes-Acute)
\\
\\
\textbf{Critiques}

Here's a summary of the negative critiques from the reviews:
\begin{itemize}
    \item Limited safety data: The biggest concern is the lack of human safety data for KIRA6, requiring thorough preclinical toxicity studies.  The drug's safety profile is currently unknown and needs experimental validation.
    \item Limited scope of cell line testing:  Testing primarily focuses on MOLM13. Expanding to other AML cell lines (with different FLT3 mutations and other genetic backgrounds) and primary patient samples is necessary.
    \item Limited comparison to other drugs:  The idea would benefit from comparing KIRA6 to other IRE1$\alpha$ inhibitors or similar drugs to justify its selection and benchmark efficacy/selectivity. Minimal data exists regarding KIRA6 compared to other drugs.
    \item Lack of pre-clinical and clinical data:  While proposed, in vivo and toxicity data are currently lacking for KIRA6, despite the comprehensive experimental plan.
    \item Assumptions requiring validation: Several assumptions about KIRA6's selectivity for AML cells over healthy cells, synergy with other drugs, impact on drug resistance mechanisms, and the suggested IC50 concentration range need experimental validation.
    \item Limited detail in some areas: The rationale for the specific IC50 range could be stronger. The exploration of resistance mechanisms and the details of synergy studies could also be more in-depth. More detail on how KIRA6 impacts MYC, NF-kB and inflammatory pathways is needed.  The connection between UPR activation and a lower incidence of FLT3 mutations needs further investigation.
    \item Potential for resistance: While the idea aims to overcome resistance, there's still a need to investigate how KIRA6 itself might contribute to new resistance mechanisms over time.
\end{itemize}

Although the idea is considered plausible, novel, and promising, these weaknesses need to be addressed through further research and experimentation.

\end{tcolorbox}
\vspace{0.1cm}
\caption{\textbf{Detailed AI co-scientist output for repurposing KIRA6, an IRE1$\alpha$ inhibitor for AML (continued)}.}
\label{fig:ex_kira6_8}
\end{figure}

\clearpage
\section{An example of tool use in the AI co-scientist with AlphaFold}
\label{sec:alphafold}
The AI co-scientist is a general purpose system broadly applicable across different areas of science and medicine. To better understand the capabilities and limitations of the system, we task it with the goal of suggesting protein sequences with specific properties. Determining the correct primary amino acid sequence with the desired properties is an essential part of protein engineering. While LLM-based systems can predict protein properties and suggest modifications~\citep{wang2025large}, they can sometimes generate incorrect sequences (i.e., hallucinations). To address this, we integrate AlphaFold~\citep{jumper2021highly}, a specialized AI system for predicting protein 3D structure, into our co-scientist. AlphaFold acts as a validation tool, evaluating the structural plausibility of sequences proposed by the co-scientist and provides feedback. This increases the reliability of the sequence design optimization process, which can be further validated with wet laboratory experiments. The approach to integrate tools highlights how specialized AI models can work in collaboration with more general AI systems like the AI co-scientist, facilitating the solution of complex challenges like protein design.

\begin{figure}[htbp!]
\centering
\includegraphics[width=0.92\textwidth,keepaspectratio]{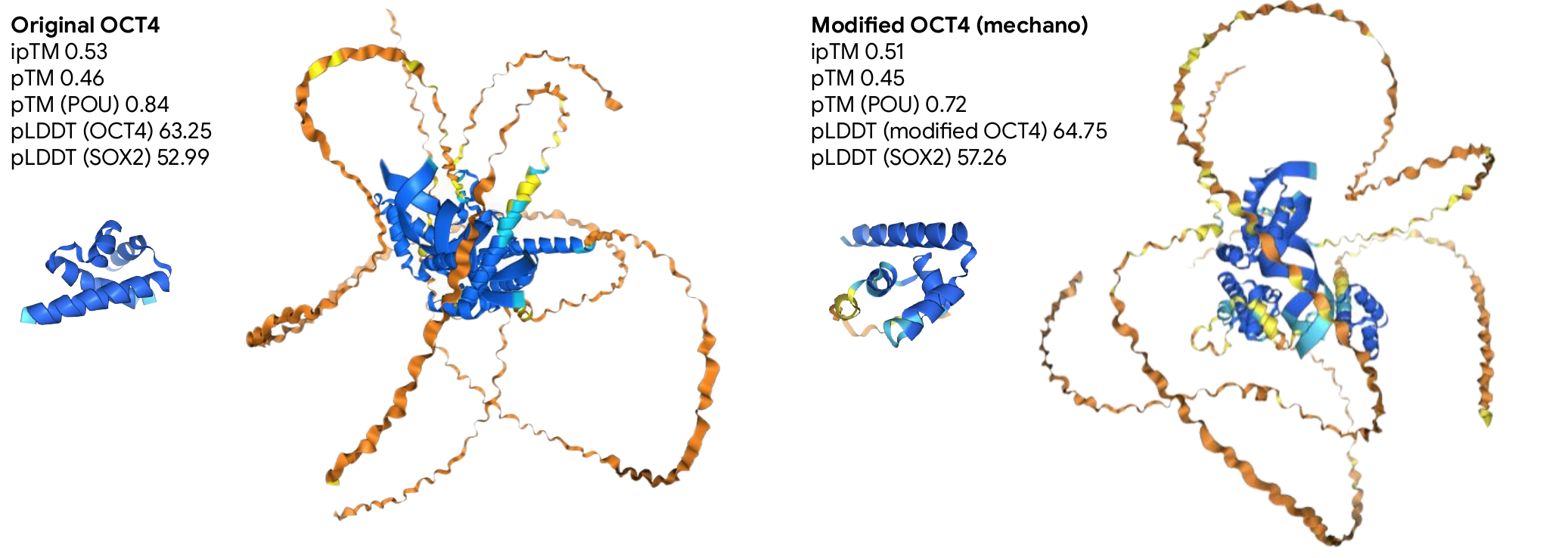}
\vspace{0.1cm}
\caption{\textbf{AlphaFold predicted protein 3D structure and metrics for original OCT4 and AI co-scientist suggested modifications.} (left panel) original OCT4 sequence with SOX2 and DNA binding (right panel) modified OCT4 sequence with SOX2 and DNA binding. The left 3D structure in each panel is the POU domain of the corresponding OCT4 sequence. The predicted template modeling (pTM) score, the interface predicted template modeling (ipTM), and predicted local distance difference test (pLDDT) are derived from the AlphaFold outputs.}
\label{fig:fig_af}
\end{figure}

As an illustrative example, we used AlphaFold to assess a co-scientist's proposed modification to the OCT4 (octamer-binding transcription factor 4) protein (Appendix \cref{fig:fig_af}), one of the four Yamanaka factors~\citep{takahashi2007induction}, to increase binding affinity of its DNA binding domain. The co-scientist suggested adding a mechano-sensitive loop to the POU domain (a family of eukaryotic transcription factors) and a dynamic phosphorylation site outside of it. The co-scientist first verified the proposed sequence against the UniProt database via web-search. AlphaFold then predicted the 3D structure of the modified protein, suggesting that the modifications maintained structural stability. These predictions were used to refine the co-scientist's hypothesis, allowing it to improve its protein sequence design in subsequent iterations. We also independently validated the modification using ESM-2 \citep{lin2022language} and RoseTTAFold \citep{baek2021accurate}. ESM-2 predicted an increased log-likelihood ratio and a similar predicted local distance difference test (pLDDT), and RoseTTAFold predicted similar confidence score (GDT), compared to the original sequence. The insertion and modification did not seem to disrupt SOX2 and OCT4 interactions, indicated by the similar pLDDT scores between the original and modified OCT4 sequences. However, this example is for demonstration purposes only. Further \textit{in silico} analysis (e.g., predicting binding affinity and off-target effects), and thorough laboratory validation are necessary to confirm that the proposed modifications actually improve the complex roles of OCT4 binding, while maintaining SOX2 interaction integrity, during pluripotency.

\vspace{-0.3cm}

Combining AlphaFold with the co-scientist framework offers a powerful approach for both improving existing proteins and designing entirely new ones. This integrated system allows researchers to iteratively optimize protein sequences for enhanced properties (e.g., stability, binding affinity, or catalytic activity) or putatively to create proteins with novel functions. It enables exploration of protein design while ensuring structural feasibility. Future work will focus on experimentally validating these capabilities and applying them to targeted protein design efforts as well as expansion to integration of other specialized AI tools with the co-scientist.

\newpage
\setlength\bibitemsep{3pt}
\printbibliography
\balance
\clearpage

\end{refsection}

\end{document}